%% file: main.tex
\PassOptionsToPackage{unicode}{hyperref}
\PassOptionsToPackage{hyphens}{url}
\PassOptionsToPackage{dvipsnames,svgnames,x11names}{xcolor}
\documentclass[
  12pt]{article}
\usepackage{graphicx} 
\usepackage{amsmath, amsthm, amsfonts,amssymb}
\usepackage{bm}
\usepackage{indentfirst}
\usepackage{amsmath}
\usepackage{mathrsfs}
\usepackage{subfiles}
\usepackage{verbatim}
\usepackage{enumitem}
\usepackage{float}
\usepackage{caption}
\usepackage{subcaption}
\usepackage{algorithm}
\usepackage{algorithmic}
\usepackage{longtable}
\usepackage{multirow}
\usepackage{multicol}
\usepackage{xcolor}
\usepackage{latexsym}
\usepackage{amsthm}
\usepackage{amsthm,amsmath,amssymb}
\usepackage{mathrsfs}
\usepackage{natbib}
\setcitestyle{authoryear,round}
\usepackage[colorlinks, citecolor=blue]{hyperref}

\newtheorem{theorem}{Theorem}

\newtheorem{assumption}{Assumption}

\newcommand{\argmin}{\mathop{\rm arg\min}}

\newtheorem{remark}{Remark}
\newcommand{\bE}{\mathbb{E}}
\newcommand{\bP}{\mathbb{P}}
\newcommand{\cP}{\mathcal{P}}

\newcommand{\cR}{\mathcal{R}}
\newcommand{\bx}{\boldsymbol{x}}
\newcommand{\bmu}{\boldsymbol{\mu}}
\newcommand{\bxi}{\boldsymbol{\xi}}

\newcommand{\btheta}{\boldsymbol{\theta}}

\newcommand{\by}{\boldsymbol{y}}

\newcommand{\bb}{\boldsymbol{b}}
\newcommand{\bz}{\boldsymbol{z}}
\newcommand{\br}{\boldsymbol{r}}
\newcommand{\bu}{\boldsymbol{u}}
\newcommand{\bv}{\boldsymbol{v}}
\newcommand{\be}{\boldsymbol{e}}
\newcommand{\cN}{\mathcal{N}}

\newtheorem{example}{Example}
\newtheorem{proposition}{Proposition}
\usepackage{amsmath,amssymb}
\usepackage{iftex}
\ifPDFTeX
  \usepackage[T1]{fontenc}
  \usepackage[utf8]{inputenc}
  \usepackage{textcomp} 
\else 
  \usepackage{unicode-math}
  \defaultfontfeatures{Scale=MatchLowercase}
  \defaultfontfeatures[\rmfamily]{Ligatures=TeX,Scale=1}
\fi
\usepackage{lmodern}
\ifPDFTeX\else  
\fi
\IfFileExists{upquote.sty}{\usepackage{upquote}}{}
\IfFileExists{microtype.sty}{
  \usepackage[]{microtype}
  \UseMicrotypeSet[protrusion]{basicmath} 
}{}
\makeatletter
\@ifundefined{KOMAClassName}{
  \IfFileExists{parskip.sty}{%
    \usepackage{parskip}
  }{
    \setlength{\parindent}{0pt}
    \setlength{\parskip}{6pt plus 2pt minus 1pt}}
}{
  \KOMAoptions{parskip=half}}
\makeatother
\usepackage{xcolor}
\makeatletter
\ifx\paragraph\undefined\else
  \let\oldparagraph\paragraph
  \renewcommand{\paragraph}{
    \@ifstar
      \xxxParagraphStar
      \xxxParagraphNoStar
  }
  \newcommand{\xxxParagraphStar}[1]{\oldparagraph*{#1}\mbox{}}
  \newcommand{\xxxParagraphNoStar}[1]{\oldparagraph{#1}\mbox{}}
\fi
\ifx\subparagraph\undefined\else
  \let\oldsubparagraph\subparagraph
  \renewcommand{\subparagraph}{
    \@ifstar
      \xxxSubParagraphStar
      \xxxSubParagraphNoStar
  }
  \newcommand{\xxxSubParagraphStar}[1]{\oldsubparagraph*{#1}\mbox{}}
  \newcommand{\xxxSubParagraphNoStar}[1]{\oldsubparagraph{#1}\mbox{}}
\fi
\makeatother

\usepackage{longtable,booktabs,array}
\usepackage{calc} 
\usepackage{etoolbox}
\makeatletter
\patchcmd\longtable{\par}{\if@noskipsec\mbox{}\fi\par}{}{}
\makeatother
\IfFileExists{footnotehyper.sty}{\usepackage{footnotehyper}}{\usepackage{footnote}}
\makesavenoteenv{longtable}
\usepackage{graphicx}
\makeatletter
\def\maxwidth{\ifdim\Gin@nat@width>\linewidth\linewidth\else\Gin@nat@width\fi}
\def\maxheight{\ifdim\Gin@nat@height>\textheight\textheight\else\Gin@nat@height\fi}
\makeatother
\setkeys{Gin}{width=\maxwidth,height=\maxheight,keepaspectratio}
\makeatletter
\def\fps@figure{htbp}
\makeatother

\makeatletter
\@ifpackageloaded{caption}{}{\usepackage{caption}}
\AtBeginDocument{%
\ifdefined\contentsname
  \renewcommand*\contentsname{Table of contents}
\else
  \newcommand\contentsname{Table of contents}
\fi
\ifdefined\listfigurename
  \renewcommand*\listfigurename{List of Figures}
\else
  \newcommand\listfigurename{List of Figures}
\fi
\ifdefined\listtablename
  \renewcommand*\listtablename{List of Tables}
\else
  \newcommand\listtablename{List of Tables}
\fi
\ifdefined\figurename
  \renewcommand*\figurename{Figure}
\else
  \newcommand\figurename{Figure}
\fi
\ifdefined\tablename
  \renewcommand*\tablename{Table}
\else
  \newcommand\tablename{Table}
\fi
}
\@ifpackageloaded{float}{}{\usepackage{float}}
\floatstyle{ruled}
\@ifundefined{c@chapter}{\newfloat{codelisting}{h}{lop}}{\newfloat{codelisting}{h}{lop}[chapter]}
\floatname{codelisting}{Listing}

\makeatother
\makeatletter
\@ifpackageloaded{caption}{}{\usepackage{caption}}
\@ifpackageloaded{subcaption}{}{\usepackage{subcaption}}
\makeatother

\ifLuaTeX
  \usepackage{selnolig}  
\fi
\usepackage[]{natbib}
\usepackage{bookmark}

\IfFileExists{xurl.sty}{\usepackage{xurl}}{} 
\hypersetup{
  pdftitle={Title},
  pdfauthor={Author 1; Author 2},
  pdfkeywords={3 to 6 keywords, that do not appear in the title},
  colorlinks=true,
  linkcolor={blue},
  filecolor={Maroon},
  citecolor={Blue},
  urlcolor={Blue},
  pdfcreator={LaTeX via pandoc}}

\newcommand{\anon}{1}

\begin{document}


\input{paper_body.tex}

\clearpage

\setcounter{section}{0}
\renewcommand{\thesection}{S\arabic{section}}
\setcounter{equation}{0}
\renewcommand{\theequation}{S\arabic{equation}}
\setcounter{theorem}{0}
\renewcommand{\thetheorem}{S\arabic{theorem}}
\setcounter{lemma}{0}
\renewcommand{\thelemma}{S\arabic{lemma}}
\setcounter{corollary}{0}
\renewcommand{\thecorollary}{S\arabic{corollary}}
\setcounter{proposition}{0}
\renewcommand{\theproposition}{S\arabic{proposition}}
\setcounter{definition}{0}
\renewcommand{\thedefinition}{S\arabic{definition}}
\setcounter{figure}{0}
\renewcommand{\thefigure}{S\arabic{figure}}

\input{supp_body.tex}

\bibliography{ref}

\end{document}

%% file: paper_body.tex
\def\spacingset#1{\renewcommand{\baselinestretch}%
{#1}\small\normalsize} \spacingset{1}


\if1\anon
{
  \title{\bf Synthetic Augmentation in Imbalanced Learning: When It Helps, When It Hurts, and How Much to Add}
  \author{Zhengchi Ma\hspace{.2cm}\\
    Department of Electrical \& Computer Engineering, Duke University\\
    and \\
    Anru R. Zhang\thanks{Corresponding author: \href{mailto: anru.zhang@duke.edu}{anru.zhang@duke.edu}} \\
    Department of Biostatistics \& Bioinformatics\\ and Department of Computer Science, Duke University}
    \date{}
  \maketitle
} \fi

\if0\anon
{
  \bigskip
  \bigskip
  \bigskip
  \begin{center}
    {\LARGE\bf Title}
\end{center}
  \medskip
} \fi

\bigskip
\begin{abstract}
Imbalanced classification often causes standard training procedures to prioritize the majority class and perform poorly on rare but important cases. A classic and widely used remedy is to augment the minority class with synthetic samples, but two basic questions remain under-resolved: when does synthetic augmentation actually help, and how many synthetic samples should be generated?

We develop a unified statistical framework for synthetic augmentation in imbalanced learning, studying models trained on imbalanced data augmented with synthetic minority samples. Our theory shows that synthetic data is not always beneficial. In a ``local symmetry'' regime, imbalance is not the dominant source of error, so adding synthetic samples cannot improve learning rates and can even degrade performance by amplifying generator mismatch. When augmentation can help (``local asymmetry''), the optimal synthetic size depends on generator accuracy and on whether the generator’s residual mismatch is directionally aligned with the intrinsic majority–minority shift. This structure can make the best synthetic size deviate from naive full balancing. Practically, we recommend Validation-Tuned Synthetic Size (VTSS): select the synthetic size by minimizing balanced validation loss over a range centered near the fully balanced baseline, while allowing meaningful departures. Extensive simulations and real data analysis further support our findings.
\end{abstract}

\noindent%
\vfill

\newpage
\spacingset{1.8} 

\section{Introduction}\label{sec:intro}

Imbalanced classification, where one class is observed far less frequently than the other, is a pervasive obstacle in modern statistical learning. Standard empirical risk minimization tends to favor overall accuracy, which often leads to systematically degraded detection of the minority class. This failure mode is especially consequential when the rare events are clinically, economically, or operationally important, as in medical data analysis \citep{salmi2024handling}, finance \citep{chen2024interpretable}, and industrial anomaly detection \citep{bougaham2024composite}. These challenges motivate a renewed statistical focus on mitigating imbalance in a principled way, without turning augmentation into a purely heuristic step \citep{chen2024survey}.

A classic response to imbalance is to augment the minority class with additional (real or synthetic) samples, so that model training is less dominated by the majority class. More broadly, synthetic volume expansion has been proposed as a way to improve downstream data analysis through the use of additional synthetic samples \citep{shen2023boosting}. The simplest mechanism is bootstrap-style oversampling, i.e., resampling minority observations with replacement \citep{efron1994introduction}. A widely used alternative is SMOTE, which creates new minority examples by interpolating between a minority observation and its neighbors \citep{chawla2002smote}. SMOTE has inspired many extensions, including ADASYN \citep{he2008adasyn}, Borderline-SMOTE \citep{han2005borderline}, safe-level-SMOTE \citep{bunkhumpornpat2009safe}, and DBSMOTE \citep{bunkhumpornpat2012dbsmote}. Related augmentation strategies include Mixup \citep{zhang2017mixup} and conditional data synthesis approaches \citep{tian2025conditional}. Beyond classical generators, modern deep generative models can also serve as synthetic sample generators, including VAEs \citep{kingma2013auto}, normalizing flows \citep{papamakarios2021normalizing}, GANs \citep{goodfellow2014generative}, diffusion/score-based models \citep{ho2020denoising} and attention-based large language models \citep{nakada2024synthetic}. 

\paragraph*{What remains unclear: when does augmentation help, and how much is enough?}
Despite the breadth of methods and applications, two basic questions remain surprisingly under-resolved from a statistical perspective.

First, \emph{synthetic augmentation is not guaranteed to help}. Because synthetic samples are generated from an estimated minority distribution, augmentation can reduce the impact of imbalance while also introducing \emph{generator mismatch} and changing the effective variance. Its overall effect on the target performance criterion therefore depends on both the data geometry and the quality of the generator. Second, \emph{even when augmentation is beneficial, it is unclear how many synthetic samples should be added}. A common practice is \emph{naive balancing}: generate enough synthetic minority samples so that the minority-plus-synthetic count matches the majority count. This rule is simple and often reasonable, but it is largely heuristic. Crucially, synthetic samples generally do not carry the same information content as independent real samples, and their mismatch can interact with the sample size in nontrivial ways. Relatedly, in a discussion paper, \citet{shen2023boosting} studies synthetic volume expansion and shows that downstream performance can be non-monotonic in the amount of synthetic data, with a problem-dependent regime beyond which additional synthetic samples may yield diminishing returns or even become counterproductive.

\paragraph*{Our perspective: treat augmentation as a controllable statistical operation.}
This paper develops a unified framework that treats synthetic augmentation as a \emph{controllable} statistical operation and analyzes the risk trade-offs it induces under a \emph{balanced} evaluation criterion, i.e., one treating two classes symmetrically at the population level. This criterion is common in imbalanced learning when minority detection is a primary goal. Our framework applies to a broad class of training losses, classification rules, and synthetic-sample generators, and it is parameterized by two quantities: the \emph{synthetic size} and the \emph{generator mismatch}. Our results show how these two factors jointly determine the statistical value of augmentation.

\paragraph*{Two guiding questions and the regimes they reveal.}
We use our framework to answer two high-level questions that map directly to practice.

\begin{quote}
    {\it Q1. Does augmenting with synthetic samples always help in imbalanced learning?}
\end{quote}

Not always. The key distinction is whether imbalance creates a \emph{first-order} distortion of the learning objective near the population optimum under the balanced criterion. When the two classes exert different first-order influence in optimization-relevant directions, changing the effective class proportions can shift the learned solution toward the balanced target. We call this the \emph{local asymmetry} regime, and it is precisely the setting in which synthetic augmentation can improve performance.
Our analysis also identifies an important complementary regime in which synthetic augmentation \emph{cannot} help, even under severe imbalance. In this \emph{local symmetry} regime, the two classes already exert equal first-order influence near the balanced optimum in the directions that matter for optimization. Then imbalance is not the bottleneck, so adding synthetic minority samples yields little benefit; if the generator is imperfect, it can instead amplify mismatch effects and degrade performance. We develop this phenomenon in detail in Section~\ref{sec:synnohelp}.

\begin{quote}
    {\it Q2. Is naive balancing truly the best?}
\end{quote}

Naive balancing is often a sensible default, but it is not universally optimal. Its optimality depends on generator quality and on whether the residual synthetic mismatch aligns with the intrinsic majority–minority difference in the learning problem. When the generator is consistent and its remaining mismatch has no exploitable directional structure, naive balancing is essentially rate-optimal: it removes the leading imbalance effect, and finer tuning can improve performance only by a constant factor. By contrast, when the mismatch is directionally aligned with the majority–minority discrepancy, a \emph{small} adjustment around naive balancing can further reduce the leading bias and produce a measurable gain. For systematically biased (inconsistent) generators, tuning the synthetic size may even be necessary for consistency, and the optimal choice can be far from naive balancing when bias cancelation is possible; when the mismatch direction is not aligned, changing the quantity alone cannot remove the bias.

\paragraph*{A practical recommendation: treat synthetic size as a tunable hyperparameter.}
In real applications, the relevant regime and mismatch direction are rarely known a priori. Motivated by the theory, we propose \emph{Validation-Tuned Synthetic Size (VTSS)}: sweep candidate synthetic sizes and select the one minimizing balanced validation loss. VTSS is simple to implement, captures gains from nontrivial sizing, and automatically avoids harmful over-synthesis when augmentation is unhelpful.

\begin{figure*}[t]
  \centering
  \includegraphics[width=\linewidth]{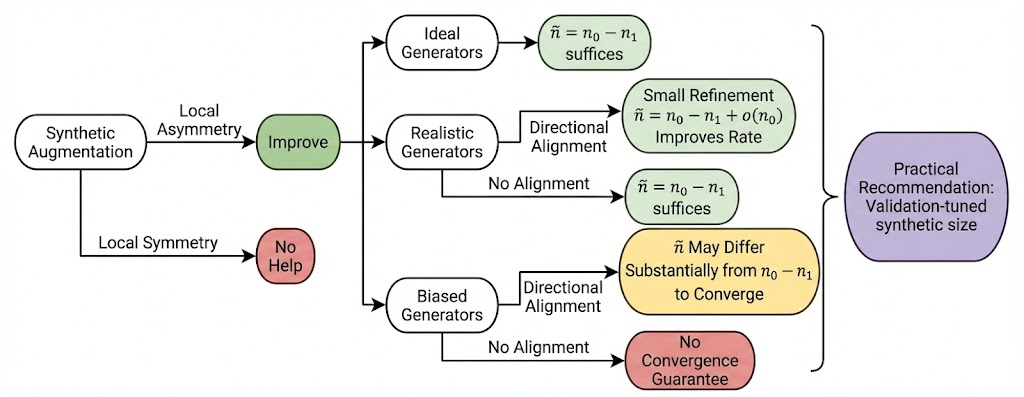}
  \caption{Flowchart of main results}
  \label{fig:flowchart}
\end{figure*}

The main messages of our theory are summarized in Figure~\ref{fig:flowchart}, which maps the local asymmetry/symmetry regimes and generator quality to when synthetic augmentation is helpful and how the synthetic size should be chosen.
Our framework accommodates a broad class of training losses, classification rules, and synthetic-sample generators.

\paragraph*{Related work.}
Recent theoretical work has begun to clarify statistical properties of learning with synthetic data. \cite{nakada2024synthetic} provides theoretical guarantees for when LLM-generated minority samples can improve learning under imbalance. \cite{lyu2025bias} emphasizes that naively treating synthetic samples as real can introduce bias and proposes bias-correction strategies. For uncertainty quantification and inference, \cite{keret2025glm} studies inference with AI-generated synthetic data for generalized linear models, while \cite{raisa2025consistent} develops a Bayesian framework that accounts for the synthetic generation mechanism and establishes conditions for consistent inference. \cite{xia2026classification} frames imbalanced classification as transfer learning under label shift. \cite{ahmad2025concentration} develops uniform concentration bounds comparing empirical risk on synthetic samples to the true minority risk and derives excess-risk guarantees for kernel-based classifiers.

A growing body of applied work illustrates synthetic augmentation in downstream analyses. In medical imaging, \cite{ktena2024generative} uses conditional diffusion models for label-steerable augmentation and studies robustness and fairness under distribution shift. \cite{wang2025self} studies self-improving generative foundation models for medical images. In finance, \cite{roy2024frauddiffuse} uses diffusion-generated transaction samples to augment scarce fraud cases. In industrial applications, \cite{stanton2024data} synthesizes time-series sensor data to support downstream predictive maintenance modeling.

Our work differs from these by providing a unified, risk-based characterization of when synthetic augmentation helps versus when it cannot, and by showing that the \emph{choice of synthetic size} is a first-order practical decision rather than a fixed heuristic. Our theory isolates qualitative regimes (local asymmetry vs.\ local symmetry), explains the bias--variance trade-offs induced by generator mismatch and synthetic size, and motivates VTSS as a robust practical procedure.

Relatedly, class imbalance can also be addressed without generating data, through the choice of objective, decision rule, and evaluation metric. Existing work considered three common paradigms: (i) classical risk minimization (often paired with reweighting or resampling), (ii) cost-sensitive learning that minimizes a weighted combination of type I and type II errors, and (iii) the Neyman--Pearson (NP) paradigm, which controls type I error at a prespecified level while minimizing type II error \citep{rigollet2011neyman,tong2020neyman,feng2021imbalanced}. These perspectives motivate alternatives to oversampling such as loss reweighting, threshold adjustment, and constraint-based formulations.

\paragraph*{Organization.}
The rest of this paper is organized as follows. Section~\ref{sec:formulation} introduces notation and the formal setup, including the balanced evaluation objective. Section~\ref{sec:excess-risk} develops our main excess-risk characterization, which makes explicit how performance depends on both class-proportion imbalance and generator-induced distributional mismatch. Section~\ref{sec:whenhelps} studies the local asymmetry regime in which synthetic augmentation can improve learning, and shows how the optimal synthetic size depends on generator quality and the directional structure of its bias. Section~\ref{sec:synnohelp} identifies a local symmetry regime in which imbalance is not the dominant bottleneck, so synthetic augmentation cannot improve rates and may degrade performance when the generator is imperfect. Section~\ref{sec:validation-selection} then presents our practical procedure, Validation-Tuned Synthetic Size (VTSS), which selects the synthetic size by minimizing balanced validation loss. Section~\ref{sec:simulation} provides simulation studies that verify the qualitative predictions of the theory and evaluate VTSS, and Section~\ref{sec:applications} reports a real-data application using electronic health records (EHR) data. Section~\ref{sec:discussions} concludes with discussion and future directions. Proofs, additional simulation and real data results, and additional technical details are deferred to the Supplementary Materials, which also include further examples for the losses (Section~\ref{sec:loss}), classifiers (Section~\ref{sec:classifiers}), and synthetic-sample generators (Section~\ref{sec:generators}) covered by our framework.

\section{Notation and Problem Formulation}\label{sec:formulation}

\paragraph*{Notation.}
For nonnegative sequences $\{a_n\}$ and $\{b_n\}$, we write $a_n\lesssim b_n$ (resp.\ $a_n\gtrsim b_n$) if there exists a universal constant $C>0$ such that $a_n\leq C b_n$ (resp.\ $a_n\geq C b_n$). We write $a_n\asymp b_n$ if both $a_n\lesssim b_n$ and $b_n\lesssim a_n$ hold, and write $a_n\ll b_n$ (resp.\ $a_n\gg b_n$) if $a_n/b_n\to 0$ (resp.\ $a_n/b_n\to\infty$).
Vectors are denoted by boldface letters (e.g., $\bx,\by,\bu,\bv$), and $\|\cdot\|$ denotes the Euclidean norm. Gradients and Hessians with respect to $\btheta$ are denoted by $\nabla$ and $\nabla^2$. For two nonzero vectors $\bu,\bv$ of the same dimension, we denote their angle and the sine value by
\(
\angle(\bu,\bv):=\arccos \left(\frac{\bu^\top\bv}{\|\bu\|  \|\bv\|}\right),~
\sin\angle(\bu,\bv)=\sqrt{1-\left(\frac{\bu^\top\bv}{\|\bu\|  \|\bv\|}\right)^2}.\)
For a symmetric matrix $A$, let $\lambda_{\max}(A)$ and $\lambda_{\min}(A)$ denote its largest and smallest eigenvalues. For symmetric matrices $A$ and $B$, we write $A\succeq B$ (resp.\ $A\succ B$) if $A-B$ is positive semidefinite (resp.\ positive definite).
Throughout, we work in an asymptotic regime where the data-generating distributions are fixed while the sample size grows. We use standard stochastic order notation $O_P(\cdot)$, $o_P(\cdot)$, $\Omega_P(\cdot)$, and $\Theta_P(\cdot)$. In particular, for sequences indexed by $(n_0,n_1)$, we write $x_{(n_0,n_1)}=O_P(a_{(n_0,n_1)})$ as $n_0,n_1\to\infty$ if $x_{(n_0,n_1)}/a_{(n_0,n_1)}$ is stochastically bounded, and write $x_{(n_0,n_1)}=o_P(1)$ if $x_{(n_0,n_1)}\to 0$ in probability. We write $x_{(n_0,n_1)}=\Omega_P(a_{(n_0,n_1)})$ if $x_{(n_0,n_1)}/a_{(n_0,n_1)}$ is stochastically bounded away from zero, and $x_{(n_0,n_1)}=\Theta_P(a_{(n_0,n_1)})$ if both $O_P(\cdot)$ and $\Omega_P(\cdot)$ hold.

\paragraph*{Problem formulation.}
We consider binary classification under class imbalance, where $y=0$ denotes the majority class and $y=1$ denotes the minority class. Let the class-conditional distributions be
$
\bx\mid(y=0)\sim \cP_0,~ \bx\mid(y=1)\sim \cP_1.
$
We observe $n_0$ majority samples and $n_1$ minority samples $(n_0>n_1)$, and we may augment the data with $\tilde n$ synthetic minority samples. Synthetic covariates are drawn from a generator-induced distribution $\cP_{\text{syn}}$; since the generator is typically trained on the observed data, $\cP_{\text{syn}}$ may depend on $(n_0,n_1)$ and should therefore be written more precisely as \(\cP_{\text{syn},n_0,n_1}\). We suppress this dependence for notational simplicity.

Let $\ell(\btheta;\bx,y)$ be a loss function with parameter $\btheta\in\Theta$. Common choices include cross-entropy/logistic loss, squared loss, hinge loss, and exponential loss (see Section~\ref{sec:loss} of the Supplementary Materials for detailed discussions). To treat the two classes symmetrically at the population level, we define the \textbf{balanced population risk}
\begin{equation}\label{def:balancedrisk_rev}
\cR(\btheta)
=\frac12\,\bE_{\cP_0}\ell(\btheta;\bx,0)+\frac12\,\bE_{\cP_1}\ell(\btheta;\bx,1),
\end{equation}
and suppose there exists a minimizer
$
\btheta^*=\argmin_{\btheta\in\Theta}\cR(\btheta).
$ Augmenting the training data with $\tilde n$ synthetic minority samples changes the effective class proportions (see Section~\ref{sec:loss} of the Supplement for a discussion of specific synthetic data generators used to produce these samples). Let
$
\pi_0=\frac{n_0}{n_0+n_1+\tilde n},~
\pi_1=\frac{n_1}{n_0+n_1+\tilde n},~
\tilde\pi=\frac{\tilde n}{n_0+n_1+\tilde n}$;
so that $\pi_0+\pi_1+\tilde\pi=1$. The resulting \textbf{synthetic population risk} is
\[
\tilde{\cR}(\btheta)
=\pi_0\,\bE_{\cP_0}\ell(\btheta;\bx,0)
+\pi_1\,\bE_{\cP_1}\ell(\btheta;\bx,1)
+\tilde\pi\,\bE_{\cP_{\text{syn}}}\ell(\btheta;\bx,1),
\]
with minimizer $\tilde\btheta=\argmin_{\btheta\in\Theta}\tilde{\cR}(\btheta)$. 
In practice, we observe $\{(\bx_i,y_i)\}_{i=1}^{n_0+n_1}$ and generate synthetic samples
$
\{\tilde{\bx}_i\}_{i=1}^{\tilde n}\sim \cP_{\text{syn}},~ \tilde y_i\equiv 1.
$
Then minimize the {\bf empirical augmented risk}
\begin{align*}
\hat{\cR}(\btheta)
&=\frac{1}{n_0+n_1+\tilde n}\Bigg(
\sum_{i=1}^{n_0+n_1}\ell(\btheta;\bx_i,y_i)
+\sum_{i=1}^{\tilde n}\ell(\btheta;\tilde{\bx}_i,1)
\Bigg) \\
&=\frac{1}{n_0+n_1+\tilde n}\Bigg(
\sum_{i=1}^{n_0}\ell(\btheta;\bx_i^{(0)},0)
+\sum_{i=1}^{n_1}\ell(\btheta;\bx_i^{(1)},1)
+\sum_{i=1}^{\tilde n}\ell(\btheta;\tilde{\bx}_i,1)
\Bigg),
\end{align*}
where $\{\bx_i^{(0)}\}_{i=1}^{n_0}$ and $\{\bx_i^{(1)}\}_{i=1}^{n_1}$ are the observed majority and minority samples. Let
$
\hat\btheta=\argmin_{\btheta\in\Theta}\hat{\cR}(\btheta).
$
Our goal is to estimate the balanced-risk optimizer $\btheta^*$ from imbalanced data, and to understand how synthetic augmentation, through $\tilde n$ and $\cP_{\text{syn}}$,  affects the balanced excess risk $\cR(\hat\btheta)-\cR(\btheta^*)$.

\section{Excess Risk Analysis for Imbalanced Classification}\label{sec:excess-risk}

In this section, we establish a fundamental limitation on balanced excess risk with synthetic minority augmentation. The central obstruction is \emph{distribution bias} arising from (i) the effective class weighting in the augmented training population and (ii) mismatch between the synthetic distribution $\cP_{\mathrm{syn}}$ and the true minority distribution $\cP_1$. We first decompose the synthetic population risk, and use it to characterize $\cR(\hat{\btheta})-\cR(\btheta^*)$.

\paragraph*{A key decomposition of the synthetic risk.}
Recall the balanced population risk $\cR(\btheta)$ and the synthetic population risk $\tilde{\cR}(\btheta)$. A direct calculation yields
\begin{align}\label{psinotation}
\nonumber \tilde{\cR}(\btheta)
&=\,\cR(\btheta)
\\&+\Big(\pi_0-\tfrac12\Big)\Big[\bE_{\cP_{0}}\ell(\btheta;\bx,0)-\bE_{\cP_{1}}\ell(\btheta;\bx,1)\Big]
+\tilde{\pi}\Big[\bE_{\cP_{\text{syn}}}\ell(\btheta;\bx,1)-\bE_{\cP_{1}}\ell(\btheta;\bx,1)\Big].
\end{align}
To streamline notation, define
\begin{align}\label{eq:psi}
\phi(\btheta)
:= \bE_{\cP_{0}}\ell(\btheta;\bx,0)-\bE_{\cP_{1}}\ell(\btheta;\bx,1),
\qquad \psi(\btheta)
:= \bE_{\cP_{\text{syn}}}\ell(\btheta;\bx,1)-\bE_{\cP_{1}}\ell(\btheta;\bx,1),
\end{align}
where $\phi(\btheta)$ measures the local majority--minority asymmetry under the loss and $\psi(\btheta)$ quantifies the synthetic-to-minority discrepancy. With these definitions, \eqref{psinotation} becomes
\begin{equation}\label{eq:decomp_compact}
\tilde{\cR}(\btheta)=\cR(\btheta)+\Big(\pi_0-\tfrac12\Big)\phi(\btheta)+\tilde{\pi}\psi(\btheta).
\end{equation}
Here, $\big(\pi_0-\tfrac12\big)\phi(\btheta)+\tilde{\pi}\psi(\btheta)$ is the \emph{risk-level} distribution bias induced by class-proportion mismatch and generator mismatch. Since our excess-risk analysis is driven by first-order optimality conditions, we will work with the gradient of this bias term.
Accordingly, we define the corresponding \emph{first-order bias vector}
\(
\bb(\btheta):=\left(\pi_0-\tfrac12\right)\nabla\phi(\btheta)+\tilde{\pi}\nabla\psi(\btheta),
\)
which is exactly the $\btheta$-gradient of the bias component in \eqref{eq:decomp_compact}.

\paragraph*{A lower bound for balanced excess risk.}
The next result shows that, even asymptotically, the estimator trained on the augmented population can incur a non-vanishing balanced excess risk unless the induced bias is controlled.

\begin{theorem}[Excess Risk Lower Bound]\label{excess-lowerbound}
Assume the parameter space is bounded, $\btheta\in \{\btheta:\|\btheta\|< B\}$. Suppose the covariates $\bx$ have bounded support and the loss $\ell(\btheta;\bx,y)$ is continuous and differentiable in $\btheta$, and bounded on bounded $(\btheta,\bx)$. Assume $\cR(\btheta)$ is continuous and $\tilde{\cR}(\btheta)$ admits a unique minimizer. Further suppose that $\nabla\tilde{\cR}(\btheta)$ is Lipschitz at its minimizer $\tilde{\btheta}$, i.e.,
$
\|\nabla\tilde{\cR}(\btheta)-\nabla\tilde{\cR}(\tilde{\btheta})\|\leq M\|\btheta-\tilde{\btheta}\|,
$
and $\cR(\btheta)$ satisfies at its minimizer $\btheta^*$ that for any $\theta$:
$
\cR(\btheta)\geq \cR(\btheta^*)+\frac{\mu}{2}\|\btheta-\btheta^*\|^2.
$
Then the balanced excess risk satisfies
\[
\cR(\hat{\btheta})-\cR(\btheta^*)\geq \frac{\mu}{2M^2}\left\|\left(\pi_0-\tfrac12\right)\nabla\phi(\btheta^*)+\tilde{\pi}\nabla\psi(\btheta^*)\right\|^2+o_P(1).
\]
\end{theorem}
Theorem~\ref{excess-lowerbound} shows that the leading lower bound is governed by the squared norm of the \emph{first-order bias vector}
$
\bb(\btheta^*):=\left(\pi_0-\tfrac12\right)\nabla\phi(\btheta^*)+\tilde{\pi}\nabla\psi(\btheta^*).
$
The term proportional to $\pi_0-\tfrac12$ reflects the class-proportion mismatch in the augmented population, whereas the term proportional to $\tilde{\pi}$ captures generator mismatch via the discrepancy between $\cP_{\mathrm{syn}}$ and $\cP_1$. In particular, a necessary condition for balanced-risk consistency, i.e., $\cR(\hat{\btheta})-\cR(\btheta^*)\to 0$, is $\|\bb(\btheta^*)\|\to 0$.

\paragraph*{Excess Risk Asymptotic Representation.} We next state the conditions under which the excess risk admits an explicit asymptotic representation. These assumptions are standard regularity conditions in statistical learning theory and allow us to refine the lower bound in Theorem~\ref{excess-lowerbound} into an exact decomposition in Theorem~\ref{representation}.
\begin{assumption}[Condition for Excess Risk Asymptotic Representation]\label{assump:excess-risk-characterization}~
\begin{enumerate}
    \item[(A1)] The balanced risk $\cR$ is differentiable and satisfies at its minimizer $\btheta^*$ for a constant $\mu>0$ that
    $
    \cR(\btheta^*)\geq \cR(\btheta)+\nabla \cR(\btheta)^\top (\btheta^*-\btheta)+\frac{\mu}{2}\|\btheta^*-\btheta\|^2
    $. Also, it satisfies $\cR(\btheta)-\cR(\btheta^*) \geq L\|\btheta-\btheta^*\|^2$ at the minimizer $\btheta^*$ for some constant $L>0$;
    \item[(A2)] The covariates $\bx$ have bounded support and a finite third moment; The gradient of total bias is finite:
    $
    \left\|\left(\pi_0-1/2\right)\nabla^2\phi(\btheta)+\tilde{\pi}\nabla^2\psi(\btheta)\right\|\leq C<\infty.
    $
    \item[(A3)] $\nabla^2\tilde{\cR}(\btheta)\succeq \lambda I$ for some $\lambda>0$, and satisfies $\left\|\nabla^2\tilde{\cR}(\btheta)-\nabla^2\tilde{\cR}(\tilde{\btheta})\right\|\leq L\|\btheta-\tilde{\btheta}\|$ at the minimizer $\tilde{\btheta}$ for a constant $L>0$; 
    \item[(A4)] $\Sigma_0(\tilde{\btheta}),\Sigma_1(\tilde{\btheta}),\tilde{\Sigma}(\tilde{\btheta})\succ \lambda I$ for some $\lambda>0$, have finite largest eigenvalue. $\Sigma(\btheta)$ satisfies $\left\|\Sigma(\tilde{\btheta})-\Sigma(\btheta^*)\right\|\leq L\|\tilde{\btheta}-\btheta^*\|$ at two minimizers $\tilde{\btheta}$ and $\btheta^*$ for a constant $L>0$;
\end{enumerate}
\end{assumption}

\begin{theorem}[Excess Risk Decomposition]\label{representation}
Define the covariance matrices of the loss gradients:
\(
\Sigma_0(\btheta)=\text{Cov}_{\cP_0}(\nabla_{\btheta}\ell(\btheta;\bx,0)),~
\Sigma_1(\btheta)=\text{Cov}_{\cP_1}(\nabla_{\btheta}\ell(\btheta;\bx,1)),~ \tilde{\Sigma}(\btheta)=\text{Cov}_{\cP_{\text{syn}}}(\nabla_{\btheta}\ell(\btheta;\bx,1)),~
\Sigma(\btheta):=\frac{n_0\Sigma_0(\btheta)+n_1\Sigma_1(\btheta)+\tilde{n}\tilde{\Sigma}(\btheta)}{n_0+n_1+\tilde{n}}.\)
Suppose Assumption~\ref{assump:excess-risk-characterization} holds. For large enough $n_0$ and $n_1$, there exists a random vector $Z \sim \cN(0, I_d)$ such that
{\small \begin{align*}
& \cR(\hat{\btheta})-\cR(\btheta^*)\\
&=\frac{1}{2}\bb(\btheta^*)^\top \left[\nabla^2\cR(\btheta^*)+\nabla \bb(\btheta^*)\right]^{-1}\nabla^2\cR(\btheta^*)\left[\nabla^2\cR(\btheta^*)+\nabla \bb(\btheta^*)\right]^{-1} \bb(\btheta^*)\\
&+\frac{1}{\sqrt{n_0+n_1+\tilde{n}}}\bb(\btheta^*)^\top \left[\nabla^2\cR(\btheta^*)+\nabla \bb(\btheta^*)\right]^{-1}\nabla^2\cR(\btheta^*)\left[\nabla^2\cR(\btheta^*)+\nabla \bb(\btheta^*)\right]^{-1}\Sigma(\btheta^*)^{\frac{1}{2}}Z\\
&+\frac{1}{2(n_0+n_1+\tilde{n})}Z^\top \Sigma(\btheta^*)^{\frac{1}{2}}\left[\nabla^2\cR(\btheta^*)+\nabla \bb(\btheta^*)\right]^{-1}\nabla^2\cR(\btheta^*)\left[\nabla^2\cR(\btheta^*)+\nabla \bb(\btheta^*)\right]^{-1}\Sigma(\btheta^*)^{\frac{1}{2}}Z + R.
\end{align*}}
Here $\bb(\btheta)=\left(\pi_0-\frac{1}{2}\right)\nabla\phi(\btheta)+\tilde{\pi}\nabla\psi(\btheta)$, so  $\nabla\bb(\btheta)=\left(\pi_0-\frac{1}{2}\right)\nabla^2\phi(\btheta)+\tilde{\pi}\nabla^2\psi(\btheta)$ denotes the Jacobian. The remainder term $R$ satisfies
\begin{align*}
R=O_P\Big(&\|\bb(\btheta^*)\|^3+(n_0+n_1+\tilde{n})^{-3/2}
+\frac{1}{n_0+n_1+\tilde{n}}\|\bb(\btheta^*)\|
+\frac{1}{\sqrt{n_0+n_1+\tilde{n}}}\|\bb(\btheta^*)\|^2\Big).
\end{align*}
\end{theorem}

Theorem~\ref{representation} expresses the excess risk as the sum of three leading terms and a higher-order remainder. The first term is a quadratic form in $\bb(\btheta^*)$ and captures the deterministic \emph{distribution bias} induced by the effective weights $(\pi_0,\pi_1,\tilde{\pi})$ and any discrepancy between $\cP_{\mathrm{syn}}$ and $\cP_1$. The second and third terms quantify stochastic fluctuations through the covariance $\Sigma(\btheta^*)$ and the Gaussian limit $Z$, thereby making the bias--variance trade-off explicit as a function of $(n_0,n_1,\tilde{n})$. This representation is useful in two ways. First, since the right-hand side depends on $\tilde{n}$ through $\bb(\btheta^*)$, $\nabla \bb(\btheta^*)$, and $\Sigma(\btheta^*)$, it can be minimized with respect to $\tilde{n}$ to obtain principled synthetic-size choices. Second, different generators affect the expansion only through $\psi(\btheta)$ and $\tilde{\Sigma}(\btheta)$, providing a common asymptotic scale for comparing generators: holding $(n_0,n_1,\tilde{n})$ fixed, generators are preferable when they yield smaller bias $\bb(\btheta^*)$ and/or more favorable covariance contributions.

\begin{remark}[User-specific class weights]\label{rm:imbalance}
Our presentation focuses on the balanced objective \((\rho=\tfrac12)\) for clarity, but all results extend directly to user-specified class weights. Fix \(\rho\in(0,1)\) as the weight on the majority class and define
\(
\cR^\rho(\btheta)=\rho\,\bE_{\cP_0}\ell(\btheta;\bx,0)+(1-\rho)\,\bE_{\cP_1}\ell(\btheta;\bx,1).
\)
Then the synthetic population risk admits the decomposition
\(
\tilde{\cR}(\btheta)=\cR^\rho(\btheta)+(\pi_0-\rho)\,\phi(\btheta)+\tilde\pi\,\psi(\btheta),
\)
so the corresponding first-order bias vector becomes
\(
\bb_\rho(\btheta):=(\pi_0-\rho)\,\nabla\phi(\btheta)+\tilde\pi\,\nabla\psi(\btheta).
\)
Consequently, Theorems~\ref{excess-lowerbound} and~\ref{representation} hold after replacing \(\bb(\btheta)\) by \(\bb_\rho(\btheta)\) (and replacing \(\tfrac12\) by \(\rho\) wherever it appears). Moreover, consistency under the \(\rho\)-weighted objective requires matching the target proportion \(\pi_0\to\rho\), i.e.,
\(
\tilde n=\tfrac{1-\rho}{\rho}n_0-n_1+o(n_0),
\)
and, as in the \(\rho=\tfrac12\) case, lower-order deviations of \(\tilde n\) around this target can affect rates.
\end{remark}

In the remainder of this paper, we distinguish two regimes described as follows.
\begin{itemize}[leftmargin=*]
    \item \textbf{Local asymmetry}: $\|\nabla\phi(\btheta^*)\| \geq c$ for a fixed constant $c> 0$. In this case, synthetic minority augmentation may improve performance. In this regime, both generator quality (through $\|\nabla\psi(\btheta^*)\|$) and synthetic size $\tilde{n}$ matter: too few synthetic samples may not sufficiently mitigate imbalance, while too many may amplify generator-induced bias.
    \item \textbf{Local symmetry}: $\|\nabla\phi(\btheta^*)\| = 0$. In this case, synthetic minority samples typically do not help and may degrade performance.
\end{itemize}

\section{Local Asymmetry: When Synthetic Minority Samples May Improve Performance}\label{sec:whenhelps}

We now focus on the locally asymmetric regime, where $\|\nabla\phi(\btheta^*)\|\geq c>0$. Here the choice of synthetic sample size $\tilde n$ interacts with the generator mismatch $\|\nabla\psi(\btheta^*)\|$, leading to a bias--variance trade-off that determines whether augmentation improves or harms performance. Accordingly, we organize the analysis by generator quality, considering the following three regimes in Sections~\ref{sec:ideal}, \ref{sec:rate-improve}, and \ref{sec:inconsistent}, respectively.
\begin{enumerate}[leftmargin=*]
    \item \textbf{Ideal synthetic minority-class sample generator:}
    $\|\nabla\psi(\btheta^*)\|\ll n_0^{-1/2}$.
    This regime primarily serves as a theoretical benchmark. If the generator is trained using only the available minority data, then even under favorable conditions that avoid the curse of dimensionality (as is often assumed in nonparametric settings), the estimation error of the learned minority distribution is fundamentally limited by the sample size $n_1$, yielding at best a parametric-rate scaling of $n_1^{-1/2}$. More broadly, even when the majority and minority distributions share exploitable structure and the generator can, in principle, leverage information from both classes, the best achievable accuracy is still bounded by parametric scaling with the total sample size, namely $(n_0+n_1)^{-1/2}\asymp n_0^{-1/2}$ in the imbalanced regime. Consequently, requiring $\|\nabla\psi(\btheta^*)\|\ll n_0^{-1/2}$ is generally unattainable for standard, data-trained generators.

    \item \textbf{Realistic synthetic minority-class sample generator:}
    $n_0^{-1/2}\ll\|\nabla\psi(\btheta^*)\|\ll 1$.
    This corresponds to the practically relevant case, where the residual synthetic mismatch is non-negligible yet bounded.

    \item \textbf{Inconsistent synthetic minority-class sample generator:}
    $\|\nabla\psi(\btheta^*)\|\gtrsim 1$.
\end{enumerate}

The thresholds separating the ideal, realistic, and inconsistent regimes are well motivated: for standard data-trained generators, residual mismatch is typically limited by parametric-order scaling with the available sample size, making $\|\nabla\psi(\btheta^*)\|\ll n_0^{-1/2}$ unrealistic. At the other extreme, a mismatch bounded away from zero (i.e., $\|\nabla\psi(\btheta^*)\|\gtrsim 1$) indicates persistent distributional error and may compromise consistency. Nevertheless, the ideal and inconsistent regimes provide useful benchmarks for isolating algorithmic and statistical behavior.

\subsection{With Ideal Synthetic Sample Generators}\label{sec:ideal}

We first study the regime of \emph{ideal} synthetic generators, where the synthetic distribution $\cP_{\mathrm{syn}}$ closely matches the true minority distribution $\cP_1$. Intuitively, when the generator-induced mismatch is negligible, the precise choice of the synthetic sample size $\tilde n$ should matter only weakly: as long as $\tilde n$ is chosen at the appropriate order of $n_0-n_1$, the estimator based on synthetic augmentation should perform well. The next theorem formalizes this intuition. It shows that, under an ideal generator, any synthetic size $\tilde n$ that stays sufficiently close to the balancing choice $\tilde n = n_0-n_1$ yields the same $1/n_0$ convergence rate. In this regime, tuning $\tilde n$ can at best improve constant factors, but cannot improve the rate.

\begin{theorem}[Excess risk under an ideal synthetic generator]\label{thm:good-generator}
Assume the conditions of Theorem~\ref{representation}. Suppose that the curvature contribution from the synthetic mismatch is controlled in the sense that
\(
\|\nabla^2 \psi(\btheta^*)\|\leq \lambda_{\min} \left(\nabla^2 \cR(\btheta^*)\right).
\)
If the generator is ideal, satisfying $\|\nabla \psi(\btheta^*)\|\ll n_0^{-1/2}$, then for any synthetic size
\(
\tilde n \in \mathscr C
:=\left\{\tilde n:\ \tilde n = n_0-n_1+o(\sqrt{n_0})\right\},
\)
we have 
\(
\cR(\hat\btheta)-\cR(\btheta^*) =\Theta_P\left( n_0^{-1}\right).
\)
\end{theorem}

The set $\mathscr C$ requires the effective class proportions to be balanced around $1/2$ up to the natural $n_0^{-1/2}$ scale of stochastic fluctuations. Under an ideal generator, the leading bias term in Theorem~\ref{representation} is dominated by the stochastic term as long as $\tilde n\in\mathscr C$, and the excess risk attains the classical balanced parametric rate $\cR(\hat\btheta)-\cR(\btheta^*)=\Theta_P\left(n_0^{-1}\right)$. In particular, the naive balancing choice $\tilde n = n_0-n_1$ already achieves the optimal rate. Adjusting $\tilde n$ within $\mathscr C$ may improve constants, but cannot surpass the $n_0^{-1}$ rate.

\subsection{With Realistic Synthetic Sample Generators}\label{sec:rate-improve}

We next consider a more realistic regime in which the synthetic generator is statistically consistent, but its residual mismatch decays more slowly than the parametric rate:
\(
n_0^{-1/2}\ll \|\nabla\psi(\btheta^*)\|\ll 1.
\)
Our goal is to understand how the interaction between the intrinsic majority--minority imbalance and the generator-induced bias determines the appropriate order of the synthetic sample size $\tilde n$. To build intuition, we consider the following toy example.
\begin{example}[Toy collinear case]\label{ex:collinear-log}
Assume the generator is consistent with $\|\nabla\psi(\btheta^*)\|\asymp(\log n_1)^{-1/2}$, so $\|\nabla\psi(\btheta^*)\|\gg n_0^{-1/2}$.
Write the leading terms in Theorem~\ref{representation} as
\(
\cR(\hat{\btheta})-\cR(\btheta^*)=T_1+T_2+T_3+R,
\)
where $T_1\asymp\|\bb(\btheta^*)\|^2$, $T_2=\Theta_P(n^{-1/2}\|\bb(\btheta^*)\|)$, and $T_3=\Theta_P(n^{-1})$, with $n=n_0+n_1+\tilde n$.
Since consistency requires $\tilde n\approx n_0-n_1$, let $\tilde n=(n_0-n_1)(1+\delta_n)$ with $\delta_n\to 0$.
If $\nabla\psi(\btheta^*)$ is collinear with $\nabla\phi(\btheta^*)$, i.e.,
$\nabla\psi(\btheta^*)=s(\log n_1)^{-1/2}\nabla\phi(\btheta^*)$ for $s\in\{\pm 1\}$,
then one can choose $\delta_n=\Theta((\log n_1)^{-1/2})$ so that $\bb(\btheta^*)=0$.
Consequently, $T_1=T_2=0$ and $\cR(\hat{\btheta})-\cR(\btheta^*)=\Theta_P(n_0^{-1})$.
In contrast, the naive balancing choice $\tilde n=n_0-n_1$ leaves $\|\bb(\btheta^*)\|=\Theta((\log n_1)^{-1/2})$, yielding a slower rate
$\cR(\hat{\btheta})-\cR(\btheta^*)=\Theta_P((\log n_1)^{-1})$.
\end{example}

Example \ref{ex:collinear-log} highlights two points. First, while $\tilde n\approx n_0-n_1$ is necessary for consistency, the lower-order refinement of $\tilde n$ around $n_0-n_1$ can change the convergence rate. Second, the relative direction of $\nabla\psi(\btheta^*)$ and $\nabla\phi(\btheta^*)$ matters: alignment can either amplify or cancel the leading bias term $\bb(\btheta^*)$. We next formalize these insights in a general setting.

\begin{theorem}[Excess risk under a realistic synthetic generator]\label{rate-improve-angle}
Suppose the assumptions of Theorem~\ref{representation} hold and $\|\nabla^2\psi(\btheta^*)\|<\infty$.
Assume the generator is realistic in the sense that $\|\nabla\psi(\btheta^*)\|\gg n_0^{-1/2}$, and that the alignment condition
\(
\sin\angle \left(\nabla\phi(\btheta^*),\nabla\psi(\btheta^*)\right)\lesssim \frac{1}{\|\nabla\psi(\btheta^*)\|\sqrt{n_0}}
\)
holds. Then,  if we choose
    \(
    \tilde{n}= \frac{n_0-n_1+O_P(\sqrt{n_0})}{1-2\|\nabla\psi(\btheta^*)\|/\|\nabla\phi(\btheta^*)\|\cos\angle \left(\nabla\phi(\btheta^*),\nabla\psi(\btheta^*)\right)+O_P(1/\sqrt{n_0})},
    \)
    we have $\cR(\hat{\btheta})-\cR(\btheta^*)=\Theta_P(n_0^{-1})$. If we choose the naive balancing rule $\tilde{n}=n_0-n_1$, then
    $\cR(\hat{\btheta})-\cR(\btheta^*)=\Theta_P(\|\nabla\psi(\btheta^*)\|^2)\gg \Theta_P(n_0^{-1})$.
\end{theorem}

Theorem~\ref{rate-improve-angle} formalizes a ``bias-cancellation'' phenomenon. When the generator mismatch $\nabla\psi(\btheta^*)$ is sufficiently aligned with the intrinsic shift $\nabla\phi(\btheta^*)$, a small refinement of $\tilde n$ around $n_0-n_1$ cancels the leading distribution bias and restores the parametric rate $\Theta_P(n_0^{-1})$. In contrast, naive balancing $\tilde n=n_0-n_1$ leaves a residual bias of order $\|\nabla\psi(\btheta^*)\|$, resulting in a slower rate.
Practically, the result suggests treating $\tilde n$ as a tuning parameter rather than a fixed balancing heuristic: useful directional information between $\nabla\phi(\btheta^*)$ and $\nabla\psi(\btheta^*)$ can guide a choice of $\tilde n$ that reduces generator-induced bias and improve the performance.

\subsection{With Inconsistent Synthetic Sample Generators}\label{sec:inconsistent}

In this subsection, we aim to study the inconsistent regime as a robustness and failure-mode analysis. In practice, generators may be misspecified, capacity-limited, or trained with insufficient data, leading to a non-vanishing distributional mismatch even asymptotically. In this case, we consider the setting in which the mismatch does not vanish, e.g.,
\(
\liminf_{n_0,n_1\to\infty}\|\nabla\psi(\btheta^*)\|\gtrsim 1,
\)
where $\psi(\cdot)$ is defined in equation (\ref{eq:psi}). This implies that the total synthetic bias term $\bb(\btheta^*)$ defined in Theorem \ref{representation} generally does not converge to zero, so the estimator may retain a non-vanishing bias unless additional directional structure between $\nabla\phi(\btheta^*)$ and $\nabla\psi(\btheta^*)$ is available and the synthetic size $\tilde n$ is tuned accordingly. 

The next theorem shows that (a) if $\bb(\btheta^*)$ remains bounded away from zero, then the excess risk cannot converge to zero; (b) when $\nabla\phi(\btheta^*)$ and $\nabla\psi(\btheta^*)$ are approximately aligned, an appropriate choice of $\tilde n$ can partially offset the persistent mismatch and restore consistency.
\begin{theorem}[Excess risk under an inconsistent synthetic generator]\label{inconsistent}
    Suppose synthetic bias satisfies $\liminf_{n_0,n_1\to\infty} \left\|\nabla\psi(\btheta^*)\right\|\geq c$ for constant $c>0$, and $\|\nabla\psi(\btheta^*)\|/\|\nabla\phi(\btheta^*)\|<1/2$.
    \begin{itemize}[leftmargin=*]
        \item When synthetic direction is well aligned such that $\sin\angle\left(\nabla\phi(\btheta^*),\nabla\psi(\btheta^*)\right)=o_P(1)$, by selecting synthetic size 
    \(
    \tilde{n}=\frac{n_0-n_1+(n_0+n_1)o(1)}{1-2\|\nabla\psi(\btheta^*)\|/\|\nabla\phi(\btheta^*)\|\cos\angle\left(\nabla\phi(\btheta^*),\nabla\psi(\btheta^*)\right)+o(1)},
    \)
    the excess risk satisfies
    \(
    \cR(\hat{\btheta})-\cR(\btheta^*)=o_P(1).
    \)

       \item If either (i) the generator mismatch is not asymptotically aligned with the intrinsic shift, in the sense that there exists $c_0>0$ such that $\sin\angle\left(\nabla\phi(\btheta^*),\nabla\psi(\btheta^*)\right)>c_0>0$ in probability and $\tilde n/n_0\to \rho$ for some $\rho\geq 0$, or (ii) the directions are aligned but we nevertheless choose the naive balancing rule $\tilde n=n_0-n_1$, then there exists $c_1>0$ such that
       \(
    \cR(\hat{\btheta})-\cR(\btheta^*)\geq c_1+o_P(1)
    \).
    \end{itemize}
\end{theorem}

Next, we provide two examples that illustrate the phenomena characterized by Theorem \ref{inconsistent}. All derivations are given in Section~\ref{sec:biasedexampledetail} of the Supplementary Materials.

\begin{example}[Two-dimensional Gaussian model]\label{biasedexample}
Let $d=2$. Suppose
\(
\bx\mid y=0 \sim \cP_0=\cN(\mathbf{0},I_2),~
\bx\mid y=1 \sim \cP_1=\cN(\bmu_1,I_2),
\)
and synthetic minority samples are generated from $\tilde{\bx}\sim \cP_{\mathrm{syn}}=\cN(\bmu_s,I_2).$ We consider the squared loss $\ell(\btheta;\bx,y)=(\btheta^\top\bx-y)^2$.
\begin{enumerate}[label=(\roman*), leftmargin=*]
\item \emph{Aligned bias directions.}
Take $\bmu_1=(\mu,0)^\top$ and $\bmu_s=(a,0)^\top$ with $a\notin\{\mu,  2/\mu\}$ to avoid degeneracy. In this case, $\nabla\phi(\btheta^*)$ and $\nabla\psi(\btheta^*)$ are collinear. Consequently, there exists a choice of synthetic sample size,
\(
\tilde n=\frac{\mu(n_0-n_1)}{a(\mu^2+2)-\mu(a^2+1)},
\)
for which the leading bias cancels and the balanced excess risk satisfies
\(
\cR(\hat{\btheta})-\cR(\btheta^*) = O_P \left((n_0+n_1+\tilde n)^{-1}\right).
\)
However, the naive ``balancing'' choice $\tilde n=n_0-n_1$ need not be sufficient for consistency. For example, when $\mu=1$ and $a=\tfrac12$, the bias-cancelling choice equals $\tilde n=4(n_0-n_1)$, whereas using $\tilde n=n_0-n_1$ yields
\(
\cR(\hat{\btheta})-\cR(\btheta^*) \to c>0
\) in probability.

\item \emph{Orthogonal bias directions.}
Take $\bmu_1=(\mu,0)^\top$ and $\bmu_s=(\mu,\mu)^\top$ with $\mu^2\leq 2$. Then $\nabla\phi(\btheta^*)$ and $\nabla\psi(\btheta^*)$ are orthogonal. If the synthetic fraction does not vanish (i.e., $\liminf \tilde\pi\ge\tau>0$), the balanced excess risk remains bounded away from zero:
\(
\cR(\hat{\btheta})-\cR(\btheta^*) \to c>0
\) in probability.
Thus, when the synthetic mismatch introduces a bias component orthogonal to $\nabla\phi(\btheta^*)$, tuning $\tilde n$ cannot eliminate the leading bias and consistency fails whenever synthetic augmentation has non-negligible weight.
\end{enumerate}
\end{example}

\section{Local Symmetry: When Synthetic Minority Samples May Degrade Performance}\label{sec:synnohelp}

Synthetic minority augmentation is often motivated as a remedy for class imbalance. 
However, there is an important regime in which augmentation cannot help, and may even hurt, despite severe imbalance. The key phenomenon is a form of \emph{local symmetry} at the balanced-risk minimizer, under which the two classes already contribute equally along the optimization-relevant directions. In such cases, imbalance is not the bottleneck, and additional synthetic samples only introduce distributional bias if the generator is imperfect.

Recall we say the problem exhibits \emph{local symmetry} at $\btheta^*$ if
\begin{equation}\label{eq:localsym}
\nabla \phi(\btheta^*)
=
\nabla_{\btheta}\Big\{\bE_{\cP_0}\ell(\btheta;\bx,0)-\bE_{\cP_1}\ell(\btheta;\bx,1)\Big\}\Big|_{\btheta=\btheta^*}
=0.
\end{equation}
Condition \eqref{eq:localsym} means that, at the balanced-risk optimum, the (population) gradient contributions from the majority and minority classes coincide; hence reweighting the classes provides little first-order benefit near $\btheta^*$. In this regime, the leading bias term that governs the excess risk reduces to the generator-mismatch component. Specifically, writing the (vector) bias as
\(
\bb(\btheta^*):=\Big(\pi_0-\tfrac12\Big)\nabla\phi(\btheta^*)+\tilde{\pi}\nabla\psi(\btheta^*),
\)
local symmetry implies $\bb(\btheta^*)=\tilde{\pi}\nabla\psi(\btheta^*)$. Thus, setting $\tilde n=0$ eliminates the synthetic component entirely, whereas adding many synthetic samples can amplify the mismatch bias through $\tilde\pi$. The next theorem quantifies this trade-off under a \emph{realistic} generator that is imperfect but improving with sample size.

\begin{theorem}[Realistic synthetic augmentation can degrade performance]\label{thm:cannothelp}
Under the assumptions of Theorem~\ref{representation}, suppose the local symmetry condition \eqref{eq:localsym} holds, and the synthetic generator is \emph{realistic} in the sense that
\(
n_0^{-1/2}\ll \|\nabla\psi(\btheta^*)\|\ll 1.
\)
Then for diminishing synthetic size $\tilde n=O \big(\sqrt{n_0}/\|\nabla\psi(\btheta^*)\|\big)\ll n_0$, we have
\(
\cR(\hat{\btheta})-\cR(\btheta^*)=O_P(n_0^{-1}).
\)
For non-diminishing synthetic size $\tilde n\gtrsim n_0$, we have
\(
\cR(\hat{\btheta})-\cR(\btheta^*)=\Omega_P \big(\|\nabla\psi(\btheta^*)\|^2\big)\gg \Theta_P(n_0^{-1}).
\)
\end{theorem}

Theorem~\ref{thm:cannothelp} shows that, under local symmetry, synthetic augmentation cannot improve the balanced excess risk when the generator mismatch is non-negligible. If the synthetic size is diminishing, the estimator already achieves the baseline $O_P(n_0^{-1})$ rate, indicating that little or no augmentation is sufficient. In contrast, if the synthetic size is non-diminishing (of order $n_0$ or larger), the mismatch bias accumulates and the excess risk scales at least as $\|\nabla\psi(\btheta^*)\|^2$, which dominates the baseline $n_0^{-1}$ rate. Intuitively, because the two classes already exert equal first-order influence near $\btheta^*$, adding synthetic minority provides no imbalance-related benefit and can only degrade performance by injecting distributional bias.

We next present two examples satisfying the local symmetry condition \eqref{eq:localsym}, with details in Sections~\ref{meanshiftmodel} and~\ref{sec:sigmoidcancel}. Section~\ref{sec:simulation} further corroborates the theory via simulation. 

\begin{example}[Mean-shift model]\label{exa:meanshiftcancel}
Consider a mean-shift model with $\bP(y=1)=\pi$ and $\bP(y=0)=1-\pi$, and class-conditionals
\(
\bx \mid (y=1)\sim \bmu+\bxi,~
\bx \mid (y=0)\sim -\bmu+\bxi,
\)
where $\bmu\in\mathbb{R}^d$ is fixed and $\bxi\in\mathbb{R}^d$ has mean zero, bounded support, and full-rank covariance. Under the linear score $f_{\btheta}(\bx)=\btheta^\top\bx$ and squared loss $\ell(\btheta;\bx,y)=(y-\tfrac12-\btheta^\top\bx)^2$, the local symmetry condition \eqref{eq:localsym} holds, i.e., $\nabla\phi(\btheta^*)=\mathbf{0}$.
\end{example}

\begin{example}[Sigmoid Bernoulli logistic regression]\label{exa:sigmoidbernoullicancel}
Consider the sigmoid Bernoulli model (without intercept) with $\bx\sim p(\bx)$ and
\(
\bP(y=1\mid \bx)=\sigma(\bx^\top\btheta_{\mathrm{true}}),~
\bP(y=0\mid \bx)=1-\sigma(\bx^\top\btheta_{\mathrm{true}}),
\)
and logistic loss $\ell(\btheta;\bx,y)=\log \big(1+\exp\{\btheta^\top \bx\}\big)-y  \bx^\top\btheta$.
If
\(
\bE_{p(\bx)} \Big[\bx  \sigma(\btheta_{\mathrm{true}}^\top \bx)\big(1-\sigma(\btheta_{\mathrm{true}}^\top \bx)\big)\Big]=\mathbf{0},
\)
then the local symmetry condition \eqref{eq:localsym} holds, i.e., $\nabla\phi(\btheta^*)=\mathbf{0}$.
\end{example}

\section{Validation-Tuned Synthetic Size (VTSS)}\label{sec:validation-selection}

Our theory shows that the balancing rule $\tilde n = n_0-n_1$ is not always optimal. Theorem~\ref{representation} characterizes the leading excess risk as a function of $\tilde n$, and different choices of $\tilde n$ (even all of the same order as $n_0-n_1$) can yield materially different performance. While Theorem~\ref{representation} also suggests an ``optimal'' $\tilde n$, the criterion depends on unobserved population quantities, making direct optimization impractical. Nevertheless, the theorem provides a clear operational message: \emph{the choice of $\tilde n$ is important}, and naive balancing may be suboptimal. Motivated by the fact that, in locally asymmetric regimes, consistency typically requires $\tilde n$ to be close to $n_0-n_1$ (equivalently, $\tilde n/(n_0-n_1)\to 1$), we restrict attention to a neighborhood of the balancing choice. Specifically, we consider candidates
\(
\tilde n(\gamma)=\mathrm{round}\big(\gamma(n_0-n_1)\big), ~ \gamma\in\Gamma,
\)
where $\Gamma\subset\mathbb{R}_+$ is centered at $1$. For each $\gamma$, we augment the training set to achieve a target minority count $n_1^{\mathrm{tar}}=n_1+\tilde n(\gamma)$, train the downstream classifier, and select the $\gamma$ that minimizes the average \emph{balanced} validation loss in a $K$-fold cross-validation. We refer to this procedure as \emph{Validation-Tuned Synthetic Size (VTSS)} and summarize it in Algorithm~\ref{alg:vtss}.

\begin{algorithm}[]
\caption{Validation-Tuned Synthetic Size (VTSS)}
\label{alg:vtss}
\begin{algorithmic}[1]
\REQUIRE Imbalanced dataset $\mathcal D=\{(\boldsymbol{x}_i,y_i)\}_{i=1}^{n_0+n_1}$ with class counts $(n_0,n_1)$; number of folds $K$; candidate multipliers $\Gamma\subset\mathbb R_+$ (centered at $1$); classifier training routine $\mathrm{Fit}(\cdot)$; balanced log-loss $\hat L_{\mathrm{bal}}(\btheta;\mathcal D_{\mathrm{val}})=\frac{1}{2n_{0,\mathrm{val}}}\sum_{i=1}^{n_{0,\mathrm{val}}}\ell(\btheta; \bx_i^{0,\mathrm{val}},0)+\frac{1}{2n_{1,\mathrm{val}}}\sum_{i=1}^{n_{1,\mathrm{val}}}\ell(\btheta; \bx_i^{1,\mathrm{val}},1)$
synthetic generator $\mathrm{Syn}(\cdot; n_1^{\mathrm{tar}})$.
\ENSURE Selected synthetic size $\tilde n^\mathrm{VTSS}$ (equivalently $\gamma^\mathrm{VTSS}$) and $\hat \btheta^{\mathrm{VTSS}}$.

\STATE Partition $\mathcal D$ into $K$ folds $\{\mathcal D^{(k)}\}_{k=1}^K$. Initialize best CV loss $L^\mathrm{VTSS} \leftarrow +\infty$.
\FOR{each $\gamma \in \Gamma$}
    \STATE Compute candidate synthetic size $\tilde n \leftarrow \mathrm{round}(\gamma (n_0-n_1))$.
    \STATE Set target minority count $n_1^{\mathrm{tar}} \leftarrow n_1 + \tilde n$. Initialize CV loss accumulator $S_\gamma \leftarrow 0$.
    \FOR{$k=1,\dots,K$}
        \STATE Set validation fold $\mathcal D_{\mathrm{val}}^{(k)} \leftarrow \mathcal D^{(k)}$ and training folds $\mathcal D_{\mathrm{tr}}^{(k)} \leftarrow \mathcal D \setminus \mathcal D^{(k)}$.
        \STATE Augment training data:
        $\widetilde{\mathcal D}_{\mathrm{tr}}^{(k)}(\gamma) \leftarrow \mathrm{Syn}(\mathcal D_{\mathrm{tr}}^{(k)}; n_1^{\mathrm{tar}})$.
        \STATE Train classifier $\hat \btheta_{\gamma}^{(k)} \leftarrow \mathrm{Fit}(\widetilde{\mathcal D}_{\mathrm{tr}}^{(k)}(\gamma))$.
        \STATE Compute balanced validation log-loss
        $L_{\gamma}^{(k)} \leftarrow \widehat L_{\mathrm{bal}}(\hat \btheta_{\gamma}^{(k)};\mathcal D_{\mathrm{val}}^{(k)})$.
        \STATE Update $S_\gamma \leftarrow S_\gamma + L_{\gamma}^{(k)}$.
    \ENDFOR
    \STATE Compute average CV loss $\bar L_\gamma \leftarrow \frac{1}{K} S_\gamma$.
    \IF{$\bar L_\gamma < L^\mathrm{VTSS}$}
        \STATE Update $L^\mathrm{VTSS} \leftarrow \bar L_\gamma$, $\gamma^{\mathrm{VTSS}} \leftarrow \gamma$.
    \ENDIF
\ENDFOR
\STATE Set $\tilde n^{\mathrm{VTSS}} \leftarrow \mathrm{round}(\gamma^\mathrm{VTSS}(n_0-n_1))$ and $n_1^{\mathrm{tar}} \leftarrow n_1+\tilde n^{\mathrm{VTSS}}$.
\STATE Train final model on the full augmented data:
$\widetilde{\mathcal D}(\gamma^\mathrm{VTSS}) \leftarrow \mathrm{Syn}(\mathcal D; n_1^{\mathrm{tar}})$ and
$\hat \btheta^{\mathrm{VTSS}} \leftarrow \mathrm{Fit}(\widetilde{\mathcal D}(\gamma^\mathrm{VTSS}))$.
\STATE Output $\tilde n^{\mathrm{VTSS}}$ and $\hat \btheta^{\mathrm{VTSS}}$.
\end{algorithmic}
\end{algorithm}

\begin{remark}\label{rmk:algorithm}
    Algorithm~\ref{alg:vtss} uses $K$-fold cross-validation to select the synthetic multiplier $\gamma$ by minimizing the average validation risk. In settings where the validation criterion is noisy, the selection can be further stabilized by repeating the entire $K$-fold procedure $m$ times with independent fold partitions and choosing $\gamma$ that minimizes the aggregated mean over all $mK$ fold evaluations. Moreover, while we present VTSS using the balanced log-loss, the same framework applies to any problem-relevant validation objective. One may replace $\hat L_{\mathrm{bal}}$ with an alternative  metric (e.g., maximize balanced accuracy, minimize a cost-sensitive loss etc.) depending on the downstream goal. Thus, VTSS should be viewed as a general tuning framework that selects $\gamma$ by optimizing a user-specified, task-dependent validation criterion.
\end{remark}


\section{Simulation Studies}\label{sec:simulation}

We conduct simulations to validate the theory and to evaluate VTSS (Section~\ref{sec:validation-selection}). Due to space constraints, we defer the detailed simulation settings to Section \ref{sec:simulation-detail} of the Supplement.

\subsection{Under Local Asymmetry}
In the local asymmetry setting, we examine classification with directional alignment. To illustrate the phenomenon induced by a biased synthetic generator (Section~\ref{sec:inconsistent}), we use the 2D Gaussian model from Example~\ref{biasedexample} with a linear predictor. We compare two fixed synthetic size rules, naive balancing, $\tilde n=n_0-n_1$, and the bias-canceling choice, $\tilde n=4(n_0-n_1)$, and evaluate both the parameter error $|\hat\btheta-\btheta^*|_2$ and the balanced excess risk.

\begin{figure*}[htb!]
  \centering

  \begin{subfigure}[htb!]{0.35\textwidth}
    \centering
    \includegraphics[width=\linewidth]{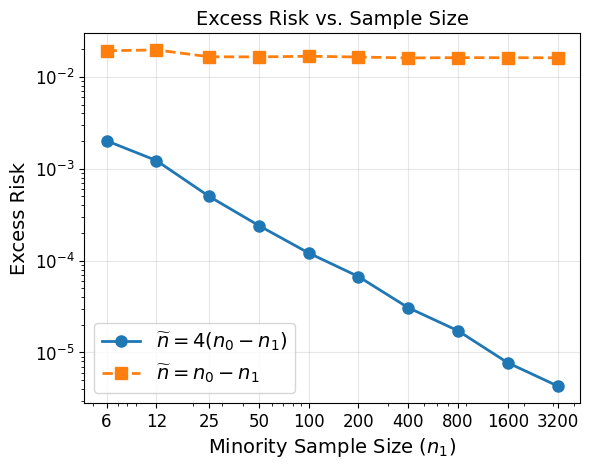}
  \end{subfigure}
  \begin{subfigure}[htb!]{0.35\textwidth}
    \centering
    \includegraphics[width=\linewidth]{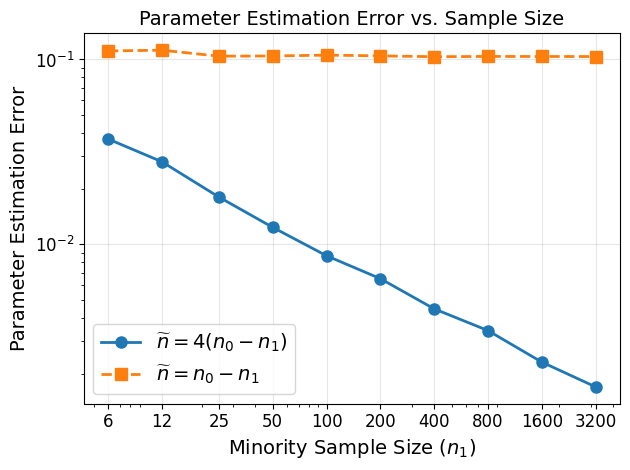}
  \end{subfigure}

  \caption{Balanced excess risk and parameter estimation error in a 2D Gaussian model with aligned synthetic bias based on $100$ simulations.}
  \label{fig:bias}
\end{figure*}

The simulation in Figure~\ref{fig:bias} confirms the analysis in Example~\ref{biasedexample}. When $\tilde n=4(n_0-n_1)$, both the balanced excess risk and the parameter error $\|\hat\btheta-\btheta^*\|_2$ decay to zero, whereas under naive balancing they converge to a strictly positive level. This demonstrates that naive balance can be inconsistent in the presence of biased synthetic data.

\begin{figure*}[htb!]
  \centering

  \begin{subfigure}[htb!]{0.4\textwidth}
    \centering
    \includegraphics[width=\linewidth]{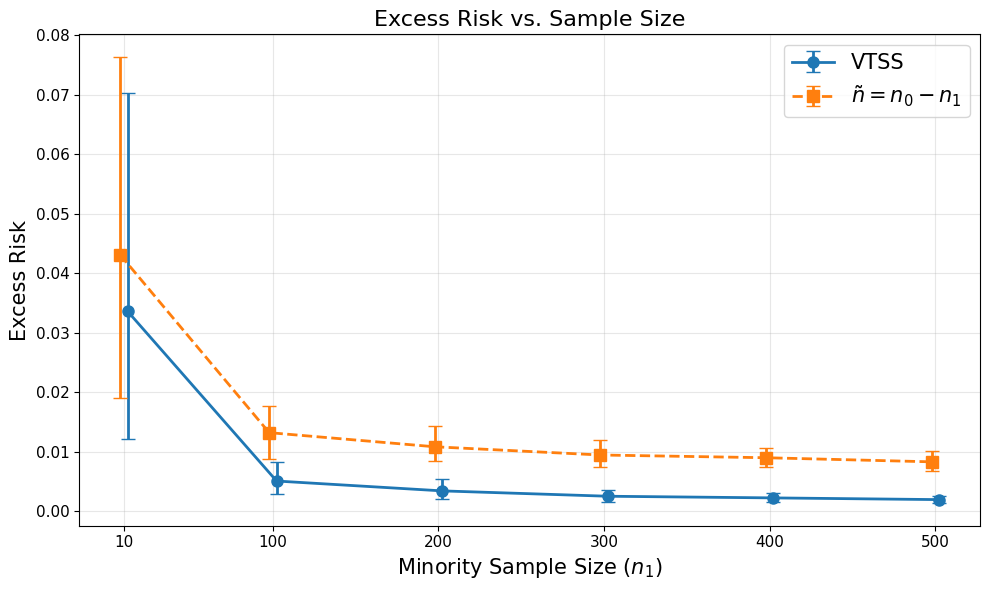}
  \end{subfigure}
  \begin{subfigure}[htb!]{0.4\textwidth}
    \centering
    \includegraphics[width=\linewidth]{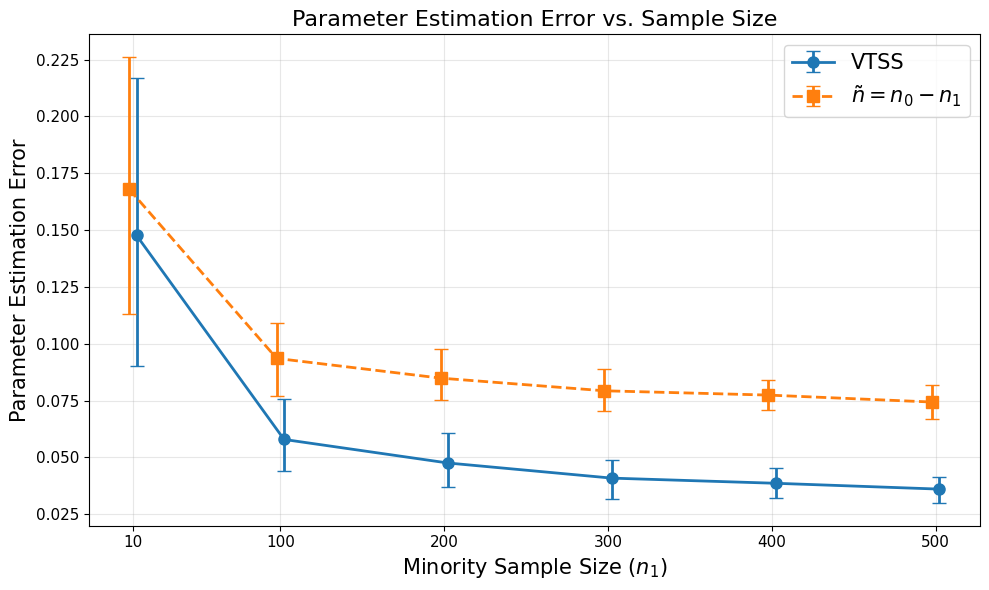}
  \end{subfigure}

  \caption{Excess risk and parameter error versus minority sample size under a 2D Gaussian mean-shift model (squared loss) based on $100$ simulations.}
  \label{fig:consistentalign}
\end{figure*}

For directional alignment with a realistic and consistent synthetic generator (Section~\ref{sec:rate-improve}), we simulate the same 2D Gaussian setup with squared loss.
Synthetic samples are produced by a realistic, consistent, and directionally aligned generator: for a requested synthetic size $k$, draw $\bx_{\mathrm{syn}}\sim \cN(\bmu_{\mathrm{syn}}, I_2)$ with
\(
\bmu_{\mathrm{syn}}=\left(1-(\log n_1)^{-1/2}\right)\bmu_1,
\)
so that the synthetic mean approaches $\bmu_1$ while remaining aligned.
We compare VTSS with the naive balancing rule $\tilde n=n_0-n_1$, and evaluate population balanced excess risk and parameter error.

Figure~\ref{fig:consistentalign} shows that, for all $n_1$, VTSS attains substantially smaller errors with a faster decay rate, supporting the theory that under a realistic yet consistent generator, directional information can be leveraged to choose $\tilde n$ and outperform naive balancing.

\subsection{Under Local Symmetry}

\begin{figure*}[htb!]
  \centering
  \includegraphics[width=0.9\linewidth]{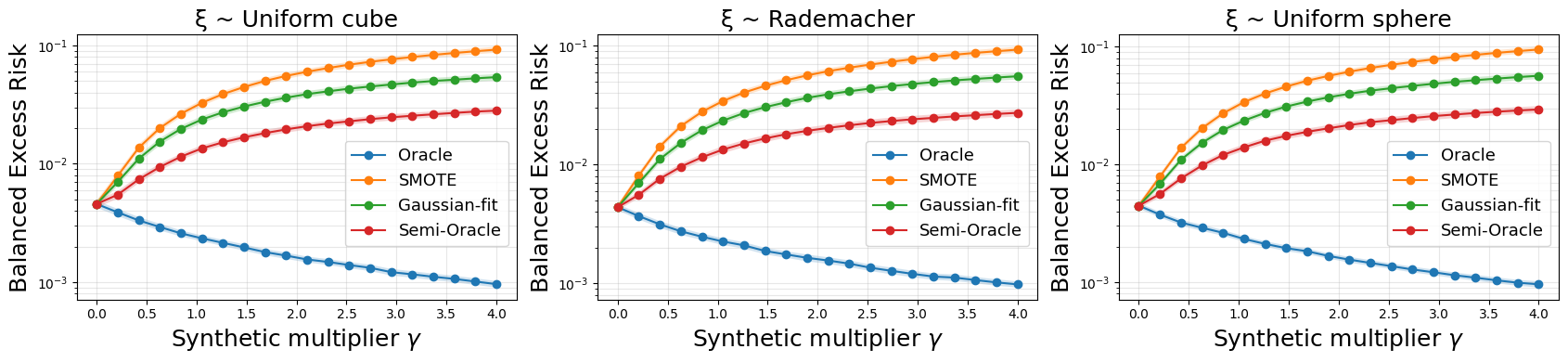}

  \caption{Balanced excess risk $\cR(\hat\btheta)-\cR(\btheta^*)$ (log scale) versus the synthetic multiplier $\gamma$.}
  \label{fig:nosyn-meanshift}
\end{figure*}

To further illustrate local symmetry (Section~\ref{sec:synnohelp}), we conduct a simulation based on the mean-shift model in Example~\ref{exa:meanshiftcancel} with a linear predictor. We consider three distributions of $\bxi$: Uniform cube, Rademacher, and Uniform sphere.
We start from imbalanced samples ($n_1=100$, $n_0=2000$), and augment the minority with $\tilde n=\mathrm{round} (\gamma (n_0-n_1))$ synthetic samples, where $\gamma$ ranges over 20 values in $[0,4]$.
We compare four synthetic generators: (i) \textbf{Oracle}; (ii) \textbf{SMOTE}; (iii) \textbf{Gaussian-fit}; and (iv) \textbf{Semi-Oracle}, which samples $\bxi$ from the true noise distribution but centers at the empirical minority mean $\hat\bmu_1$ (i.e., $\hat\bmu_1+\bxi$).

Figure~\ref{fig:nosyn-meanshift} aligns with the theory. In the mean-shift setting, the majority--minority difference cancels at the balanced optimum ($\nabla\phi(\btheta^*)=0$), so realistic generators cannot reduce this part of the error. Any mismatch between the synthetic and true minority distributions induces bias and can only increase balanced excess risk. Empirically, realistic generators, including the Semi-Oracle generator that is already close to the truth, do not improve performance and typically degrade it as $\gamma$ increases, whereas the Oracle improves monotonically by adding perfect minority samples. Consistent with this, the Supplementary Figure~\ref{fig:select0} shows VTSS selects very small $\gamma$ in most runs, often exactly $\gamma=0$, justifying its effectiveness.

We also have analogous local symmetry results showing that synthetic data provides no benefit for the sigmoid Bernoulli logistic model as in Example \ref{exa:sigmoidbernoullicancel}. These results are presented in Section \ref{sec:simu-symmetry-sigmoid} of the supplementary materials.

\subsection{Further Evaluation of VTSS}\label{sec:furtherevaluation}

\begin{figure*}[htb!]
  \centering

  \begin{subfigure}[t]{0.4\textwidth}
    \centering
    \includegraphics[width=\linewidth]{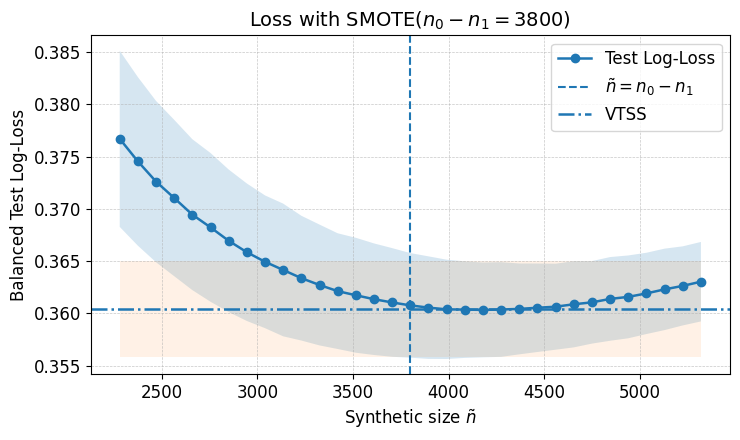}
  \end{subfigure}
  \begin{subfigure}[t]{0.4\textwidth}
    \centering
    \includegraphics[width=\linewidth]{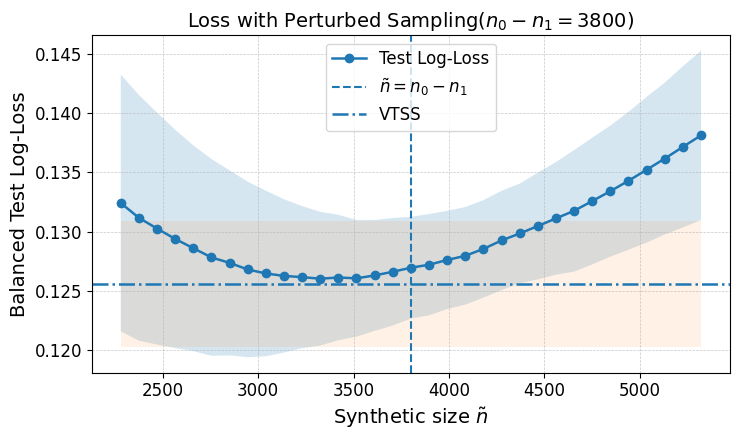}
  \end{subfigure}\hfill

  \caption{Balanced test log-loss as a function of the synthetic size $\tilde n$.
\textbf{Left:} logistic regression with SMOTE as the synthetic generator.
\textbf{Right:} kernel logistic regression with perturbed sampling as the synthetic generator.}
  \label{fig:curves}
\end{figure*}

In this subsection, we further illustrate the advantage of VTSS by plotting the test loss as a function of the synthetic size $\tilde n$ and comparing it with the loss achieved by VTSS. We consider two configurations. (i) Linear classification (SMOTE + logistic regression) and 
(ii) Nonlinear classification (perturbed sampling + kernel logistic).
In both configurations of Figure~\ref{fig:curves}, the empirical curves confirm the theoretical insight that, although the naive balancing choice $\tilde n=n_0-n_1$ often lies near the optimum, it is generally not the loss-minimizing synthetic size. Moreover, VTSS, which selects $\tilde n$ by minimizing balanced validation loss on each dataset, can always capture the lowest average loss. This highlights the benefit of data-dependent tuning around a theory-guided target.

\section{Real Data Applications}\label{sec:applications}

\paragraph*{Data source, cohort construction, features.} Clinically, sepsis is a life-threatening condition arising from a dysregulated response to infection that can rapidly progress to organ dysfunction, making timely identification essential \citep{singer2016third}. We conduct experiments on MIMIC-III (Medical Information Mart for Intensive Care), a large, de-identified, single-center ICU dataset covering admissions at Beth Israel Deaconess Medical Center from 2001–2012, with rich clinical information \citep{johnson2016mimic}.
Using MIMIC-III, we construct an adult admission cohort (age $>15$ years at admission). We consider three binary prediction tasks, with in-hospital mortality, sepsis, and septic shock as responses. These represent clinically consequential outcomes capturing (i) overall patient prognosis, (ii) infection-related critical illness, and (iii) progression to the most severe end of the sepsis spectrum. Sepsis status is identified using ICD-9 diagnosis codes, labeling admissions with severe sepsis or septic shock as sepsis cases.

To capture early physiologic trajectories, we extract 10 clinical variables during the first 120 hours after admission and use binned averages to stabilize irregularly sampled measurements:
\begin{itemize}
\item Bedside vital signs from \texttt{CHARTEVENTS}, averaged into 6-hour bins (20 bins): heart rate, respiratory rate, temperature, and systolic blood pressure.
\item Laboratory measurements from \texttt{LABEVENTS}, averaged into 24-hour bins (5 bins): white blood cell count, platelets, creatinine, INR, lactate, and total bilirubin.
\end{itemize}
For each admission, we convert the multivariate time series into a mean feature vector over the observation window. Missing features are imputed using the corresponding cohort-level feature means. After preprocessing, all three datasets are substantially imbalanced; the class counts and imbalance ratios are summarized in Table~\ref{tab:realdatasummary}.
\begin{table}[]
    \centering
    \resizebox{0.8\textwidth}{!}{
    \begin{tabular}{c|ccc}
         &  Positive Samples ($n_1$) & Negative Samples ($n_0$) & Imbalance Ratio ($n_0/n_1$)\\ \hline
       Mortality  & $5{,}789$ & $45{,}068$ & $7.785$ \\
       Sepsis & $4{,}084$ & $46{,}773$ & $11.45$\\
       Septic Shock & $2{,}585$ & $48{,}272$ & $18.67$
    \end{tabular}}
    \caption{Class imbalance for the mortality, sepsis, and septic shock prediction tasks.}
    \label{tab:realdatasummary}
\end{table}
We train (i) logistic regression and (ii) support vector machine (SVM) classifiers on these features to predict each outcome.

\paragraph*{Evaluation metric: balanced excess risk.}
Performance is primarily assessed using the balanced excess risk. Because the population minimizer $\theta^*=\arg\min_{\theta}\cR(\theta)$ depends on unknown population expectations, we approximate the subtractive term using an oracle plug-in estimate on each test split. Specifically, we compute $\theta^*_{\text{test}}=\arg\min_{\theta}\hat{\mathcal R}_{\text{test}}(\theta)$ (the parameter that minimizes the balanced loss on the test set) and report the balanced excess risk as $\hat{\cR}_{\text{test}}(\hat\theta)-\hat{\cR}_{\text{test}}(\theta^*_{\text{test}})$. We repeat the train-test split 100 times with different random seeds and report the average balanced excess risk across repetitions.

\paragraph*{Synthetic data generation and tuning protocol.} 
We compare four generators: SMOTE, ADASYN, Borderline-SMOTE, and Minority Jitter (randomly sampling a minority instance and adding feature-wise Gaussian perturbations). We tune synthetic multiplier $\gamma$ via 5-fold cross-validation and repeat the cross-validation procedure 10 times. We sweep $\gamma$ over a range centered around 1 and report the mean balanced excess risk with a 95\% confidence interval at each $\gamma$. For reference, we also plot a horizontal dashed line (with 95\% confidence interval) corresponding to the average balanced excess risk achieved by VTSS.

\begin{figure}[htbp]
  \centering

  \begin{subfigure}[t]{0.32\textwidth}
    \centering
    \includegraphics[width=\linewidth]{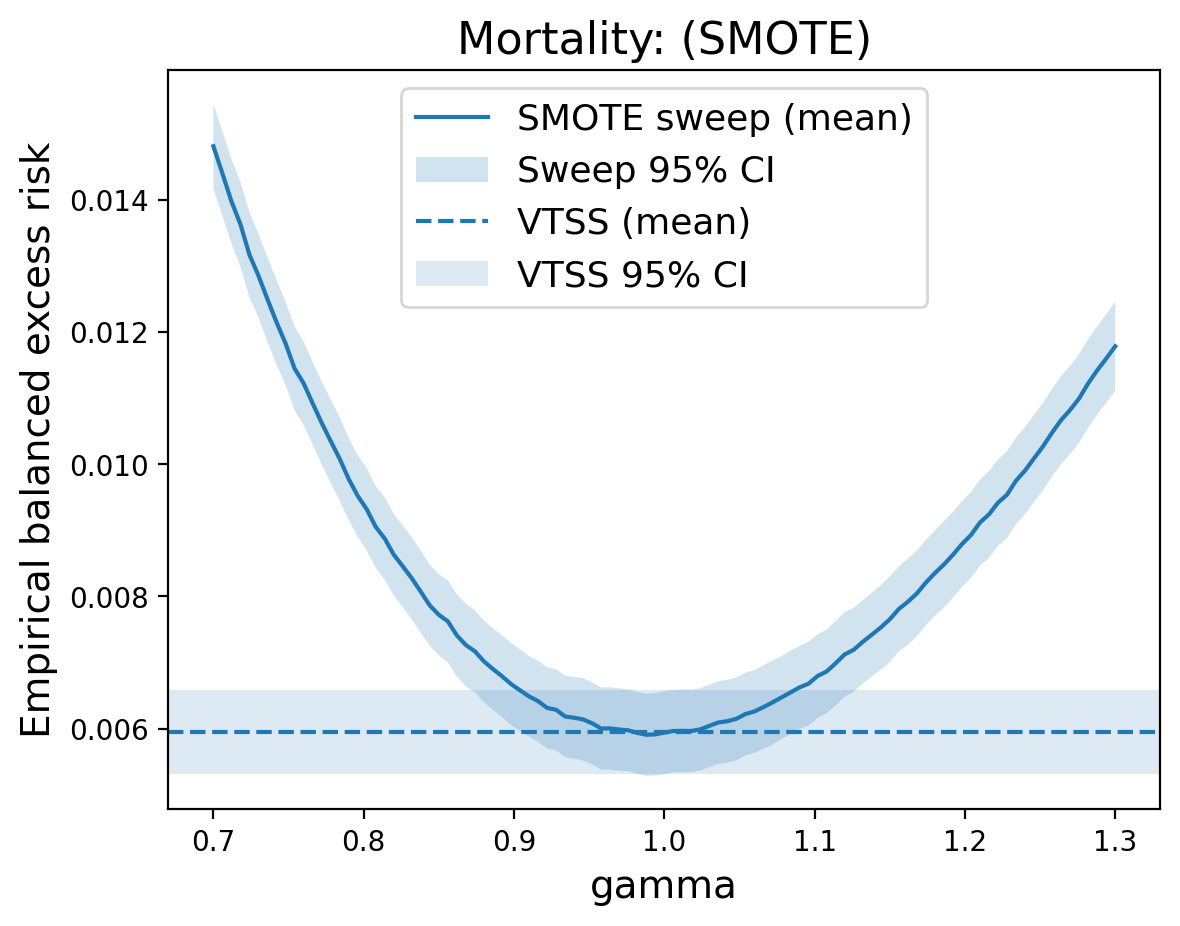}
  \end{subfigure}
  \begin{subfigure}[t]{0.32\textwidth}
    \centering
    \includegraphics[width=\linewidth]{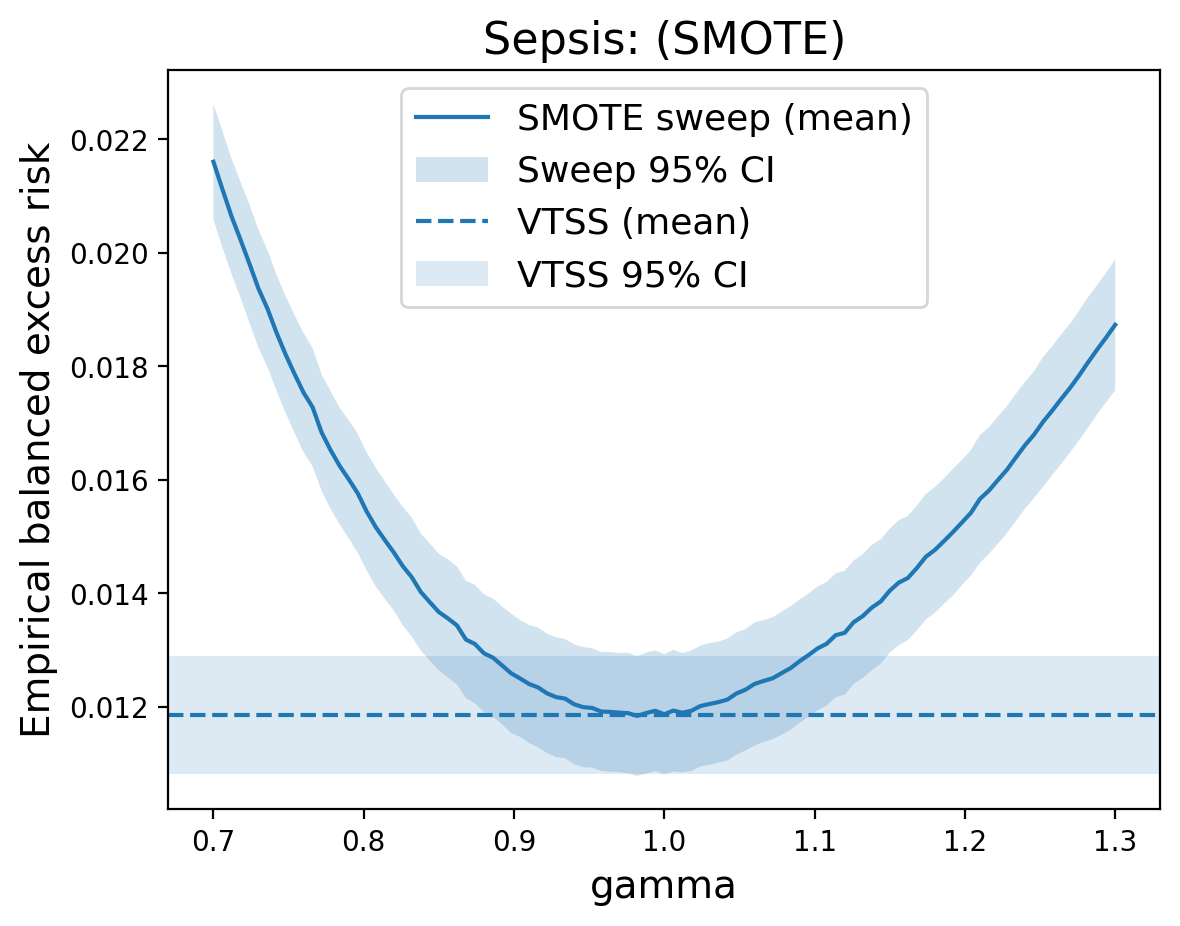}
  \end{subfigure}
  \begin{subfigure}[t]{0.32\textwidth}
    \centering
    \includegraphics[width=\linewidth]{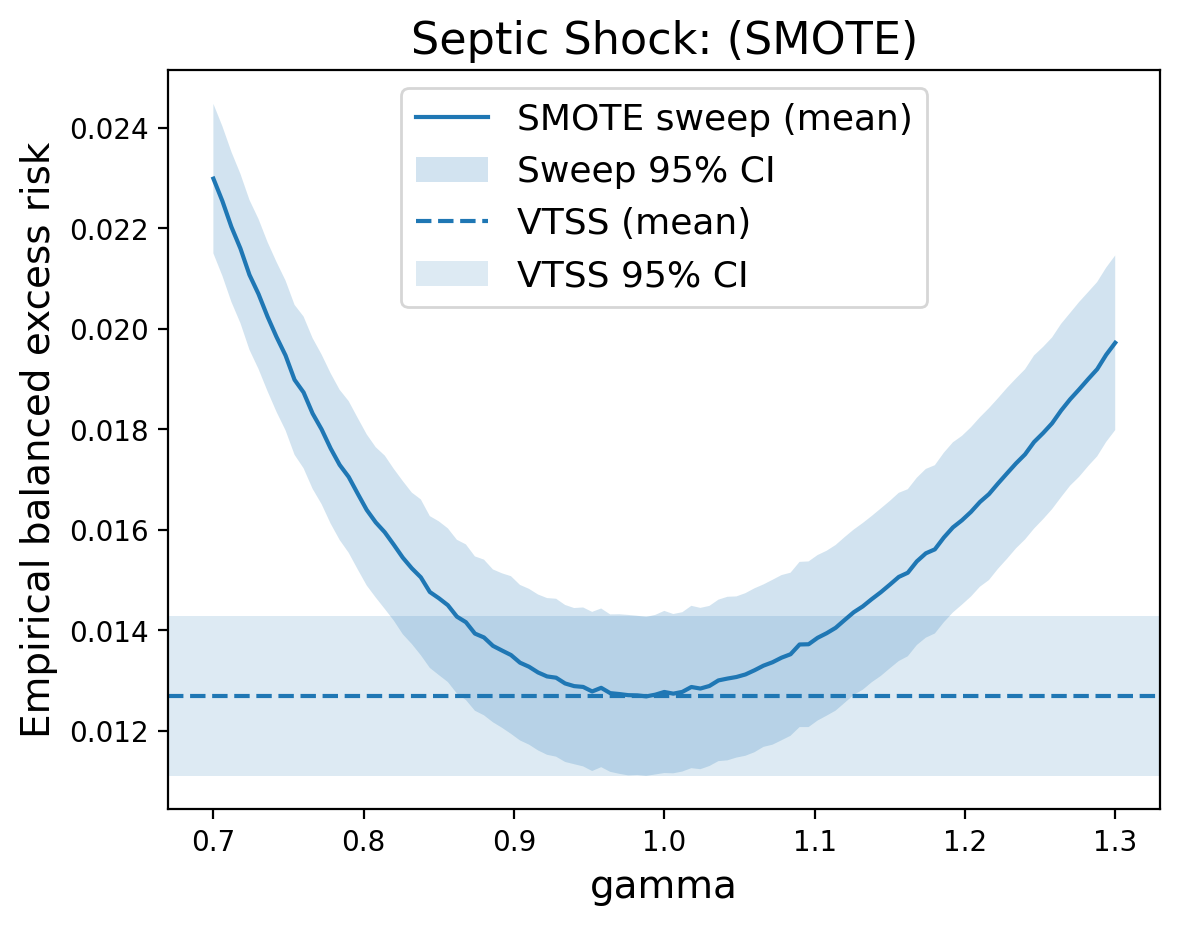}
  \end{subfigure}

  \begin{subfigure}[t]{0.32\textwidth}
    \centering
    \includegraphics[width=\linewidth]{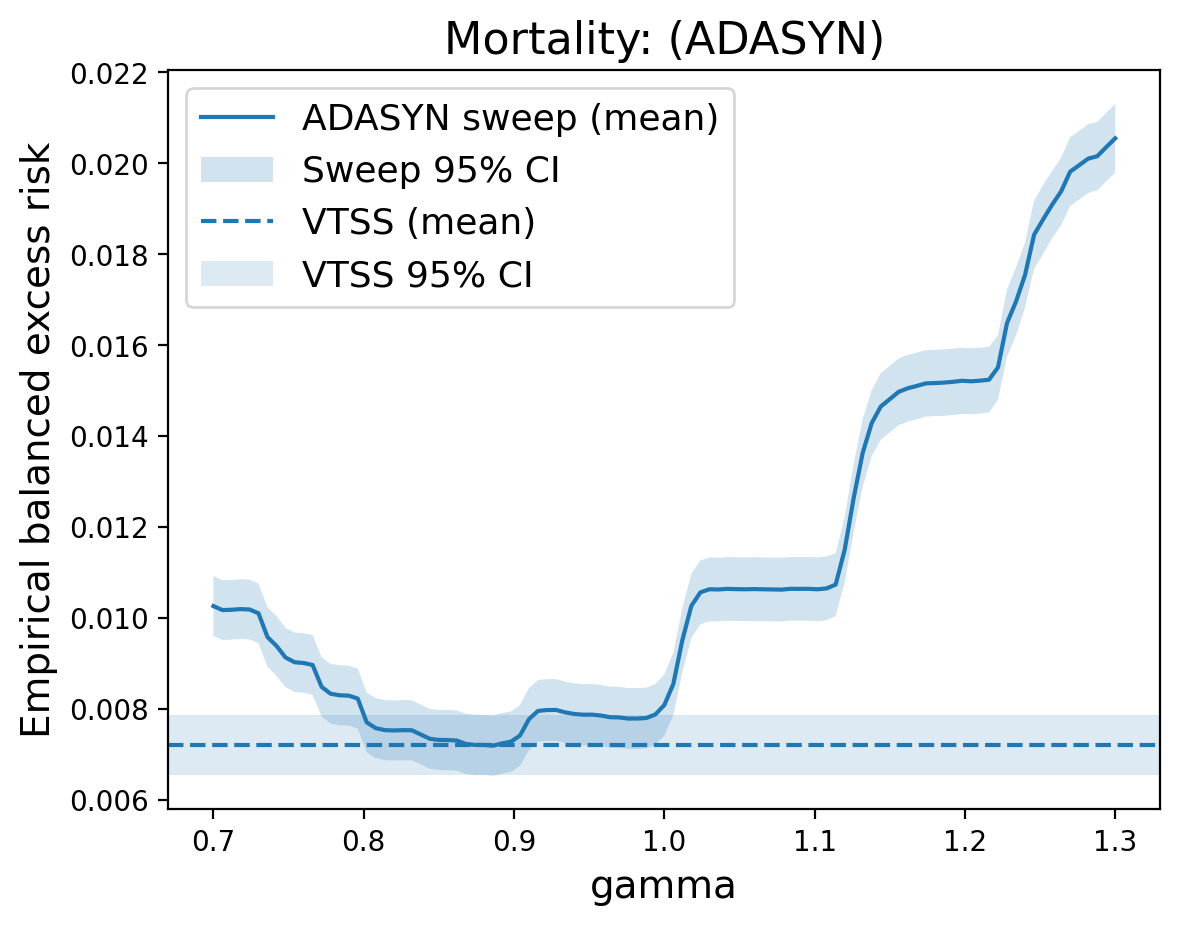}
  \end{subfigure}
  \begin{subfigure}[t]{0.32\textwidth}
    \centering
    \includegraphics[width=\linewidth]{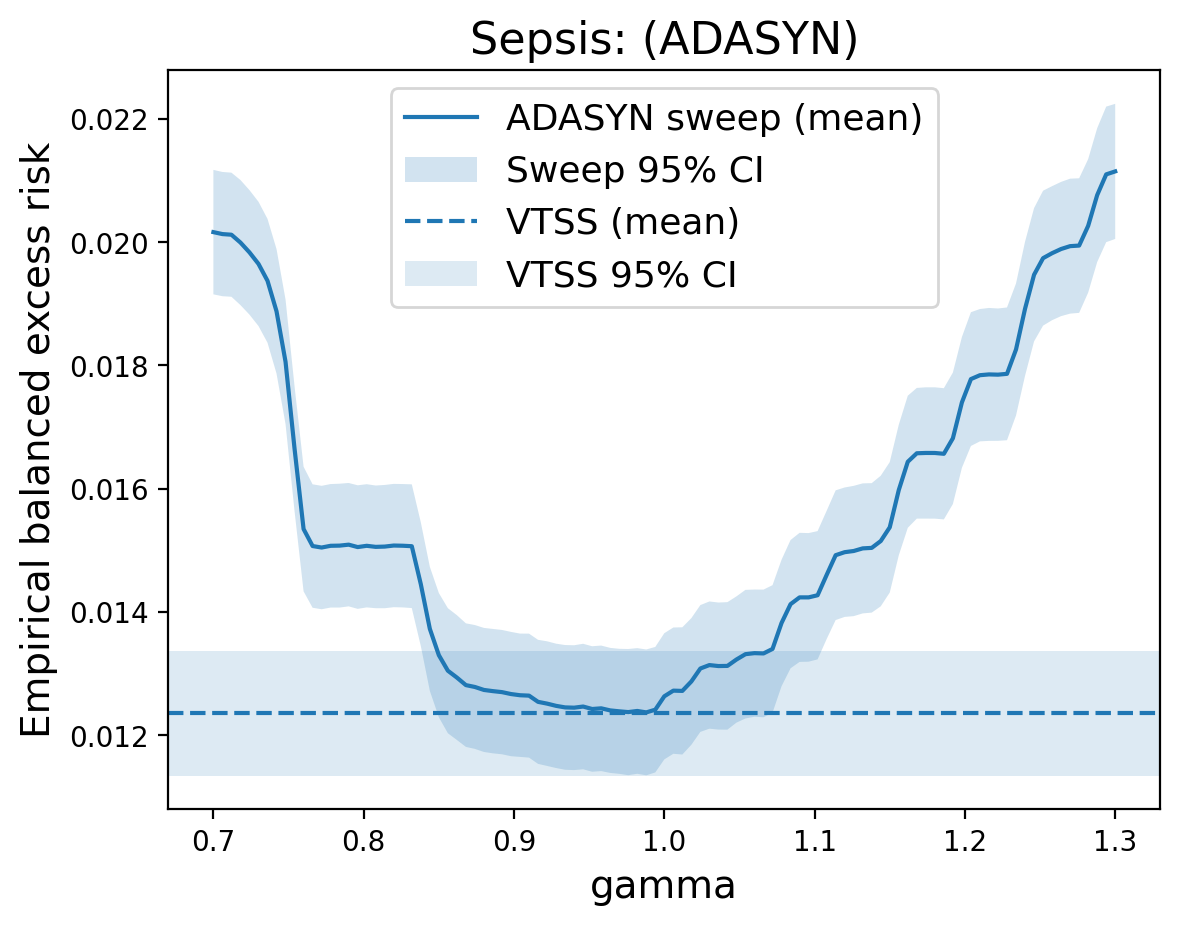}
  \end{subfigure}
  \begin{subfigure}[t]{0.32\textwidth}
    \centering
    \includegraphics[width=\linewidth]{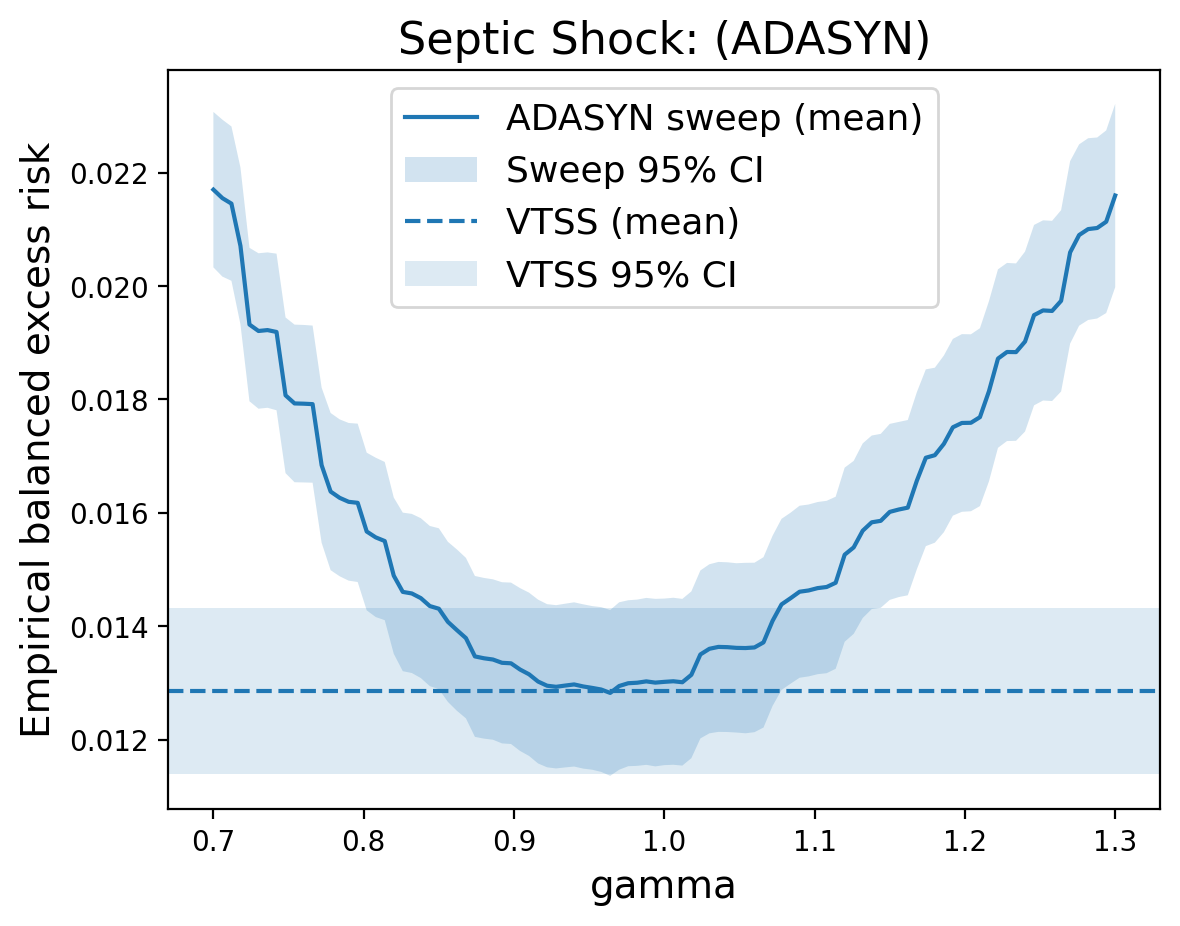}
  \end{subfigure}

  \begin{subfigure}[t]{0.32\textwidth}
    \centering
    \includegraphics[width=\linewidth]{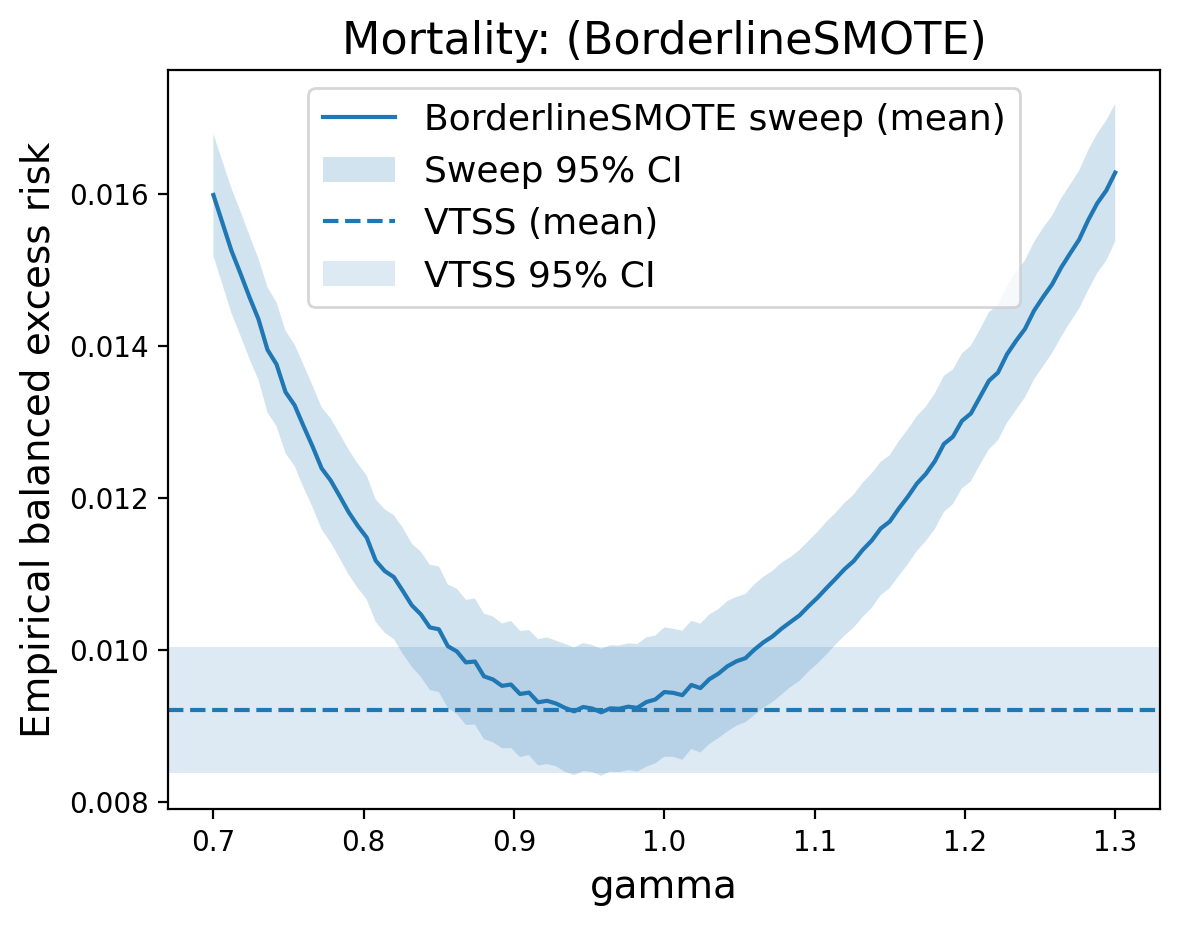}
  \end{subfigure}
  \begin{subfigure}[t]{0.32\textwidth}
    \centering
    \includegraphics[width=\linewidth]{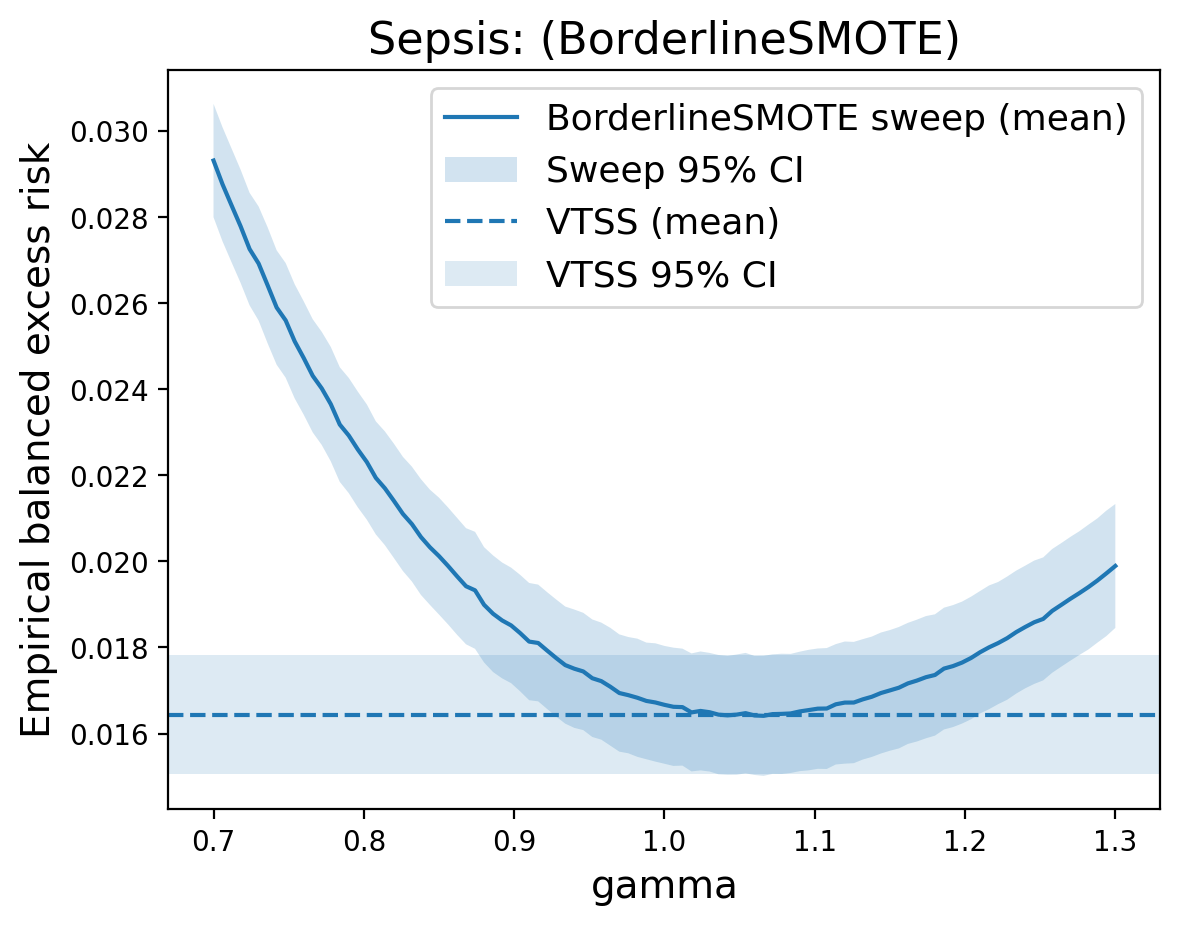}
  \end{subfigure}
   \begin{subfigure}[t]{0.32\textwidth}
    \centering
    \includegraphics[width=\linewidth]{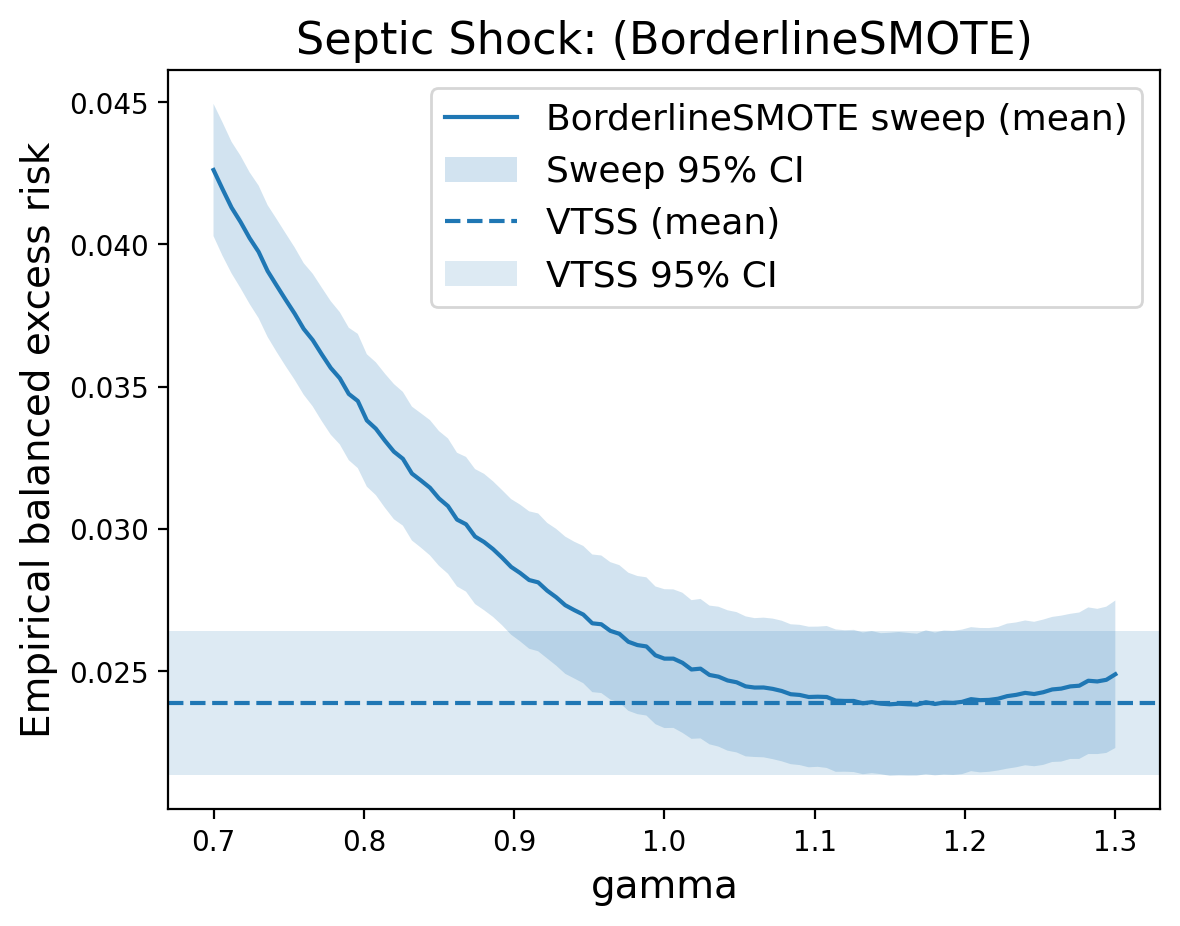}
  \end{subfigure}

  \begin{subfigure}[t]{0.32\textwidth}
    \centering
    \includegraphics[width=\linewidth]{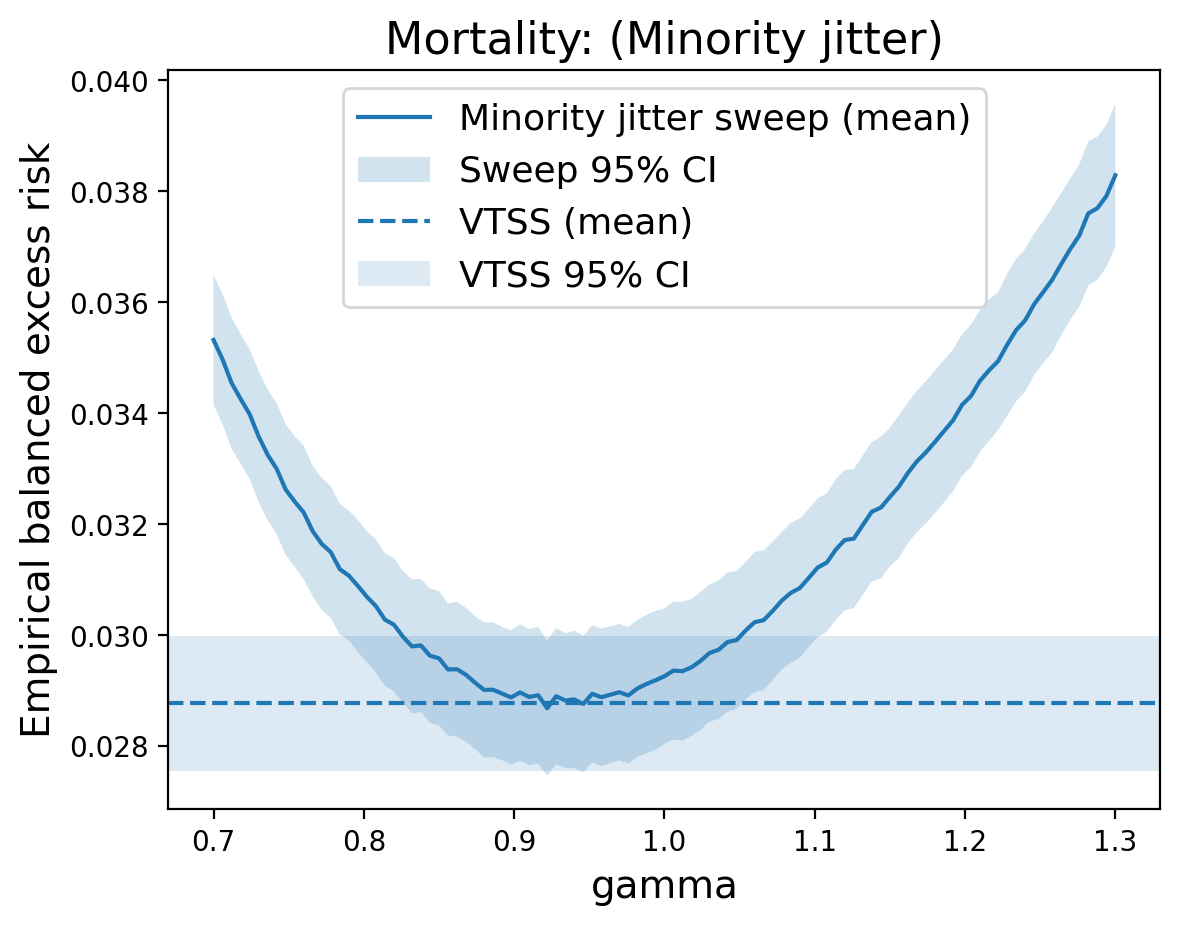}
  \end{subfigure}
  \begin{subfigure}[t]{0.32\textwidth}
    \centering
    \includegraphics[width=\linewidth]{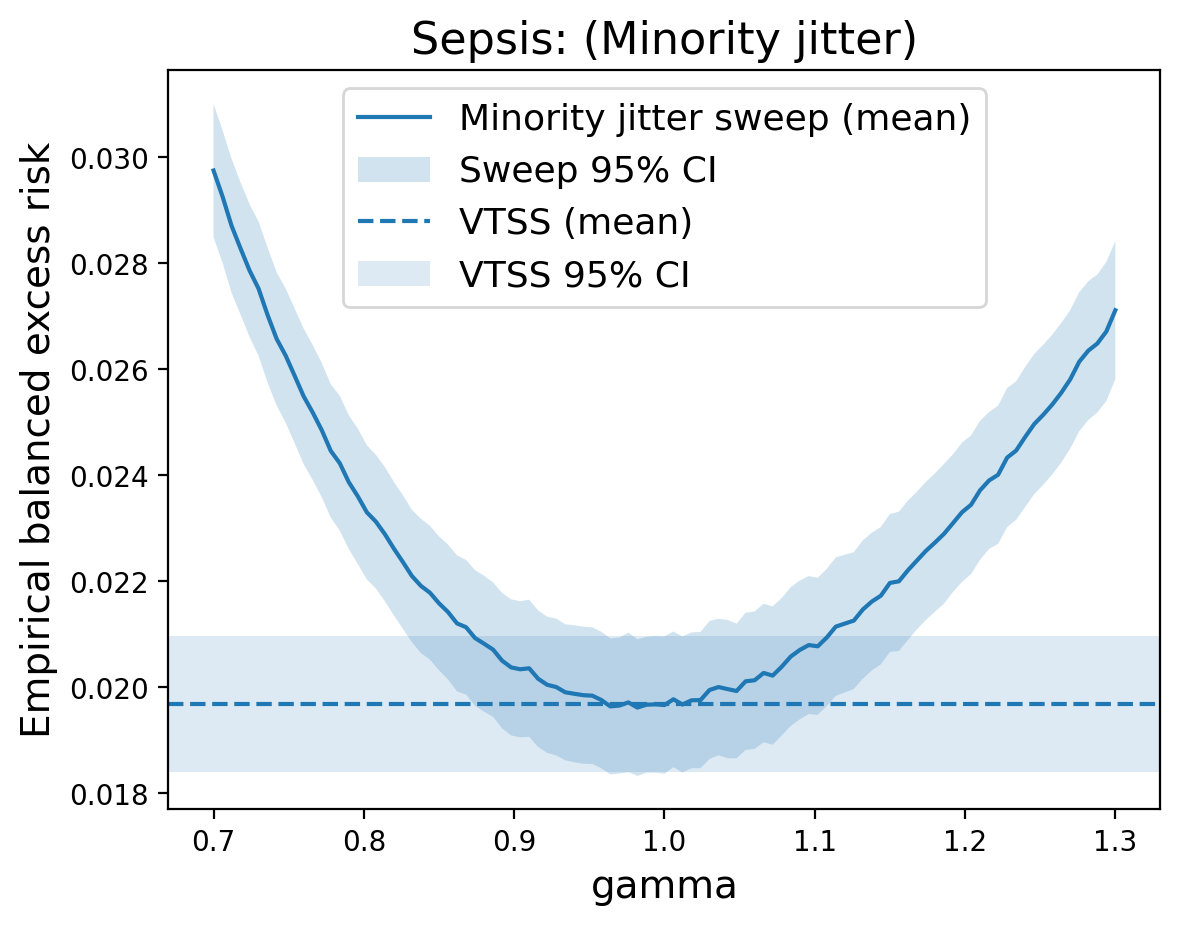}
  \end{subfigure}
  \begin{subfigure}[t]{0.32\textwidth}
    \centering
    \includegraphics[width=\linewidth]{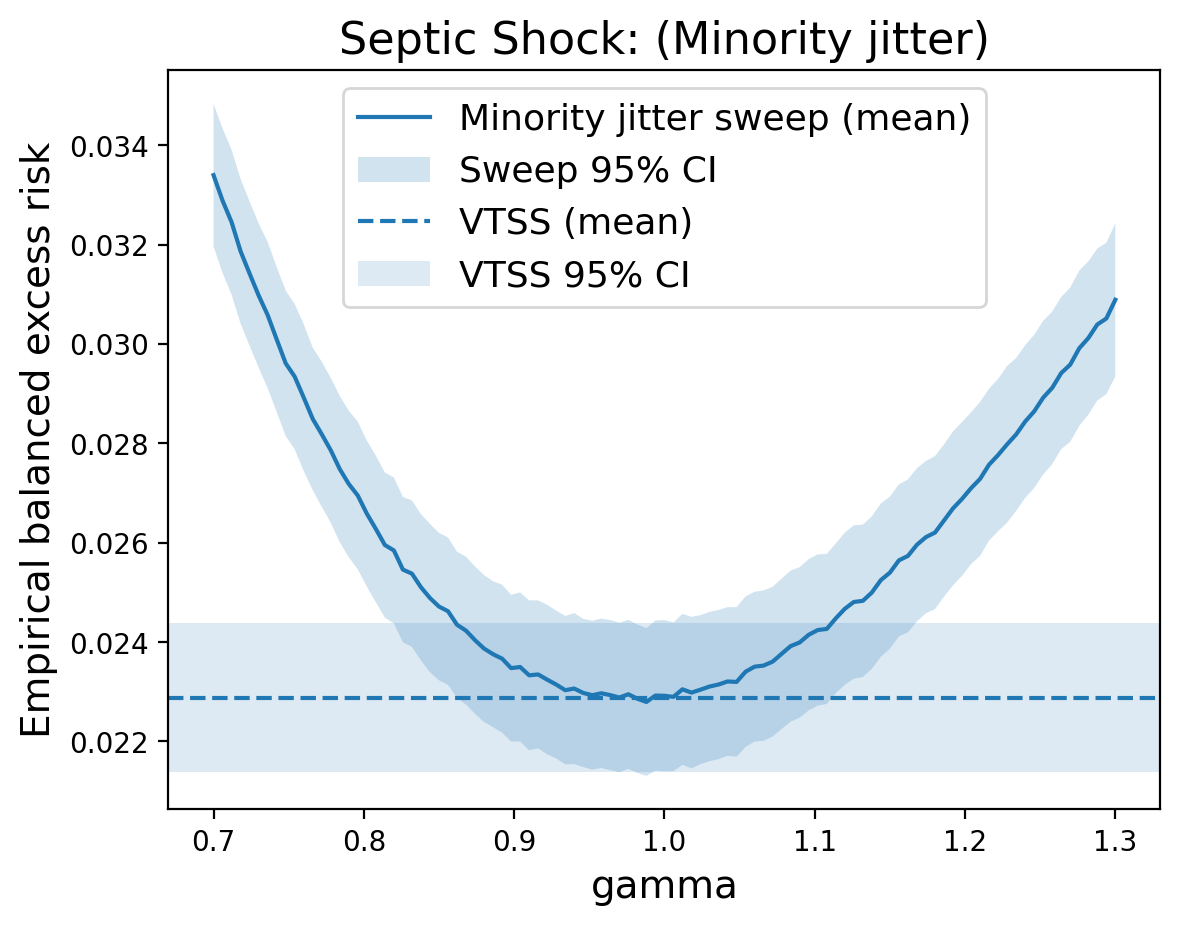}
  \end{subfigure}

  \par\medskip
\begin{minipage}[t]{0.32\textwidth}
  \centering
  \footnotesize
  \textbf{Raw-Mortality} \\
  $0.3825 ~[0.3809,0.3842]$
\end{minipage}\hfill
\begin{minipage}[t]{0.32\textwidth}
  \centering
  \footnotesize
  \textbf{Raw-Sepsis}\\
   $0.5721 ~[0.5700,0.5742]$
\end{minipage}\hfill
\begin{minipage}[t]{0.32\textwidth}
  \centering
  \footnotesize
  \textbf{Raw-Septic shock}\\
  $0.7786 ~[0.7761,0.7810]$
\end{minipage}

  \caption{Excess risk in predicting hospital mortality, sepsis and septic shock across four synthetic generators using logistic regression. Bottom row is the excess risk obtained by training with raw imbalanced data  without synthetic data augmentation.}
  \label{fig:realdataexcessrisk}
\end{figure}

\paragraph*{Results for logistic regression.} Figure \ref{fig:realdataexcessrisk} highlights that the optimal synthetic sample size is highly sensitive to both the clinical endpoint and the generator. Training on the raw imbalanced data ($\gamma=0$) is substantially worse (bottom row), underscoring the need to mitigate class imbalance. Across tasks, the balanced excess risk curves  exhibit minima at markedly different locations: for different tasks and generators, the best can be $\gamma^*<1$ (conservative oversampling), $\gamma^*\approx 1$, or $\gamma^*>1$ (aggressive oversampling). Overall, VTSS yields a stable selection that matches, within uncertainty, the best achievable risk over the candidate $\gamma$ values, supporting its use as a robust default across tasks and generators.

\paragraph*{Results for SVM.}
We repeat the same pipeline with SVM. Results are summarized in Figure~\ref{fig:realdataexcessrisksvm} of the Supplement. The qualitative conclusions remain consistent: performance is sensitive to $\gamma$, the optimal $\gamma$ depends on both the endpoint and generator, and training on the raw imbalanced data is frequently suboptimal. Notably, for in-hospital mortality the risk-minimizing multipliers are often concentrated around $\gamma\in[0.4,0.6]$, suggesting that under SVM this task favors substantially less oversampling than the naive balancing choice $\gamma=1$. VTSS captures this behavior and consistently recovers near-minimal balanced excess risk, supporting its use as a robust default across classifiers, tasks, and generators.

\begin{figure}[t]
  \centering

  \begin{subfigure}[t]{0.244\textwidth}
    \centering
    \includegraphics[width=\linewidth]{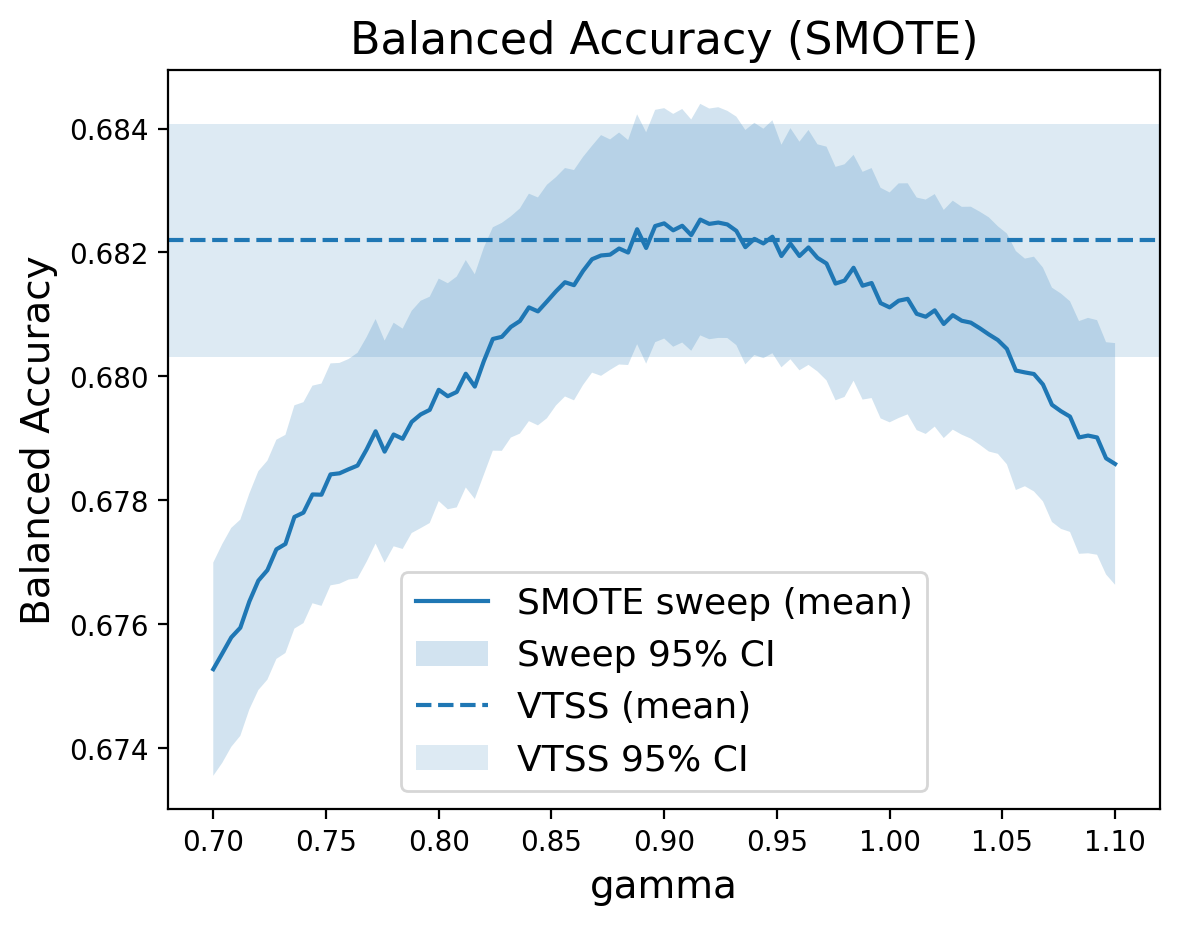}
  \end{subfigure}
  \begin{subfigure}[t]{0.244\textwidth}
    \centering
    \includegraphics[width=\linewidth]{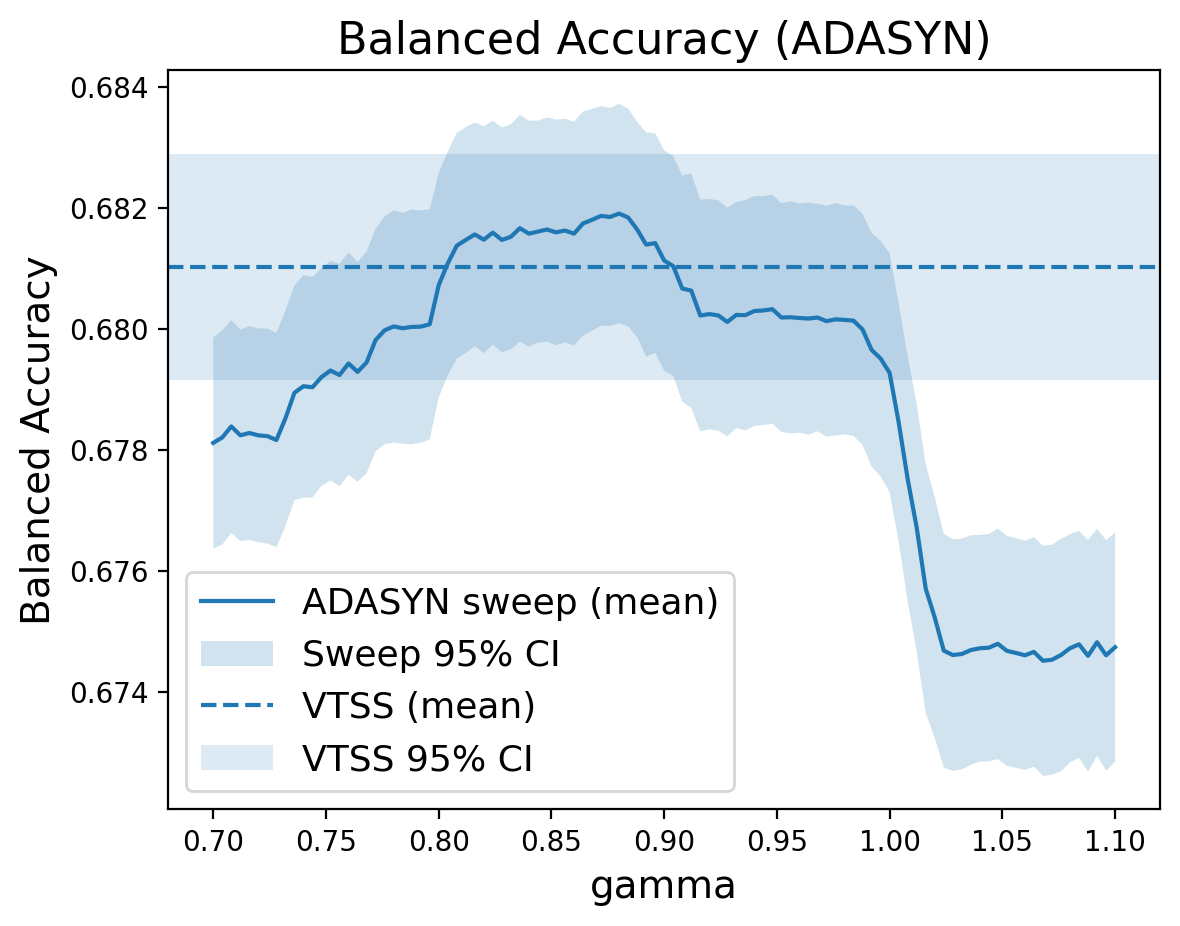}
  \end{subfigure}
  \begin{subfigure}[t]{0.244\textwidth}
    \centering
    \includegraphics[width=\linewidth]{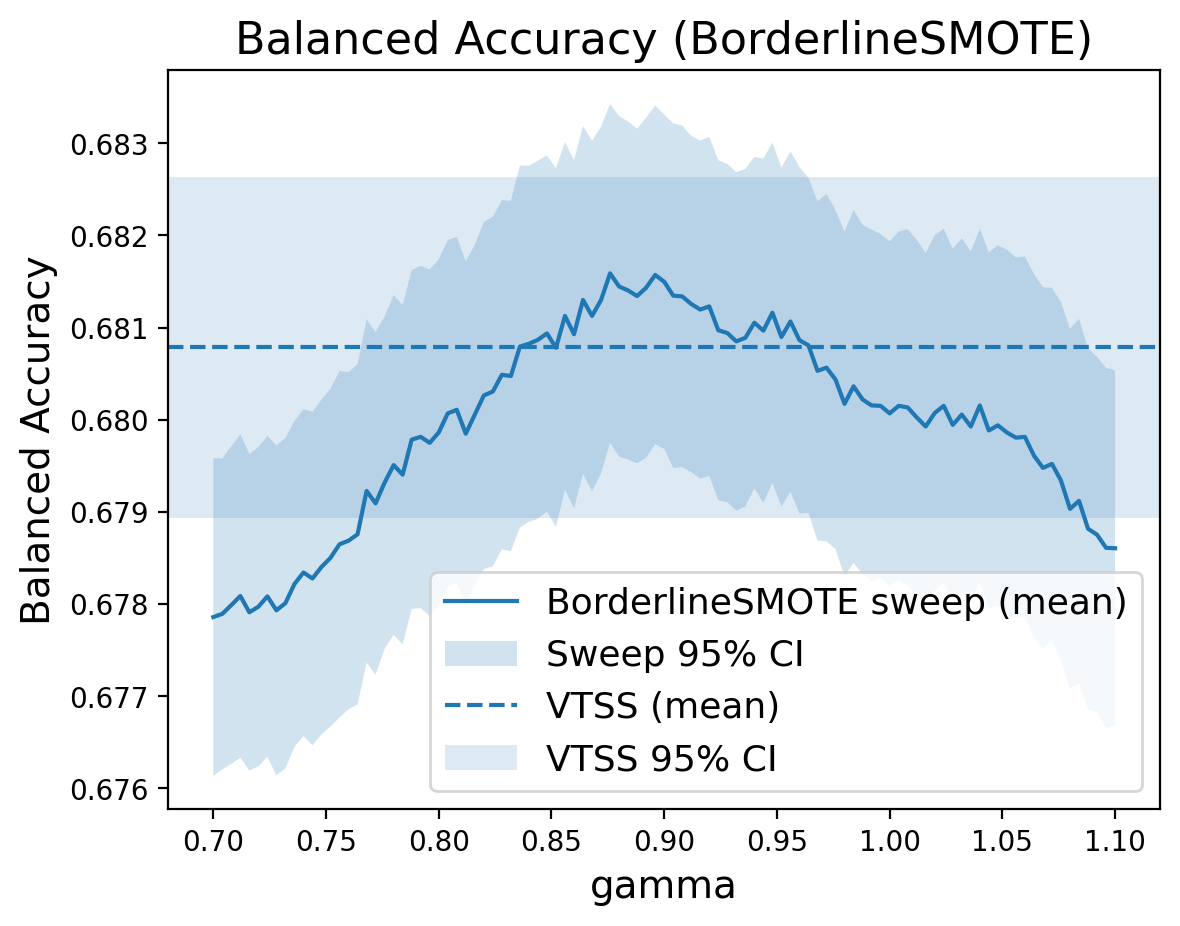}
  \end{subfigure}
  \begin{subfigure}[t]{0.244\textwidth}
    \centering
    \includegraphics[width=\linewidth]{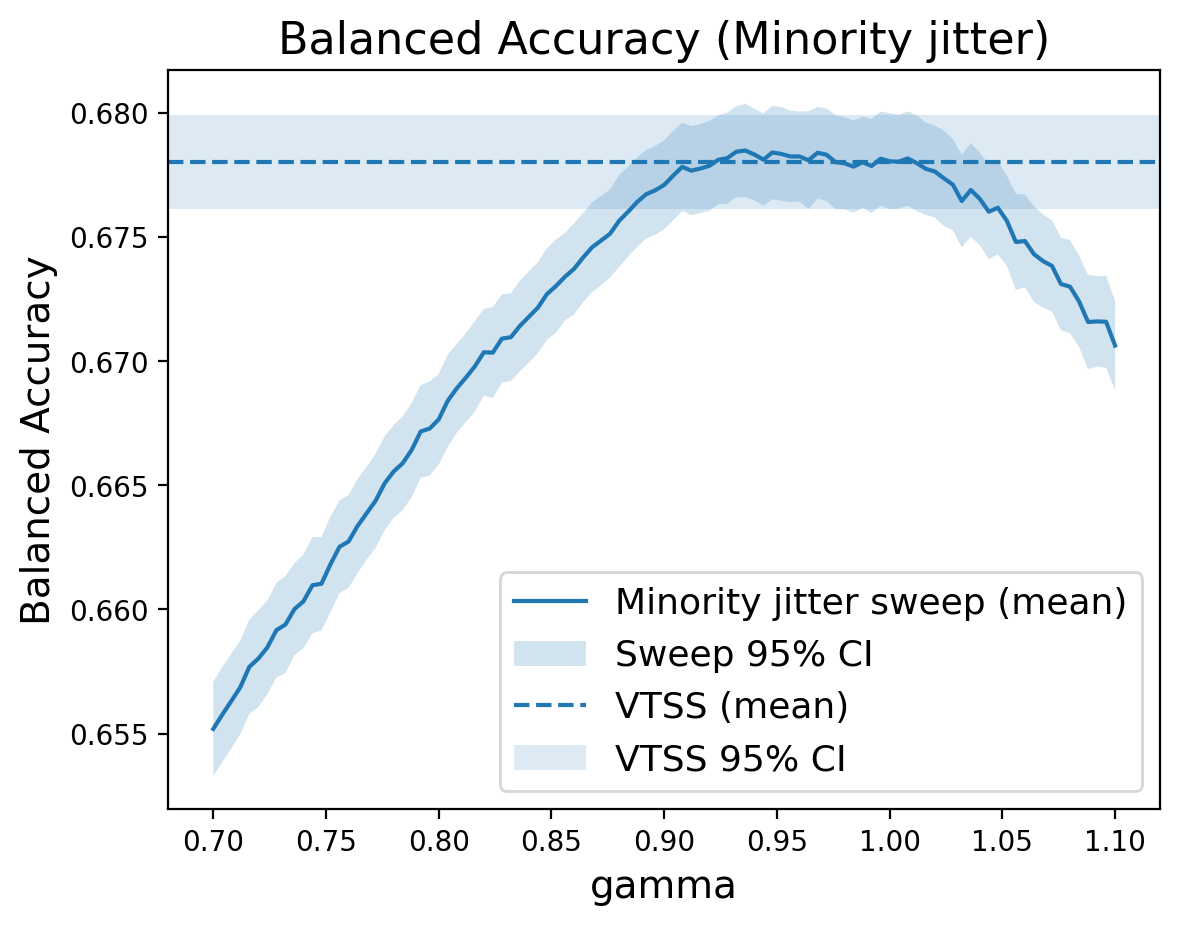}
  \end{subfigure}
  
  \caption{Balanced accuracy in predicting hospital mortality across four synthetic generators.}
  \label{fig:mortalitybalancedaccuracy}
\end{figure}

\paragraph*{Alternative validation objective: balanced accuracy.} To further assess the flexibility of VTSS, we also consider an alternative validation objective motivated by Remark~\ref{rmk:algorithm}. Instead of selecting $\gamma$ by minimizing validation balanced log-loss, we instantiate VTSS with a decision-oriented criterion and choose $\gamma$ to maximize validation balanced accuracy, keeping the pipeline unchanged. Figure~\ref{fig:mortalitybalancedaccuracy} reports balanced accuracy as a function of $\gamma$ for each generator, along with the VTSS-selected operating point and associated uncertainty.

Figure~\ref{fig:mortalitybalancedaccuracy} shows that VTSS consistently selects $\gamma$ values achieving near-optimal balanced accuracy across generators, while the naive balancing choice can be noticeably suboptimal. We also observe that VTSS does not always match the exact global maximizer, likely because balanced accuracy is thresholded and non-smooth, producing flat or step-like validation curves that are difficult to localize under finite-sample variability. Nonetheless, VTSS remains close to the best achievable balanced accuracy while outperforming naive balancing, supporting its use as a robust default for synthetic size selection under alternative metrics.

\section{Discussion}\label{sec:discussions}

This paper develops a unified risk-based theory for synthetic minority augmentation in imbalanced classification under a balanced evaluation objective. We derive an excess-risk decomposition separating imbalance-induced objective distortion from generator-induced mismatch, and show how the synthetic size $\tilde n$ controls the bias–variance trade-off. The theory identifies two regimes: under local asymmetry, augmentation can help, but the optimal $\tilde n$ depends on generator quality and may deviate from naive full balancing (especially under directional alignment); under local symmetry, imbalance is not the first-order bottleneck, so realistic augmentation typically cannot help and may hurt by amplifying mismatch. These insights motivate VTSS, which selects $\tilde n$ using balanced validation loss to capture gains when tuning matters while avoiding over-synthesis when augmentation is unhelpful.

Our analysis focuses on a classical low-dimensional asymptotic regime with fixed covariate dimension $d$. Extending the framework to high-dimensional settings, where both downstream learning and generator mismatch can change qualitatively, and where mismatch may worsen due to curse-of-dimensionality effects, is an important direction. In such regimes, understanding how mismatch scales with $d$, how the bias–variance trade-off shifts, and how to adapt VTSS when validation is noisy or limited remain open problems.

Our results also clarify settings where oversampling is less likely to help. For likelihood-based generative discriminant models such as LDA, class imbalance primarily enters through an explicit prior term, so balanced or cost-sensitive operating points can often be achieved by adjusting priors at prediction time rather than altering the training sample composition. Moreover, because these methods are sensitive to moment misspecification, synthetic samples can be counterproductive if they introduce even mild mean or covariance distortions.

Another message is that the direction of generator mismatch matters. This connects naturally to guided generative modeling (e.g., guided diffusion or instruction-tuned LLMs): if guidance can steer the dominant mismatch component toward the imbalance-relevant direction, augmentation becomes more effective. This suggests future work on augmentation-aware generator training that optimizes not only sample realism but also mismatch in the directions most relevant to the downstream balanced risk.

Synthetic and real minority samples are generally dependent when the generator is trained on the same minority data, so treating them as independent can invalidate common arguments. A simple fix is data-splitting (train the generator on one subset, augment using the other), and when data are scarce, K-fold cross-fitting can preserve sample efficiency while avoiding within-fold reuse. These strategies are widely used \citep{tian2025conditional, xia2024advancing}.

Finally, synthetic data is widely used beyond imbalance, including in fairness and privacy settings \citep{ktena2024generative}. Our analysis suggests a common principle across these applications: benefits depend jointly on how the synthetic distribution differs from the target and how much synthetic data is introduced. Extending our risk decomposition to group-weighted objectives and integrating mismatch control with privacy notions are promising directions, but the practical takeaway remains: treat the synthetic size as a tunable decision rather than a fixed balancing heuristic.

\section*{Disclosure statement}\label{disclosure-statement}
The authors declare no conflict of interest.

\section*{Data Availability Statement}\label{data-availability-statement}

The MIMIC-III database is publicly available, but access is restricted due to patient privacy concerns. Access can be obtained by completing the required CITI training and a data use agreement through PhysioNet. See details at \url{https://physionet.org/content/mimiciii/1.4/}.


%% file: supp_body.tex
\def\spacingset#1{\renewcommand{\baselinestretch}%
{#1}\small\normalsize} \spacingset{1}


\if1\anon
{
  \begin{center}
{\LARGE\bfseries
Supplementary Materials for 
“Synthetic Augmentation in Imbalanced Learning:
When It Helps, When It Hurts, and How Much to Add”
\par}

\vspace{1em}

{\large
Zhengchi Ma\\
Department of Electrical \& Computer Engineering, Duke University\\[0.8em]
Anru R. Zhang\\
Department of Biostatistics \& Bioinformatics\\ and Department of Computer Science, Duke University
\par}
\end{center}

\vspace{1em}
} \fi

\if0\anon
{
  \bigskip
  \bigskip
  \bigskip
  \begin{center}
{\LARGE\bfseries
Supplementary Materials for\\[0.5em]
“Synthetic Augmentation in Imbalanced Learning:
When It Helps, When It Hurts, and How Much to Add”
\par}
\end{center}

\vspace{1em}
  \medskip
} \fi

\bigskip
\begin{abstract}
 In the supplementary materials we provide full details of the simulation studies and additional results on both simulation and real data analysis. The details of examples in the main paper are presented and the proofs of all technical results are included. Finally, we discuss related losses, classifiers, and synthetic data generators.
\end{abstract}

\spacingset{1.8} 


\section{Additional Results and Details for Simulations}\label{sec:simulation-detail}

\subsection{Under Local Asymmetry}
\subsubsection{For Biased Synthetic Generator}
In the local asymmetry case, we study the classification with directional alignment.
To illustrate the directional-alignment phenomenon for biased synthetic generator discussed in Section~\ref{sec:inconsistent} of the main paper, we consider a 2D Gaussian setting as in Example~\ref{biasedexample}, with imbalance ratio $n_0/n_1=20$. For $n_1$ varying from 6 to 3200, we set $n_0=20n_1$, draw majority samples $\bx\mid y=0\sim\cN(0,I_2)$ and minority samples $\bx\mid y=1\sim\cN(\bmu_1,I_2)$ with $\bmu_1=(\mu,0)$ and $\mu=1$.
We augment the training data with $\tilde n$ synthetic points labeled as minority, generated from a biased but directionally aligned generator $\bx_{\mathrm{syn}}\sim\cN(\bmu_s,I_2)$ with $\bmu_s=(a,0)$ and $a=0.5$.
For each repetition (100 Monte Carlo runs per $n_1$), we also generate a fresh balanced test set of size 5000 per class from the true distributions.
We compare two fixed synthetic-size rules: the naive balancing choice $\tilde n=n_0-n_1$ and the bias-canceling choice $\tilde n=4(n_0-n_1)$ (from the analysis of Example~\ref{biasedexample}).
On each augmented training set, we fit the linear predictor $\hat\btheta\in\mathbb R^2$ by least squares and evaluate (i) parameter error $\|\hat\btheta-\btheta^*\|_2$ and (ii) balanced excess risk $\cR(\hat\btheta)-\cR(\btheta^*)$, where $\cR(\btheta)=\|\btheta\|_2^2+\tfrac12(\btheta^\top \bmu_1-1)^2$ and $\btheta^*=(\mu/(\mu^2+2),0)$.
Finally, we plot the mean and standard deviation of the excess risk and parameter error versus $n_1$ on log scales for the two choices of $\tilde n$. The results are reported in Figure \ref{fig:bias} in the main paper.

\subsubsection{For Consistent Synthetic Generators}
For directional alignment with a realistic and consistent synthetic generator (Section~\ref{sec:rate-improve} of the main paper), we simulate the same 2D Gaussian setup as above under imbalance $n_0/n_1=20$ and squared loss.
For $n_1\in[10,500]$, we run 100 Monte Carlo repetitions.
Synthetic minority samples are produced by a realistic, consistent, and directionally aligned generator: for a requested synthetic size $k$, draw $\bx_{\mathrm{syn}}\sim \cN(\bmu_{\mathrm{syn}}, I_2)$ with
\(
\bmu_{\mathrm{syn}}=\left(1-(\log n_1)^{-1/2}\right)\bmu_1,
\)
so that the synthetic mean approaches $\bmu_1$ as $n_1$ increases while remaining aligned with $\bmu_1$.
We compare VTSS with the naive balancing rule $\tilde n=n_0-n_1$, fit a least-squares predictor on the augmented training data, and evaluate both methods using population excess risk $\cR(\hat\btheta)-\cR(\btheta^*)$ and parameter error $\|\hat\btheta-\btheta^*\|_2$, reporting mean and empirical $95\%$ confidence interval over repetitions as functions of $n_1$.
The results are shown in Figure~\ref{fig:consistentalign} in the main paper.

\subsection{Under Local Symmetry}
\subsubsection{Mean-Shift Model}
To further illustrate local symmetry (Section~\ref{sec:synnohelp} of the main paper), we conduct a simulation study based on the mean-shift model in Example~\ref{exa:meanshiftcancel}. We generate imbalanced binary data in dimension $d=20$ from
\(
\bx\mid y=1\sim \bmu+\bxi,~ \bx\mid y=0\sim -\bmu+\bxi,
\)
where $\bmu=\delta \be_1$ with $\delta=1$ and $\be_1=(1,0,\ldots,0)^\top$. Here $\bxi$ has mean zero and identity covariance. We consider three choices of $\bxi$: (i) Uniform cube with independent coordinates $\xi_j\sim \mathrm{Unif}[-\sqrt{3},\sqrt{3}]$, (ii) Rademacher with $\xi_j\in\{-1,+1\}$ equally likely, and (iii) Uniform-on-sphere obtained by sampling $\boldsymbol{g}\sim\cN(0,I_d)$ and setting $\bxi=\sqrt{d}\boldsymbol{g}/\|\boldsymbol{g}\|$.
We start from an imbalanced training sample with $n_1=100$ and $n_0=2000$, and augment the minority class with $\tilde n=\mathrm{round} (\gamma (n_0-n_1))$ synthetic samples, where $\gamma$ ranges over 20 values in $[0,4]$.
We fit a linear predictor under squared loss on the augmented sample and report the balanced excess risk $\mathcal R(\hat\btheta)-\mathcal R(\btheta^*)$, where $\btheta^*=\bmu/(1+\|\bmu\|_2^2)$ and $\mathcal R(\btheta)$ is evaluated in closed form under the population model.
We compare four synthetic generators: (i) \textbf{Oracle}, which draws synthetic points from the true minority distribution $\bmu+\bxi$; (ii) \textbf{SMOTE}, which interpolates between minority neighbors (with $k=5$); (iii) \textbf{Gaussian-fit}, which samples from $\cN(\hat\bmu_1,\hat\Sigma)$ using the empirical minority mean and covariance estimate; and (iv) \textbf{Semi-Oracle}, which samples $\bxi$ from the true noise distribution but centers at the empirical minority mean $\hat\bmu_1$ (i.e., $\hat\bmu_1+\bxi$).
Results are averaged over 100 Monte Carlo repetitions, and we plot the mean balanced excess risk across repetitions as a function of $\gamma$ on a log scale. The results are reported in Figure~\ref{fig:nosyn-meanshift} of the main paper.

\begin{figure*}[htb!]
  \centering
  \includegraphics[width=\linewidth]{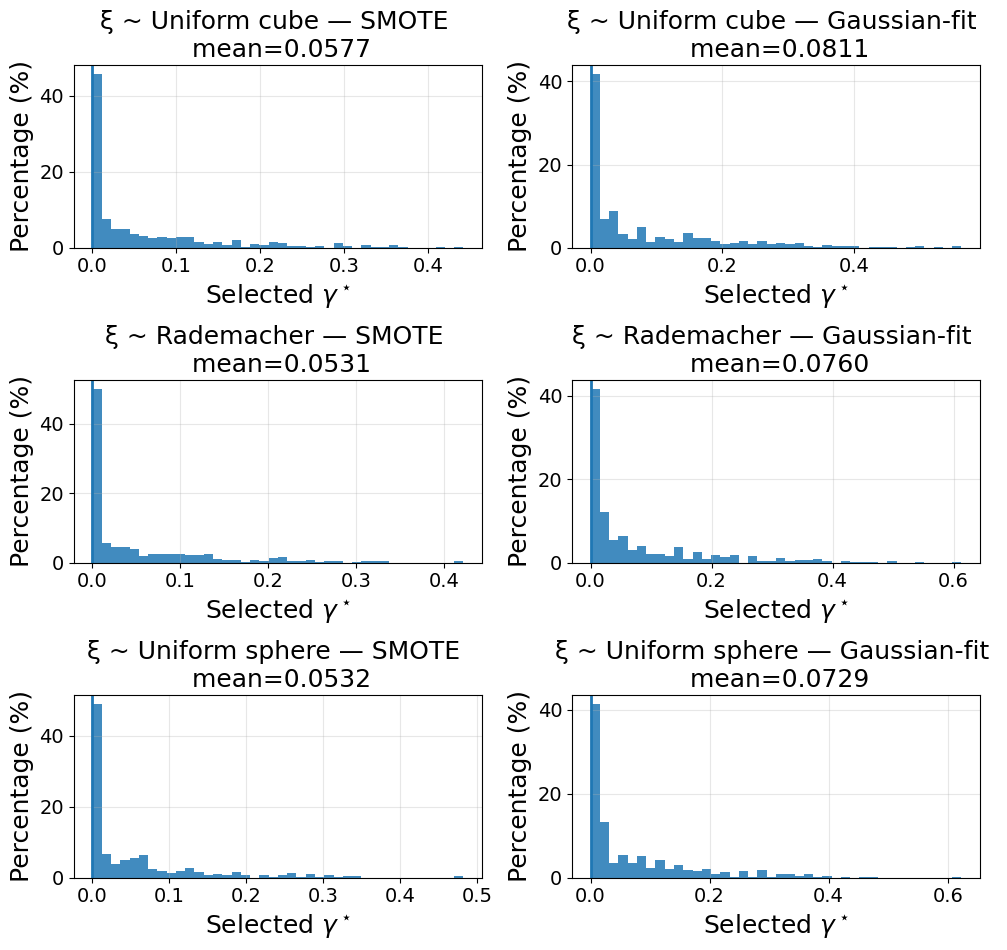}

  \caption{Histograms of the selected synthetic multiplier $\gamma^*$ chosen by VTSS.}
  \label{fig:select0}
\end{figure*}

We also examine the VTSS-selected multiplier $\gamma^*$ in the same mean-shift setup. Rather than sweeping $\gamma$ to plot excess-risk curves, we focus on the empirical distribution of the minimizer $\gamma^*$.
For each noise distribution $\bxi$ (Uniform cube, Rademacher, Uniform sphere), we run 500 repetitions. In each repetition, we generate a fresh balanced validation set with 2000 samples per class and tune over a grid of 200 equally spaced candidates $\gamma\in[0,2]$ (including $\gamma=0$).
For each candidate $\gamma$ and each generator, we fit the least-squares predictor on the augmented training set and set $\gamma^*$ to the minimizer of the validation loss.
Figure~\ref{fig:select0} reports the empirical distribution of $\gamma^*$. VTSS selects a very small $\gamma^*$ in the vast majority of runs and frequently chooses $\gamma^*=0$ exactly. The intuition is that when synthetic augmentation provides no genuine benefit and can introduce mismatch-induced bias, the balanced validation objective typically does not decrease as $\gamma$ increases. Hence, the minimizer is attained at (or near) the boundary $\gamma=0$, allowing the procedure to automatically opt out of synthetic data in this regime.

\subsubsection{Sigmoid Bernoulli Logistic Model}\label{sec:simu-symmetry-sigmoid}
\begin{figure*}[htb]
  \centering

  \begin{subfigure}[htb]{0.45\textwidth}
    \centering
    \includegraphics[width=\linewidth]{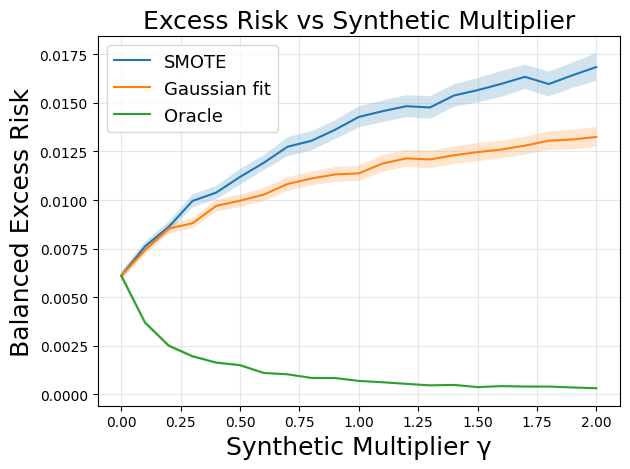}
  \end{subfigure}\hfill
  \begin{subfigure}[htb!]{0.45\textwidth}
    \centering
    \includegraphics[width=\linewidth]{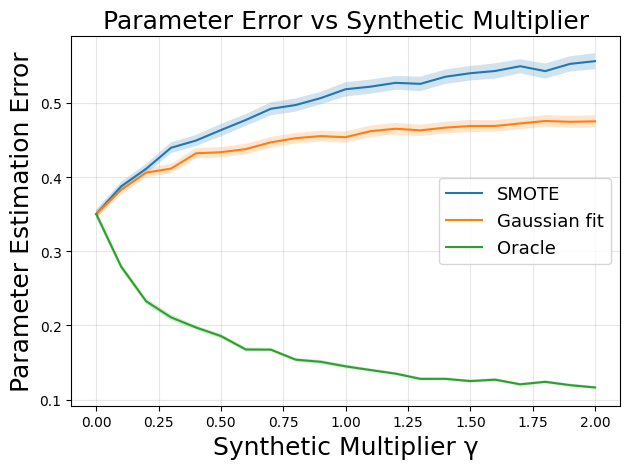}
  \end{subfigure}\hfill

  \caption{Synthetic augmentation does not help under a bias-cancellation sigmoid Bernoulli logistic model.}
  \label{fig:nosynsigmoid}
\end{figure*}

To further demonstrate a regime where class imbalance is not the bottleneck, we consider the sigmoid Bernoulli logistic model in Example~\ref{exa:sigmoidbernoullicancel} with dimension $d=20$.
We let $v=\be_1$ and $\btheta_{\mathrm{true}}=c\bv$ with $c=1$, and generate covariates as $\bx=T \bv+W$, where $T\in\{-a,b\}$ with $a=5$, $b=1$, and $\bP(T=-a)=\alpha$ chosen by the bias-cancellation formula (equation~(\ref{alphachoice}) in the Supplement). The noise $W$ is orthogonal to $\bv$ (its first coordinate is identically zero) and has a symmetric, non-Gaussian mixture distribution in the remaining coordinates: $W_{2:d}=\bz+s\bmu$ with $\bz\sim\cN(0, I)$, $s\in\{\pm1\}$ equally likely, and $\bmu$ having its first three entries equal to $2.5$ (others zero), making Gaussian-fit imperfect.
Labels are drawn as $y\in\{0,1\}$ with $\bP(y=1\mid \bx)=\sigma(\btheta_{\mathrm{true}}^\top \bx)$.
For each of 100 repetitions, we sample $n_{\mathrm{train}}=5000$ i.i.d.\ training pairs $(\bx,y)$ (with random realized class counts), define the minority class as the label with fewer observations, and augment only that minority class to size $n_{1}+\tilde n$, where $\tilde n=\mathrm{round}(\gamma (n_{0}-n_{1}))$ and $\gamma$ ranges over $[0,2]$.
We compare three generators for minority augmentation: SMOTE (nearest-neighbor interpolation with $k=5$), Gaussian-fit (sampling from a fitted Gaussian using the minority empirical mean and covariance with a small ridge), and an Oracle generator that draws fresh $(\bx,y)$ from the true model and keeps only minority-labeled points.
For each augmented dataset, we fit logistic regression and report (i) balanced excess logistic risk relative to $\btheta^*$ and (ii) parameter error $\|\hat\btheta-\btheta^*\|_2$, where $\btheta^*$ is obtained by fitting the same model on a large balanced sample of 20{,}000 per class.

Figure~\ref{fig:nosynsigmoid} shows that, for realistic generators such as SMOTE and Gaussian-fit, increasing $\gamma$ does not reduce balanced excess risk or parameter error; in fact, performance typically worsens. In contrast, the Oracle generator improves monotonically, as expected.

\subsubsection{Further Evaluation of VTSS}
We consider two configurations in Section \ref{sec:furtherevaluation} in the main paper.

\paragraph*{Linear classification (SMOTE + logistic regression).}
We generate imbalanced training data in $d=5$ with ratio $r=n_0/n_1=20$ (in the plotted curve, $n_1=200$, $n_0=4000$, and $n_0-n_1=3800$). The majority class is $\cN(0,I)$, while the minority class is a balanced two-component Gaussian mixture, with each component having covariance $I$ and means $\bmu_{\pm}=\delta \be_1\pm \frac{\xi}{2}\be_2$.
For each of 100 repetitions, we sweep $\tilde n$ over a grid induced by $\gamma\in[0.6,1.4]$ via $\tilde n=\mathrm{round}(\gamma(n_0-n_1))$, apply SMOTE to reach minority size $n_1+\tilde n$, fit logistic regression, and evaluate balanced test log-loss on a fixed balanced test set (10{,}000 per class). VTSS chooses $\tilde n$ in each repetition by minimizing the same balanced log-loss on a validation set, and we plot its average test loss as a horizontal reference line.

\paragraph*{Nonlinear classification (perturbed sampling + kernel logistic).}
We use a non-Gaussian data-generating process in $d=5$.
For the majority class $y=0$, we first draw a mixture indicator $z\sim\mathrm{Bernoulli}(0.30)$, then sample the first two coordinates from a correlated Gaussian $(x_1,x_2)\sim\cN\left(( 1.5\cdot z,0),\begin{pmatrix}1&0.5\\0.5&1\end{pmatrix}\right)$, and sample the remaining coordinates i.i.d.\ as $x_{3:5}\sim\cN(0,I_3)$.
For the minority class $y=1$, we generate a noisy ring in the $(x_1,x_2)$-plane by drawing an angle $\omega\sim\mathrm{Unif}[0,2\pi]$ and radius $r=2.2+\varepsilon_r$ with $\varepsilon_r\sim\cN(0,0.12^2)$, then setting
\(
x_1=\delta + r\cos \omega+\varepsilon_1,~
x_2=r\sin \omega+\varepsilon_2,
\)
with $\varepsilon_1,\varepsilon_2\sim\cN(0,0.08^2)$ and $\delta=4.8$. The remaining coordinates are small Gaussian noise $x_{3:5}\sim\cN(0,0.08^2 I_3)$. This simulation is a controlled nonlinear-boundary benchmark, designed to test whether VTSS can tune synthetic size when the generator introduces distributional blur. We embed the 2D structure in $d=5$ by adding nuisance coordinates, reflecting the common setting where only a subset of features drives class separation.
Synthetic data are generated by perturbed resampling: bootstrap minority points and add Gaussian jitter $\cN(0,I)$.
We train an RBF-kernel probabilistic classifier, sweep $\tilde n=\mathrm{round}(\gamma(n_0-n_1))$ over $\gamma\in[0.6,1.4]$, and let VTSS pick $\tilde n$ via validation minimization in each repetition.

The results are reported in Figure~\ref{fig:curves} of the main paper.

\section{Additional Results for Real Data Analysis}
\begin{figure}[htbp]
  \centering

  \begin{subfigure}[t]{0.32\textwidth}
    \centering
    \includegraphics[width=\linewidth]{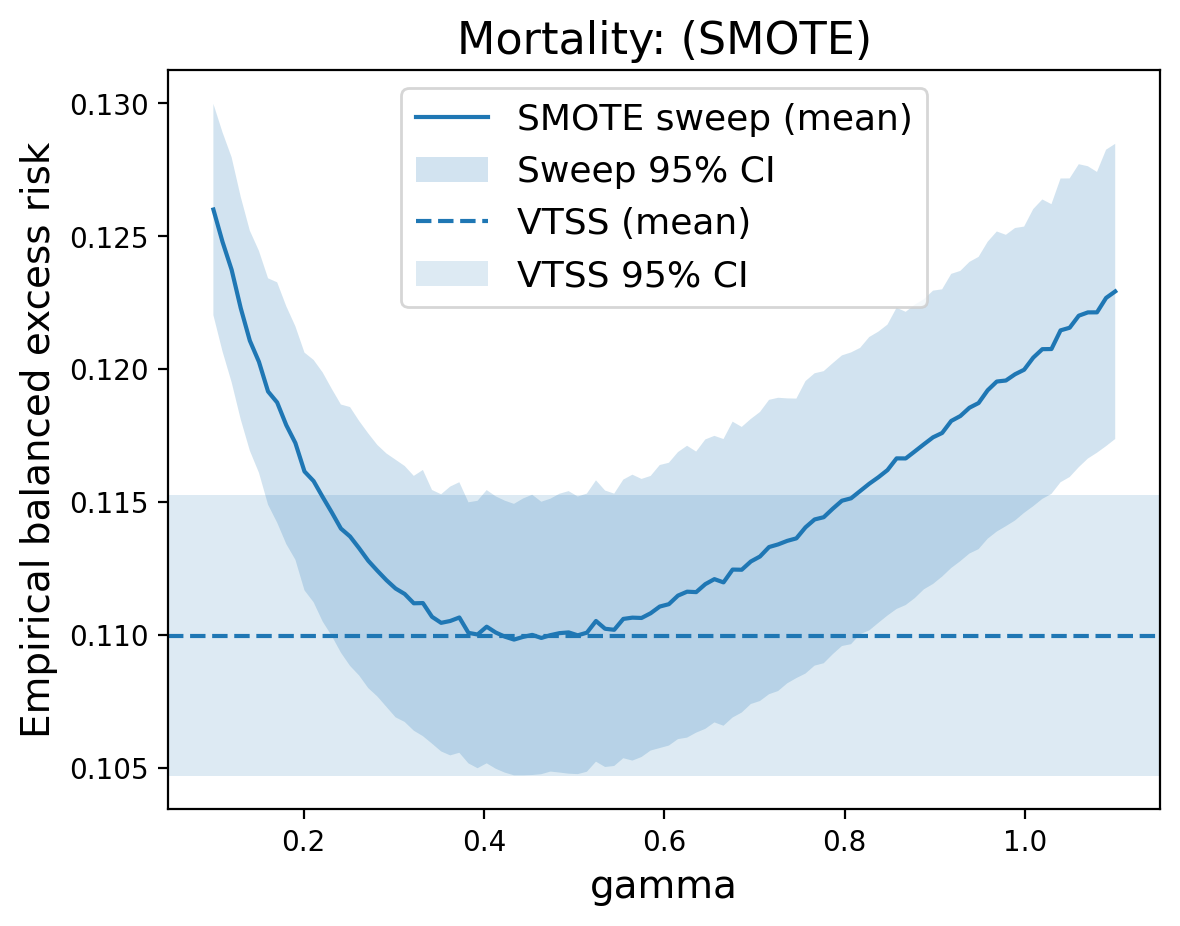}
  \end{subfigure}
  \begin{subfigure}[t]{0.32\textwidth}
    \centering
    \includegraphics[width=\linewidth]{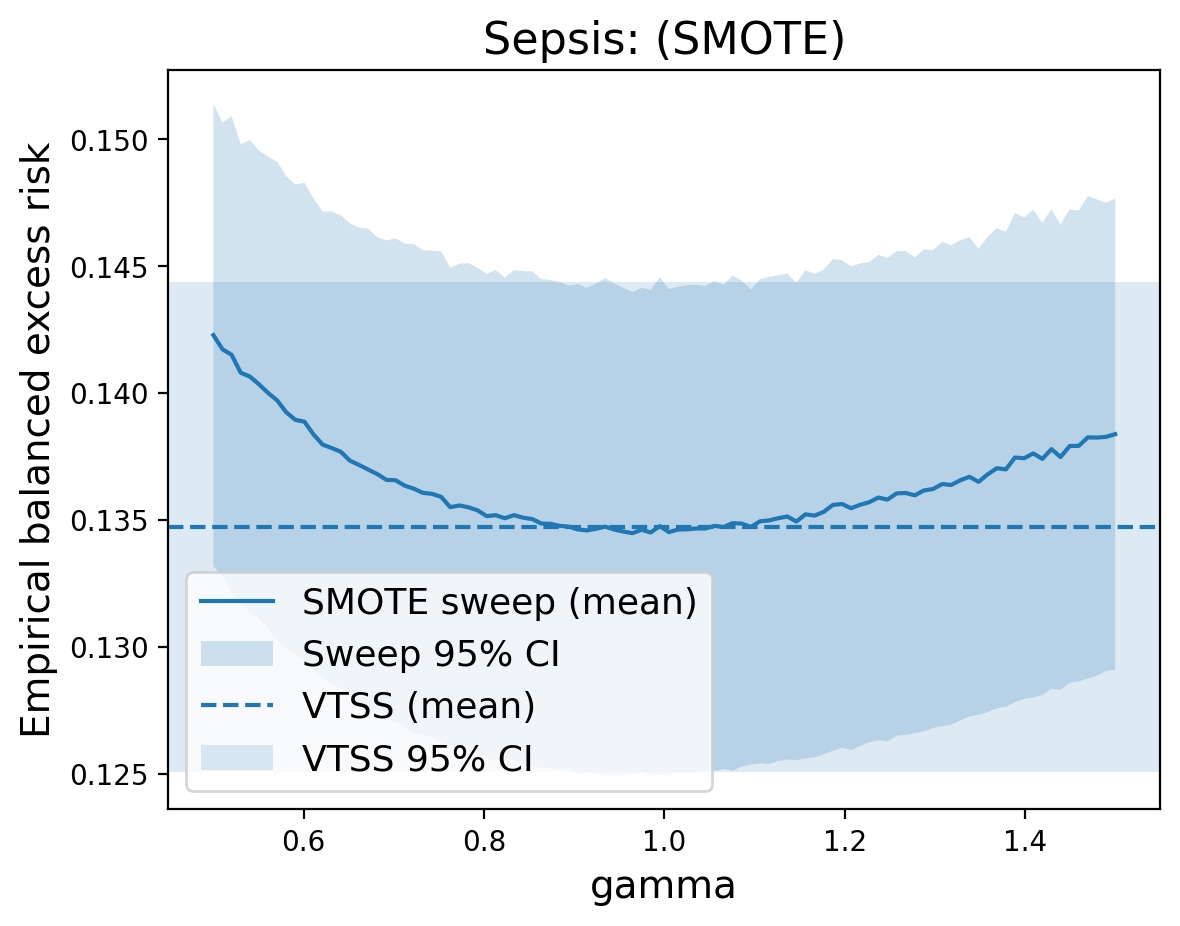}
  \end{subfigure}
  \begin{subfigure}[t]{0.32\textwidth}
    \centering
    \includegraphics[width=\linewidth]{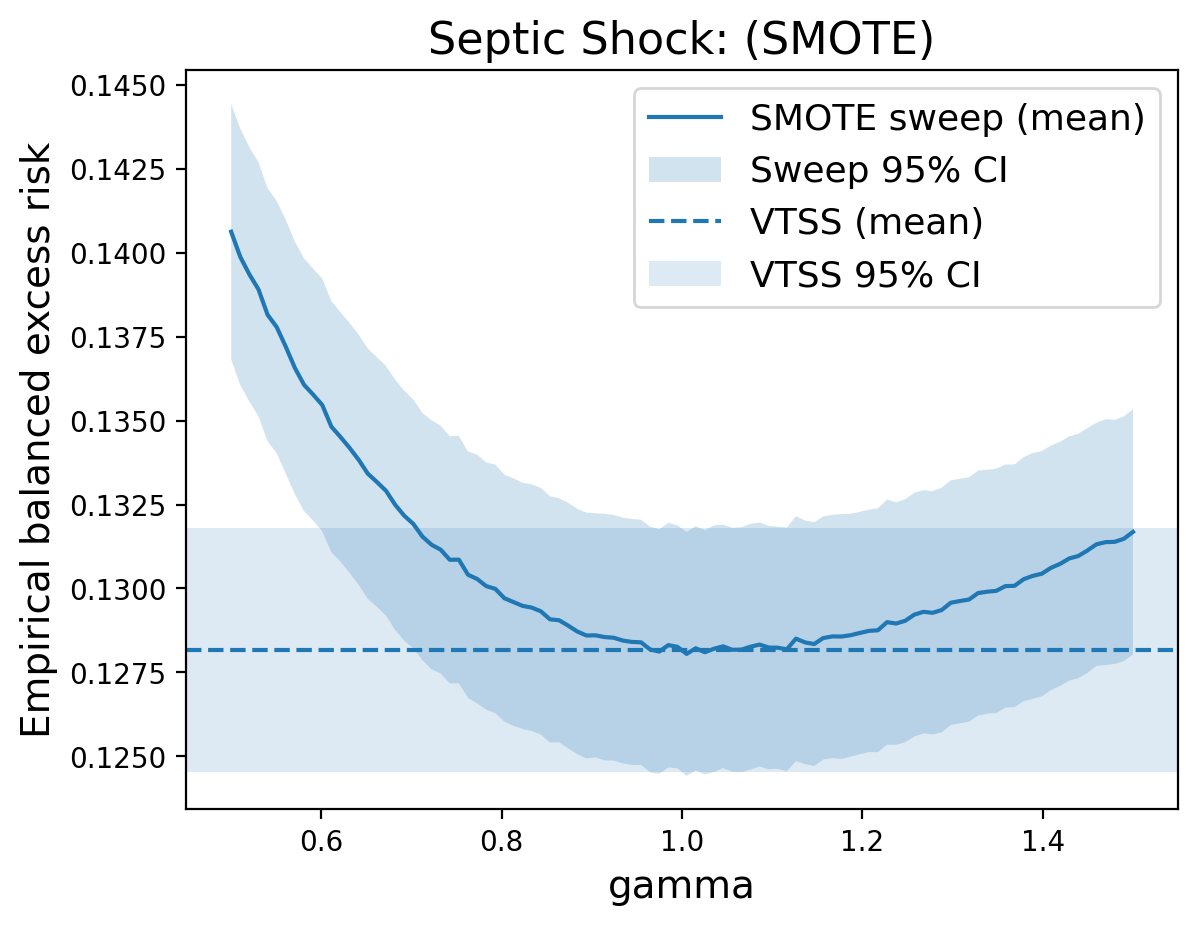}
  \end{subfigure}

  \begin{subfigure}[t]{0.32\textwidth}
    \centering
    \includegraphics[width=\linewidth]{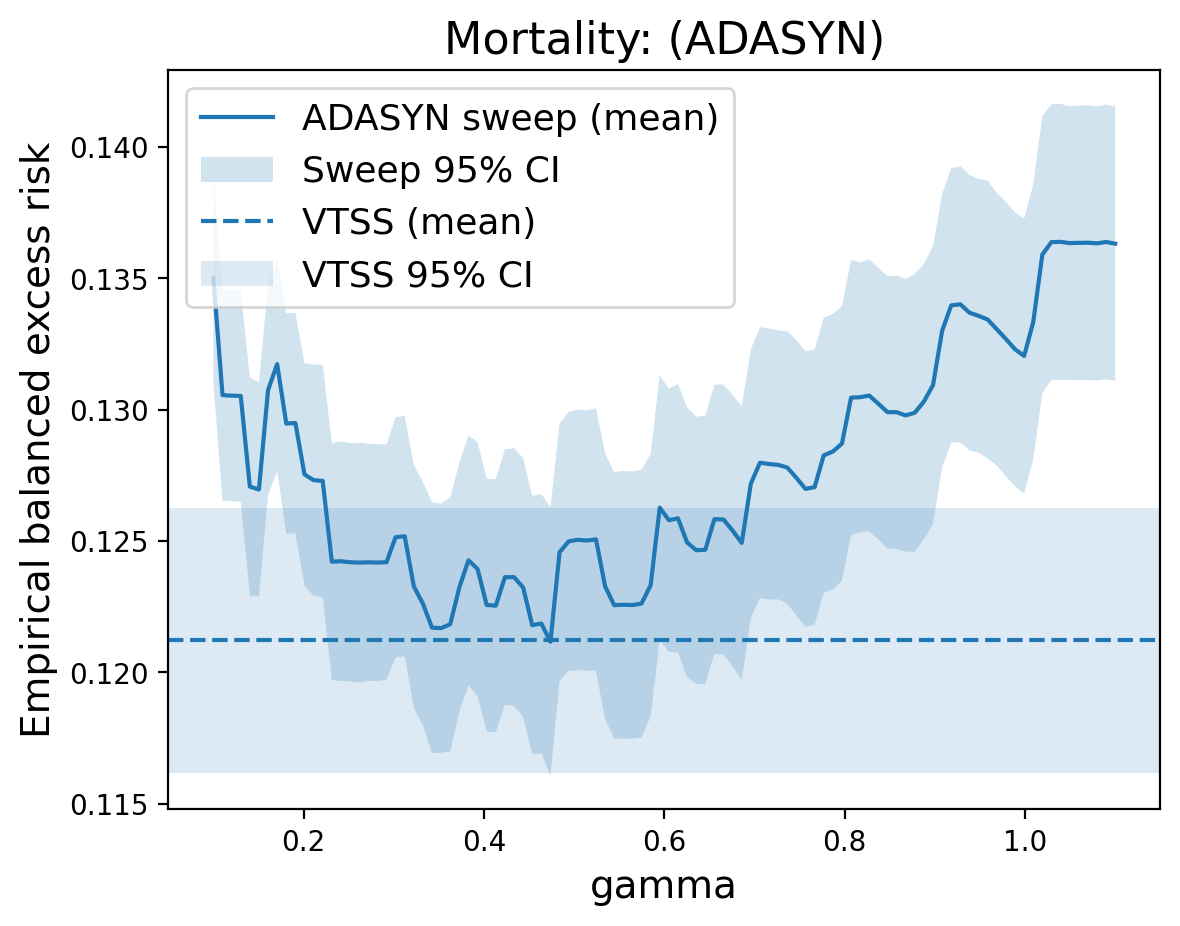}
  \end{subfigure}
  \begin{subfigure}[t]{0.32\textwidth}
    \centering
    \includegraphics[width=\linewidth]{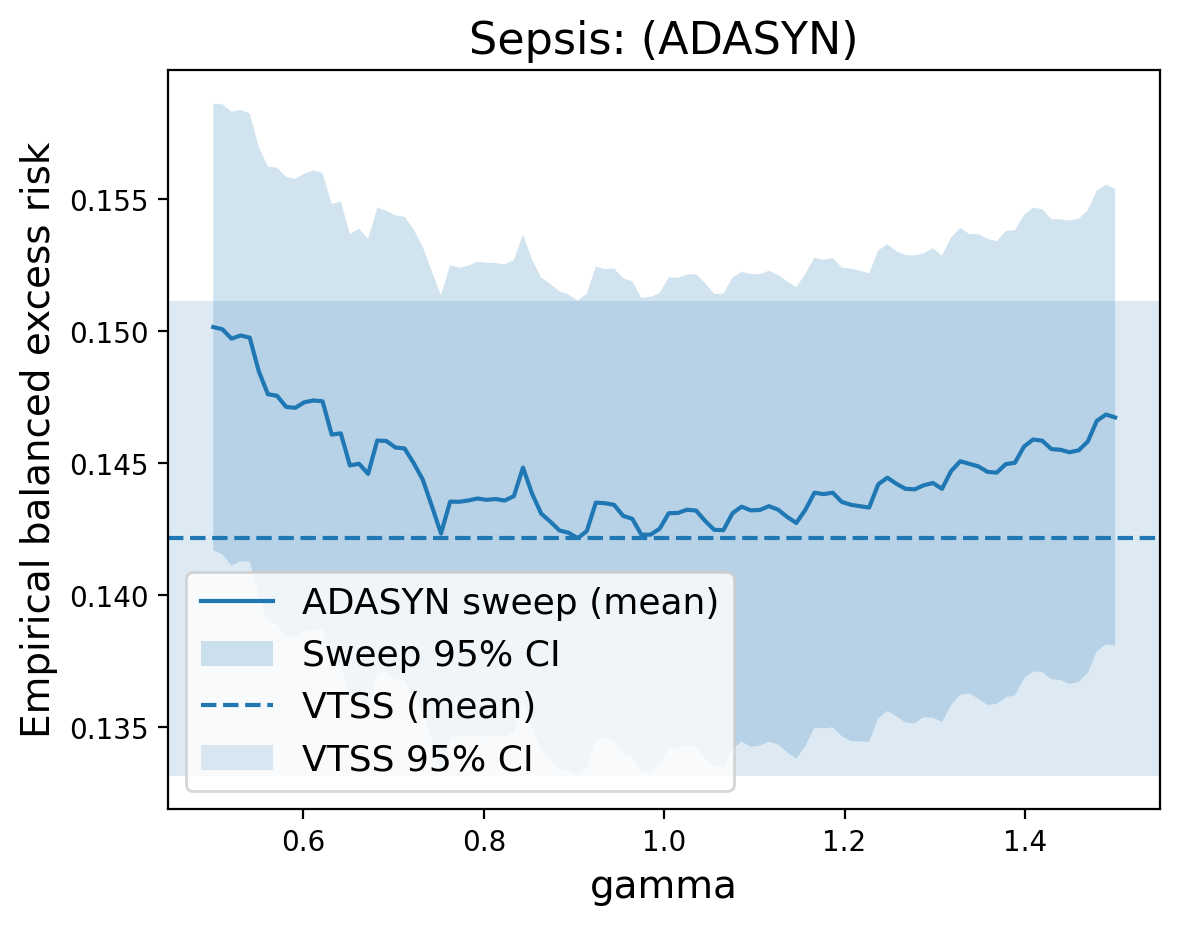}
  \end{subfigure}
  \begin{subfigure}[t]{0.32\textwidth}
    \centering
    \includegraphics[width=\linewidth]{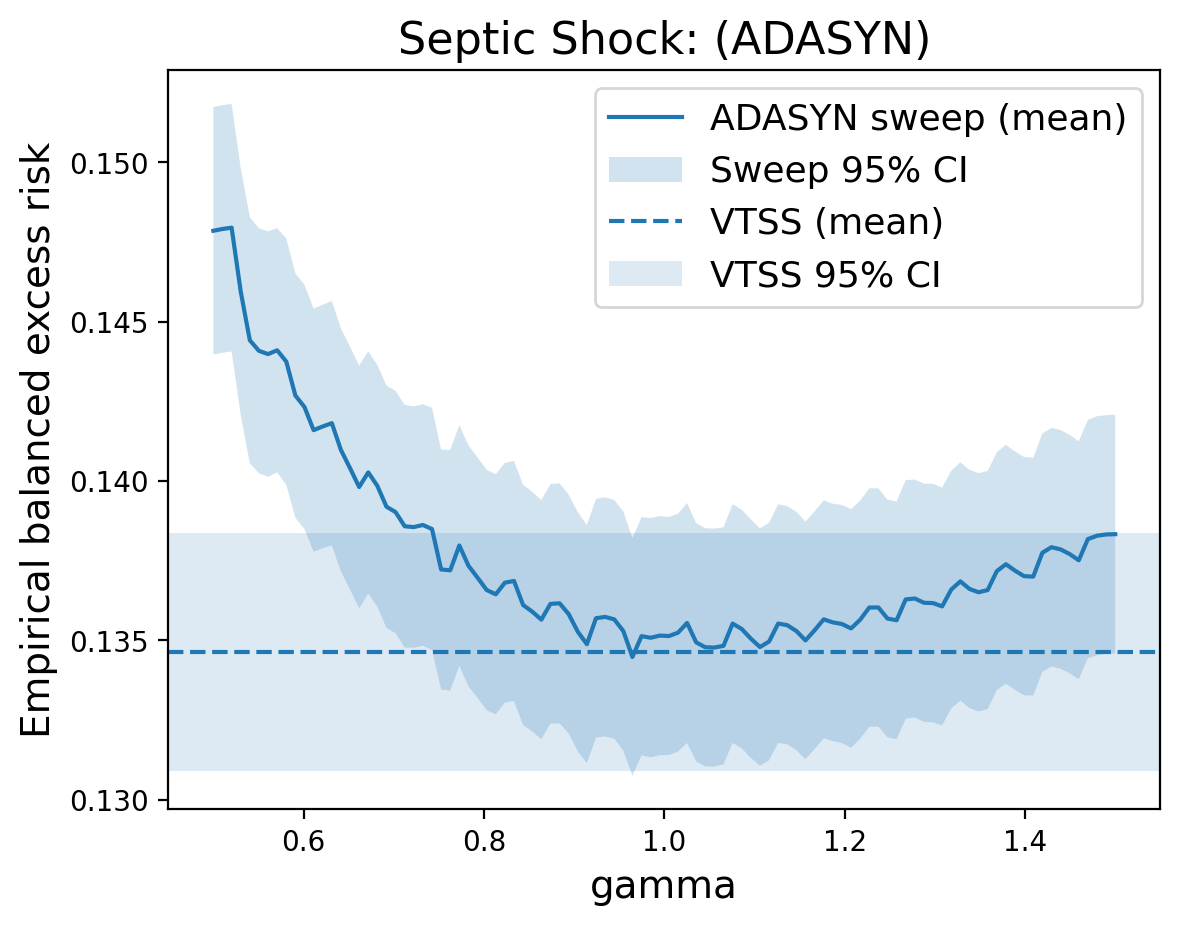}
  \end{subfigure}

  \begin{subfigure}[t]{0.32\textwidth}
    \centering
    \includegraphics[width=\linewidth]{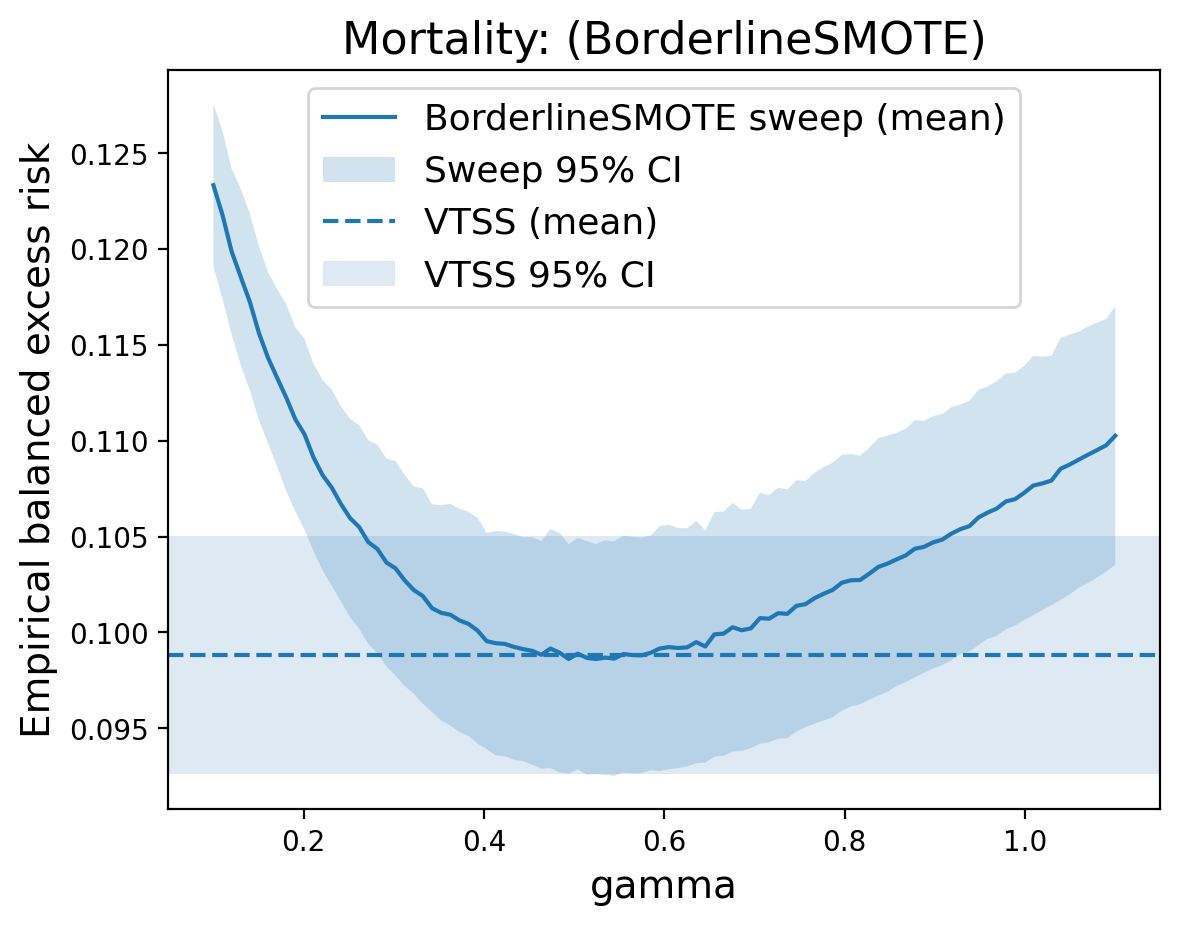}
  \end{subfigure}
  \begin{subfigure}[t]{0.32\textwidth}
    \centering
    \includegraphics[width=\linewidth]{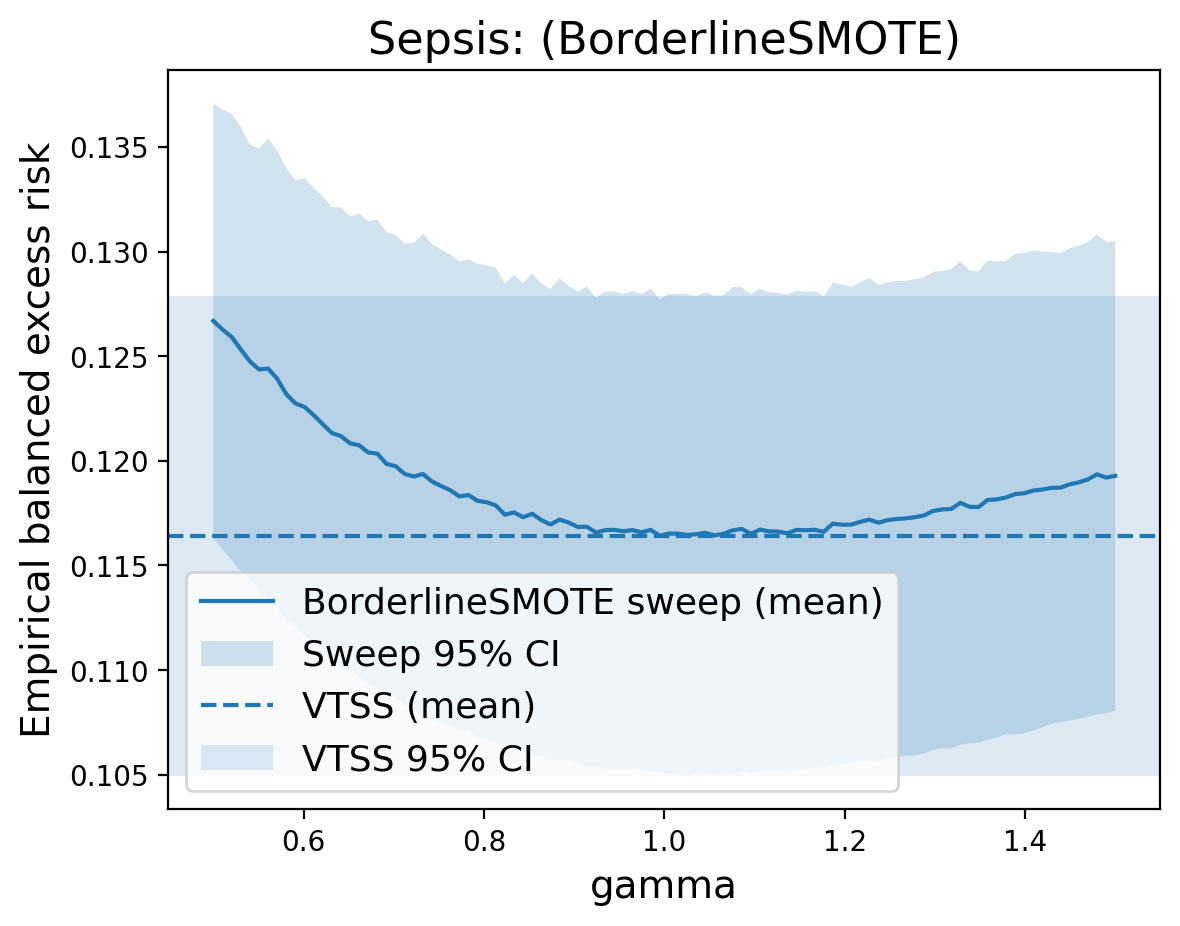}
  \end{subfigure}
  \begin{subfigure}[t]{0.32\textwidth}
    \centering
    \includegraphics[width=\linewidth]{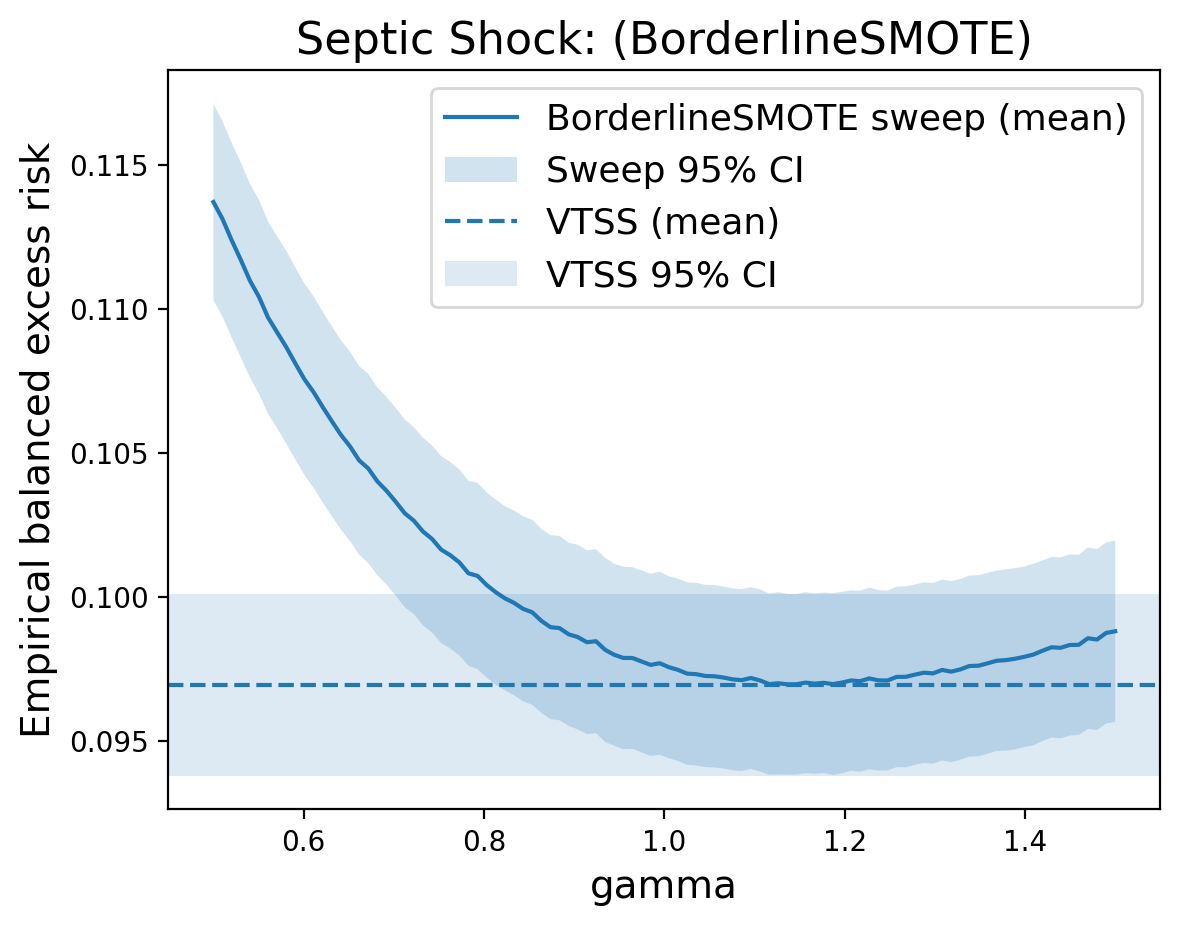}
  \end{subfigure}

  \begin{subfigure}[t]{0.32\textwidth}
    \centering
    \includegraphics[width=\linewidth]{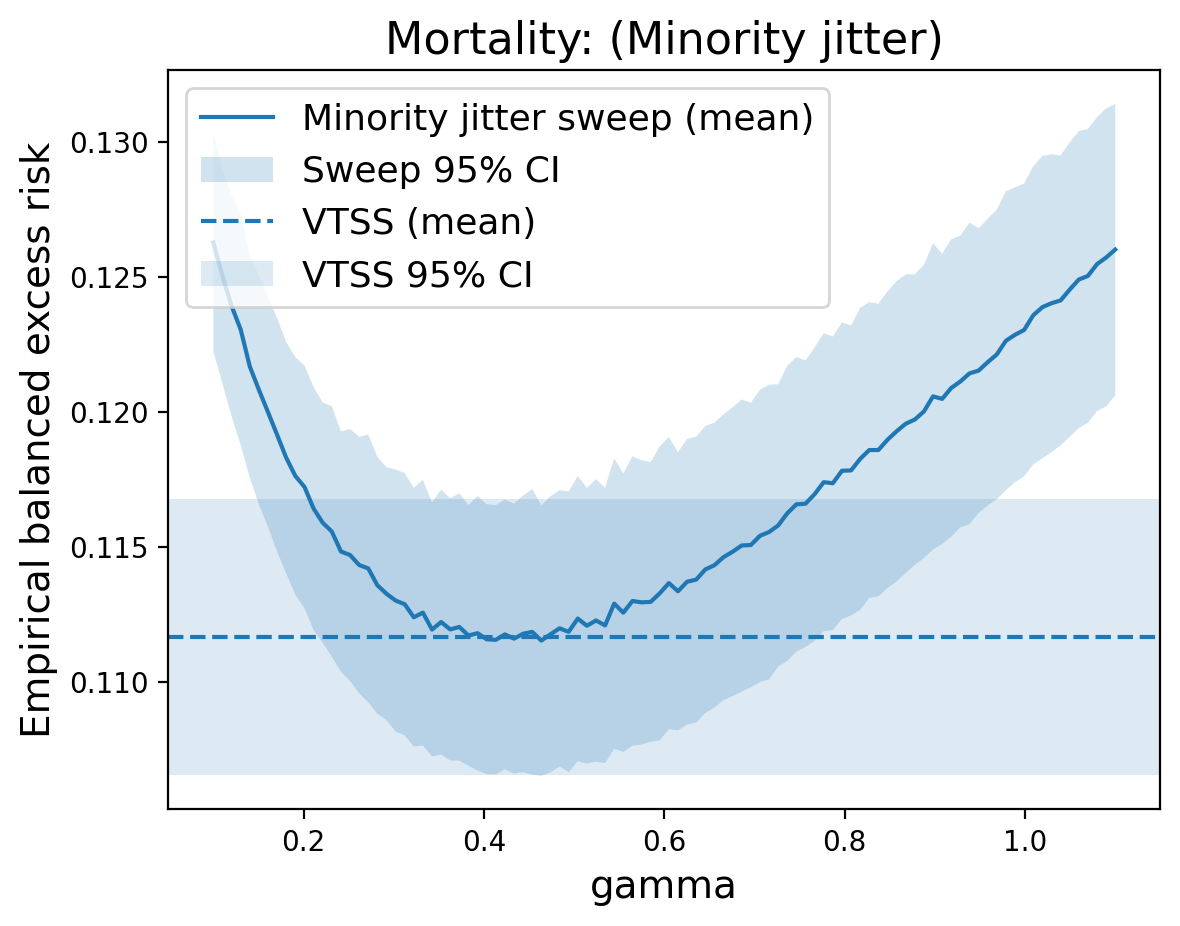}
  \end{subfigure}
  \begin{subfigure}[t]{0.32\textwidth}
    \centering
    \includegraphics[width=\linewidth]{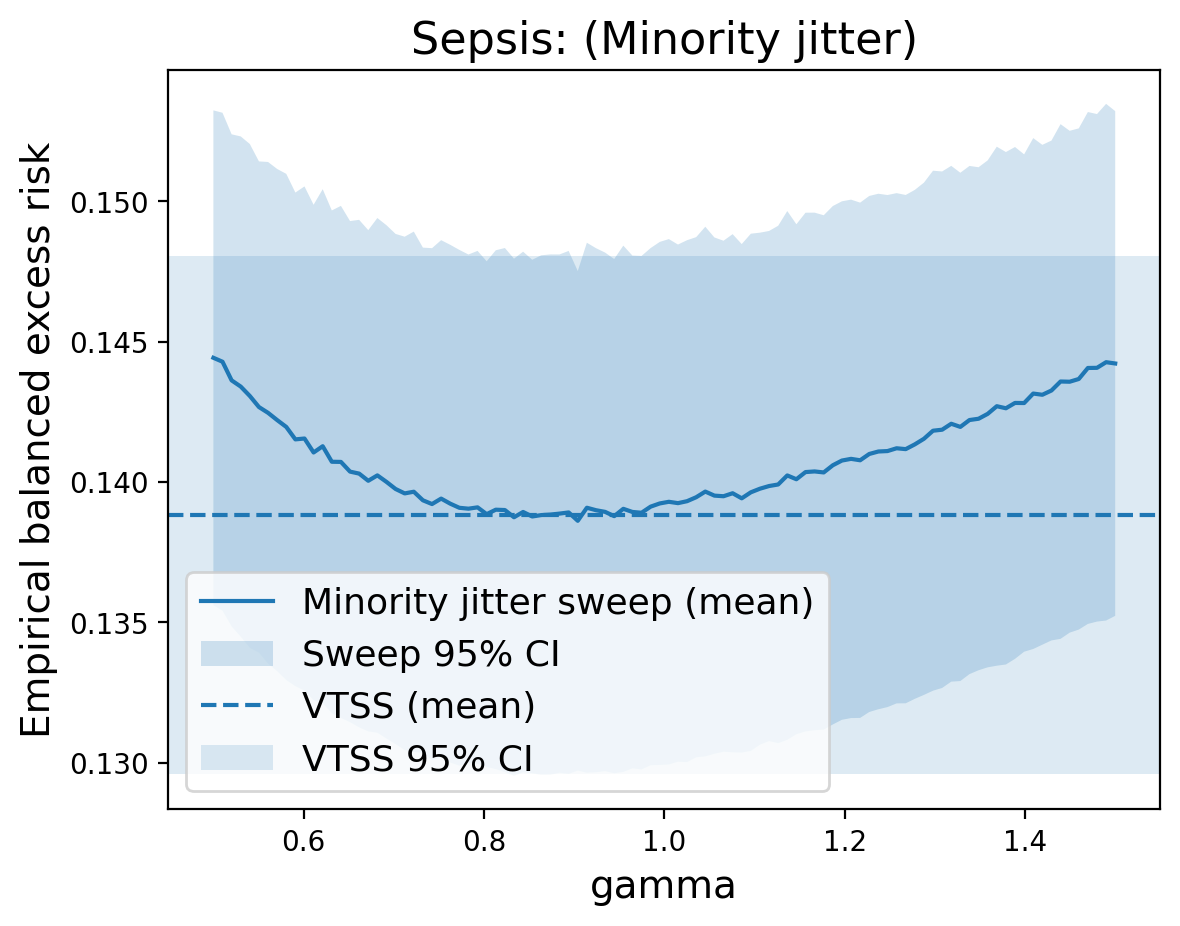}
  \end{subfigure}
  \begin{subfigure}[t]{0.32\textwidth}
    \centering
    \includegraphics[width=\linewidth]{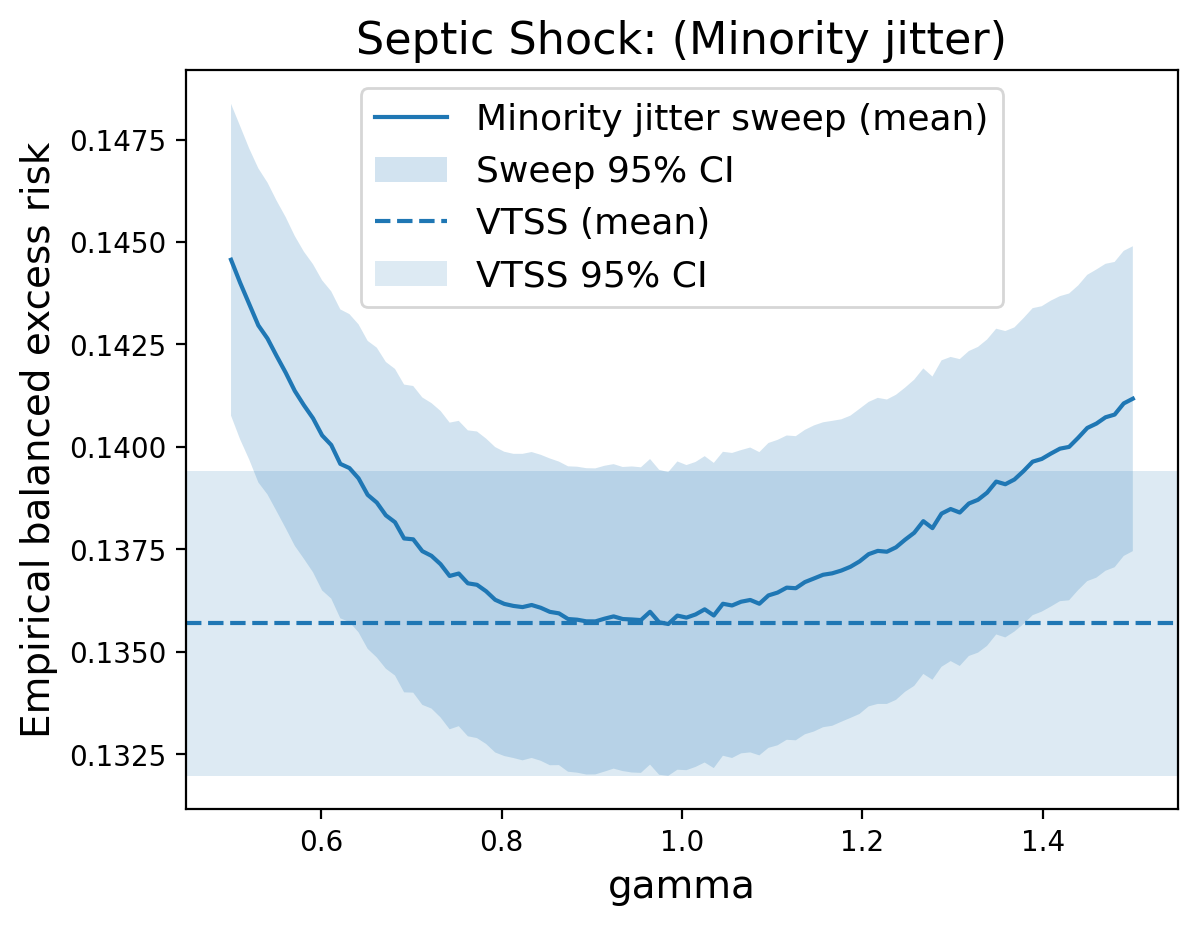}
  \end{subfigure}

  \par\medskip
\begin{minipage}[t]{0.32\textwidth}
  \centering
  \footnotesize
  \textbf{Raw-Mortality} \\
  $0.1463~[0.1428,0.1497]$
\end{minipage}\hfill
\begin{minipage}[t]{0.32\textwidth}
  \centering
  \footnotesize
  \textbf{Raw-Sepsis}\\
  $0.2138~[0.2091,0.2185]$
\end{minipage}\hfill
\begin{minipage}[t]{0.32\textwidth}
  \centering
  \footnotesize
  \textbf{Raw-Septic shock}\\
   $0.2478~[0.2427,0.2529]$
\end{minipage}

  \caption{Excess risk in predicting hospital mortality, sepsis and septic shock across four synthetic generators using SVM. Bottom row is the excess risk obtained by training with raw imbalanced data without synthetic data augmentation.}
  \label{fig:realdataexcessrisksvm}
\end{figure}

The results for using SVM as classifier are summarized in Figure \ref{fig:realdataexcessrisksvm}.

\section{Details of Examples}\label{sec:examples}
In this section, we present the detailed calculations of the examples in the main paper.

\subsection{Details of Example \ref{ex:collinear-log}}
\paragraph*{Toy Collinear Example:} The details of this example are split into following cases:
     \begin{itemize}
         \item \textbf{(Same Direction)} When $s=+1$, the direction of $\nabla\psi(\btheta^*)$ and $\nabla\phi(\btheta^*)$ are the same. Then we select $\delta_n$ as 
         \[
         \delta_n=\frac{2}{\sqrt{\log n_1}-2}\quad \Rightarrow \quad\tilde{n}=\frac{\sqrt{\log n_1}}{\sqrt{\log n_1}-2}(n_0-n_1).
         \]
         Then we have
         \[
         \bb(\btheta)=\left(\frac{n_0}{n_0+n_1+\tilde{n}}-\frac{1}{2}+\frac{\tilde{n}}{n_0+n_1+\tilde{n}}\frac{1}{\sqrt{\log n_1}}\right)\nabla\phi(\btheta)=0.
         \]
         Hence,
         \[
           T_1=T_2=0, \quad T_3=\Theta_P\left(n_0^{-1}\right)\quad \Rightarrow \quad 
           \cR(\hat{\btheta})-\cR(\btheta^*)=\Theta_P\left( n_0^{-1}\right).
           \]
          
         \item \textbf{(Opposite Direction)} Similarly, when $s=-1$, the direction of $\nabla\psi(\btheta^*)$ and $\nabla\phi(\btheta^*)$ are the opposite. Then we select $\delta_n$ as
         \[
         \delta_n=\frac{-2}{\sqrt{\log n_1}+2}\quad \Rightarrow \quad\tilde{n}=\frac{\sqrt{\log n_1}}{\sqrt{\log n_1}+2}(n_0-n_1).
         \]
         Then we still have $\bb(\btheta^*)=0$ and 
         \[
           T_1=T_2=0, \quad T_3=\Theta_P\left(n_0^{-1}\right)\quad \Rightarrow \quad 
           \cR(\hat{\btheta})-\cR(\btheta^*)=\Theta_P\left( n_0^{-1}\right).
           \]

         \item \textbf{(Exactly Synthesizing $\tilde{n}=n_0-n_1$)} In contrast, consider the above two cases but now simply generate $\tilde{n}=n_0-n_1$ synthetic samples. Then, for the same and opposite direction settings,
         \[
         \bb(\btheta^*)=\left(\pi_0-\frac{1}{2}+\frac{\tilde{n}}{n_0+n_1+\tilde{n}}\frac{s}{\sqrt{\log n_1}}\right)\nabla\phi(\btheta^*)=s\frac{n_0-n_1}{2n_0}\frac{1}{\sqrt{\log n_1}}\nabla\phi(\btheta^*).
         \]
         Therefore, 
         \[
           T_1\asymp (\log n_1)^{-1},\quad T_2=\Theta_P\left((n_0\log n_1)^{-1/2}\right), \quad T_3=\Theta_P\left(n_0^{-1}\right),
           \]
        and thus we have
        \[
        \cR(\hat{\btheta})-\cR(\btheta^*)=\Theta_P\left( (\log n_1)^{-1}\right).
        \]
     \end{itemize}

\subsection{Details of Example \ref{biasedexample}}\label{sec:biasedexampledetail}

\paragraph*{Two-Dimensional Gaussian Example:}

We consider a two dimensional Gaussian case where the majority follows $\bx|y=0\sim\cP_0=\mathcal{N}(0,I_2)$ while the minority $\bx|y=1\sim\cP_1=\mathcal{N}(\bmu_1,I_2)$. The synthetic data are generated from $\tilde{\bx}\sim\cP_{\text{syn}}=\mathcal{N}(\bmu_s,I_2)$. We consider the mean squared loss $\ell(f_{\btheta}(\bx),y)=(\btheta^\top \bx-y)^2$. Generally, when $\bx\sim\mathcal{N}(\bmu,I)$, the risk is
    \[
    \bE_{\mathcal{N}(\mu,I)}\ell(\btheta;\bx,y)=\bE(\btheta^\top \bx-y)^2=\btheta^\top \btheta+(\btheta^\top \bmu-y)^2.
    \]
    Thus the balanced risk
    \[
    \cR(\btheta)=\frac{1}{2}\bE_{\cP_0}\ell(\btheta;\bx,0)+\frac{1}{2}\bE_{\cP_1}\ell(\btheta;\bx,1)=\btheta^\top \btheta+\frac{1}{2}(\btheta^\top \bmu_1-1)^2,
    \]
    \[
    \nabla\cR(\btheta)=2\btheta+(\btheta^\top \bmu_1-1)\bmu_1=0\quad \Rightarrow \quad \btheta^*=(\bmu_1\bmu_1^\top +2I)^{-1}
    \bmu_1.
    \]
    Then,
    \begin{align}\label{eq80}
        \nabla\phi(\btheta)=-2(\btheta^\top \bmu_1-1)\bmu_1,\quad \nabla\psi(\btheta)=2(\btheta^\top \bmu_s-1)\bmu_s-2(\btheta^\top \bmu_1-1)\bmu_1.
    \end{align}
    \begin{itemize}
        \item \textbf{Aligned Synthetic Bias Case.} For the aligned bias case, we select $\bmu_1=(\mu,0)^\top $ and $\bmu_s=(a,0)^\top $, where $a\neq\left\{\mu,\frac{2}{\mu}\right\}$. By equation (\ref{eq80}), two gradients are exactly aligned
        \[
        \nabla\phi(\btheta^*)=\left(\frac{4\mu}{\mu^2+2},0\right)^\top ,\quad \nabla\psi(\btheta^*)=\left(\frac{2(\mu a-2)(a-\mu)}{\mu^2+2},0\right)^\top .
        \]
        By empirical estimation, we have
        \[
        \hat{\btheta}-\tilde{\btheta}=O_P\left((n_0+n_1+\tilde{n})^{-1/2}\right).
        \]
        Here we choose $\tilde{n}$ to enforce $\tilde{\btheta}=\btheta^*$, then the excess risk is consistent by Taylor expansion
        \[
        \cR(\hat{\btheta})-\cR(\btheta^*)=\cR(\hat{\btheta})-\cR(\tilde{\btheta})\lesssim \frac{1}{2}\lambda_{\max}(\nabla^2\cR(\btheta^*))\|\hat{\btheta}-\tilde{\btheta}\|^2=O_P\left((n_0+n_1+\tilde{n})^{-1}\right).
        \]
    
       To solve the corresponding synthetic size, $\tilde{\btheta}=\btheta^*$ implies $\nabla\tilde{\cR}(\btheta^*)=0$. Calculation gives
        \begin{align*}
            \nabla\tilde{\cR}(\btheta^*)&=2\btheta^*+2\pi_1(\btheta^{*T}\bmu_1-1)\bmu_1+2\tilde{\pi}(\btheta^{*T}\bmu_s-1)\bmu_s\\
            &=\left(\frac{2\mu}{\mu^2+2},0\right)^\top +\left(\frac{-4\pi_1\mu}{\mu^2+2},0\right)^\top +\left(\frac{2\tilde{\pi}(\mu a-\mu^2-2)a}{\mu^2+2},0\right)^\top =0.
        \end{align*}
        Solving this gives 
        \[
        \tilde{n}=\frac{\mu(n_0-n_1)}{a(\mu^2+2)-\mu(a^2+1)}.
        \]
        For example, when $\mu=1,a=\frac{1}{2}$, this value is $\tilde{n}=4(n_0-n_1)$, which is far away from the balanced synthetic size $n_0-n_1$.

        \item \textbf{Sub-Optimal Synthetic Size $n_0-n_1$.} In the previous case, if we still choose the balanced synthetic size $\tilde{n}=n_0-n_1$, the excess risk is not consistent anymore. The population risk 
        \[
        \tilde{\cR}(\btheta)=\btheta^\top \btheta+\pi_1(\btheta_1\mu-1)^2+\tilde{\pi}(\btheta_1a-1)^2.
        \]
        Therefore,
        \begin{equation*}
            \nabla\tilde{\cR}(\btheta)=0\Rightarrow\begin{cases}
    \btheta_1+\pi_1(\btheta_1\mu-1)\mu+\tilde{\pi}(\btheta_1a-1)a=0,\\
    \btheta_2=0,
    \end{cases}\Rightarrow\tilde{\btheta}=\left(\frac{\mu\pi_1+a\tilde{\pi}}{1+\pi_1\mu^2+\tilde{\pi}a^2},0\right)^\top 
        \end{equation*}

        Define $\frac{n_1}{n_0}\to \rho$, where $\rho\geq 0$. Therefore, $\pi_0\to \frac{1}{2}$, $\pi_1\to \frac{\rho}{2}$ and $\tilde{\pi}\to \frac{1-\rho}{2}$.
        The Hessian for the risk 
        \[
        \nabla^2\cR(\btheta)=\begin{pmatrix}
            \mu^2+2 & 0\\ 0 &2
        \end{pmatrix}\succeq 2I,
        \]
        thus $\cR(\btheta)$ is 2-strongly convex and thus $\cR(\btheta)-\cR(\btheta^*)\geq \|\btheta-\btheta^*\|^2$. Therefore,
        \[
        \lim\cR(\tilde{\btheta})-\cR(\btheta^*)\geq \lim\|\tilde{\btheta}-\btheta^*\|^2=\left(\frac{\mu\pi_1+a\tilde{\pi}}{1+\pi_1\mu^2+\tilde{\pi}a^2}-\frac{\mu}{\mu^2+2}\right)^2=c>0,
        \]
        here for $a\neq \frac{\mu^2+2}{\mu}$ when $\rho=0$. Also by empirical estimation and Taylor expansion
        \[
        \cR(\hat{\btheta})-\cR(\tilde{\btheta})=O_P\left(\frac{1}{n_0+n_1+\tilde{n}}\right),
        \]
        we have
        \[
        \cR(\hat{\btheta})-\cR(\btheta^*)= \cR(\hat{\btheta})-\cR(\tilde{\btheta})+\cR(\tilde{\btheta})-\cR(\btheta^*)\to c>0
        \]
        in probability.

        \item \textbf{Orthogonal Synthetic Bias.} Construct $\bmu_1=(\mu,0)^\top $ and $\bmu_s=(\mu,\mu)$ with $\mu^2\leq 2$, we have the orthogonal synthetic bias
        \[
        \nabla\phi(\btheta^*)=\left(\frac{4\mu}{\mu^2+2},0\right)^\top ,\quad \nabla\psi(\btheta^*)=\left(0,-\frac{4\mu}{\mu^2+2}\right)^\top .
        \]
        Solving the balanced minimizer
        \begin{align*}
            \nabla\cR(\btheta)=0\Rightarrow \btheta^*=\left(\frac{\mu}{\mu^2+2},0\right)^\top \Rightarrow \cR(\btheta^*)=\frac{1}{\mu^2+2}.
        \end{align*}
        Thus, the excess risk for any $\btheta$ satisfies
        \[
        \cR(\btheta)-\cR(\btheta^*)=\btheta_1^2+\btheta_2^2+\frac{1}{2}(\btheta_1\mu-1)^2-\frac{1}{\mu^2+2}\geq \btheta_2^2.
        \]
        Then for $\tilde{\btheta}$,
        \begin{align*}
            \nabla\tilde{\cR}(\btheta)=0\Rightarrow\begin{cases}
    \btheta_1+\pi_1(\btheta_1\mu-1)\mu+\tilde{\pi}\mu(\mu\btheta_1+\mu\btheta_2-1)=0,\\
    \btheta_2+\tilde{\pi}\mu(\mu\btheta_1+\mu\btheta_2-1)=0,
    \end{cases}\\\Rightarrow\tilde{\btheta}_2=\frac{\tilde{\pi}\mu}{1+\pi_1\mu^2+2\tilde{\pi}\mu^2+\pi_1\pi_2\mu^4}
        \end{align*}
    Thus for any non-diminishing synthetic size ($\lim\tilde{\pi}\geq \tau>0$), we have $\lim \tilde{\btheta}_2\geq c>0$. By similar steps in the last example,
    \[
        \cR(\hat{\btheta})-\cR(\btheta^*)= \cR(\hat{\btheta})-\cR(\tilde{\btheta})+\cR(\tilde{\btheta})-\cR(\btheta^*)\to c>0
        \]
        in probability.

    \item \textbf{Arbitrary Direction.} For any direction $\nabla\psi(\btheta^*)$, we can always make a projection onto $\nabla\phi(\btheta^*)$ and an orthogonal vector. As long as the projection onto the orthogonal vector is $o_P(1)$, we can always find some proper synthetic size to make the excess risk consistent.
    \end{itemize}

\subsection{Details of Example \ref{exa:meanshiftcancel}}\label{meanshiftmodel}

For the mean shift model, we use labels $y\in\{0,1\}$ with class prior 
\[
\bP(y=1)=\pi \quad \bP(y=0)=1-\pi,
\]
allowing $\pi$ to be arbitrarily imbalanced. Let $\bxi\in \mathbb{R}^d$ be a random vector from any distribution such that 1) $\bxi$ has bounded support almost surely and 2) $\bxi$ is mean zero and with full-rank covariance. Conditioned on $y$, features follow symmetric shifts of the common noise distribution $\bxi$: 
\[
\bx|(y=1)\sim \bmu+\bxi\quad \bx |(y=0)\sim -\bmu+\bxi.
\]
We consider the linear classifier with squared loss
\[
f_{\btheta}(\bx)=\btheta^\top \bx\quad \ell(\btheta;\bx,y)=(y-1/2-\btheta^\top \bx)^2.
\]
This model isolates how class imbalance interacts with the geometry of the population risk and provides a clean setting to see when rebalancing via synthetic data is inherently unnecessary. We then have the following theorem.

\begin{proposition}[Local symmetry for mean shift model]\label{meanshiftcancel}
   Under the mean shift model, for any distribution of $\bxi$ satisfying the above conditions, the balanced minimizer has a closed form solution, and majority-minority bias cancels.
   \[
   \btheta^*=\frac{1}{2}\left[\bmu\bmu^\top +\bE[\bxi\bxi^\top ]\right]^{-1}\bmu,\quad \nabla \phi(\btheta^*)=0.
   \]
\end{proposition}

Proposition \ref{meanshiftcancel} shows that, under the mean shift model, the balanced risk minimizer admits a closed form expression and, crucially, the majority–minority bias cancels at the optimum $\nabla \phi(\btheta^*)=0$. This is precisely the local symmetry property discussed earlier. As a result, even under severe imbalance, adding synthetic minority samples cannot improve the estimator’s performance under any realistic generator.

\subsection{Details of Example \ref{exa:sigmoidbernoullicancel}}\label{sec:sigmoidcancel}
For the logistic regression model, we continue to use the label convention $y\in\{0,1\}$. The data are generated from a sigmoid Bernoulli model without intercept:
\begin{equation*}
\bx\sim p(\bx),\quad \mathbb{P}(y=1\mid \bx)=\sigma(\bx^\top \btheta_{\mathrm{true}}), \quad \mathbb{P}(y=0\mid \bx)=1-\sigma(\bx^\top \btheta_{\mathrm{true}}).
\end{equation*}
Under this formulation, the logistic loss can be written as
\[
\ell(\btheta;\bx,y)=\log(1+\exp\{\btheta^\top \bx\})-y\bx^\top\btheta.
\]
Our goal in this subsection is to identify conditions on $p(\bx)$ under which class imbalance does not induce a first-order shift in the balanced risk optimum, implying that synthetic minority augmentation is not expected to help.
\begin{proposition}[Local symmetry for Sigmoid Bernoulli model]\label{Logisticcancel}
    Any data generating distribution $p(\bx)$ satisfying the following condition gives a canceled majority-minority bias
    \begin{equation}\label{logisticcondition}
    \bE_{p(\bx)}\left[\bx\sigma(\btheta_{\mathrm{true}}^\top \bx)\left(1-\sigma(\btheta_{\mathrm{true}}^\top \bx)\right)\right]=0\quad \Rightarrow \quad \nabla\phi(\btheta^*)=0.
    \end{equation}
\end{proposition}

Proposition \ref{Logisticcancel} provides a sufficient condition on the feature distribution $p(\bx)$ under which the majority–minority bias cancels at the balanced risk minimizer, i.e., $\nabla\phi(\btheta^*)=0$. Intuitively, this condition enforces a form of local symmetry: the weighted contribution of features to the gradient through 
$\bE_{p(\bx)}\left[\bx\sigma(\btheta_{\mathrm{true}}^\top \bx)\left(1-\sigma(\btheta_{\mathrm{true}}^\top \bx)\right)\right]=0$ averages out, so changing class proportions does not create a first-order perturbation of the target solution. Consequently, imbalance is not the bottleneck in this regime, and adding synthetic minority samples cannot improve excess risk under realistic generators. To demonstrate that this situation is not pathological, we next construct concrete examples of $p(\bx)$ that satisfy the condition and illustrate how the cancellation can occur in practice.

\begin{example}[One-Dimensional Discrete Case]\label{1Dexample}
    Under the sigmoid Bernoulli logistic regression model, suppose the covariate is generated from the discrete distribution
    \begin{equation}\label{discretedistribution}
        x=\left\{\begin{array}{cc}
           -a & \text{with probability }\alpha;\\
            b & \text{with probability }1-\alpha;
        \end{array}\right.
    \end{equation}
    Here $a,b,\alpha>0$. Let $\btheta_{\mathrm{true}}=c>0$. Then the condition in equation (\ref{logisticcondition}) is
    \begin{equation*}
        \bE_{p(x)}\left[x\sigma(cx)\left(1-\sigma(cx)\right)\right]=-\alpha a\sigma(ca)(1-\sigma(ca))+(1-\alpha)b\sigma(cb)(1-\sigma(cb))=0
    \end{equation*}
    The condition is satisfied as long as the probability $\alpha$ satisfies
    \begin{equation}\label{alphachoice}
    \alpha=\frac{b\sigma(cb)(1-\sigma(cb))}{ a\sigma(ca)(1-\sigma(ca))+b\sigma(cb)(1-\sigma(cb))}.
    \end{equation}
    Also, the imbalance of this problem can have any severity:
    \begin{equation}\label{imbalancetozero}
    \mathbb{P}(y=1)=\bE_{p(x)}\left[\sigma(cx)\right]=\alpha(1-\sigma(ca))+(1-\alpha)\sigma(cb)\mathop{\to}\limits_{a\to\infty}0.
    \end{equation}
    Thus, we can construct any imbalance for this problem by creating constants $a$ and $b$, instead of just a naive balancing case.
\end{example}

\begin{example}[Multivariate Case]\label{exa:nosynsigmoid2}
    Suppose the true parameter is \[
    \btheta_{\mathrm{true}}=c\cdot \boldsymbol{v},\quad \|\boldsymbol{v}\|=1,\quad c>0.
    \]
    And the covariates are generated from the distribution
    \[
    \bx=T\boldsymbol{v}+W, 
    \]
    where $T$ follows the distribution of x in equation (\ref{discretedistribution}) in example \ref{1Dexample}. $W$ is a mean zero random noise of any distribution with bounded support and full-rank covariance in the orthogonal space of $v$ such that $\boldsymbol{v}^\top W=0$ almost surely. Also, $W$ is independent of $T$. Therefore, we have
    \[
    \btheta_{\mathrm{true}}^\top \bx=c\boldsymbol{v}^\top (T\boldsymbol{v}+W)=cT.
    \]
    Then the condition (\ref{logisticcondition}) is
    \begin{align*}
    \bE_{p(\bx)}\left[(T\boldsymbol{v}+W)\sigma(cT)\left(1-\sigma(cT)\right)\right]=\boldsymbol{v}\bE_{p(\bx)}\left[T\sigma(cT)\left(1-\sigma(cT)\right)\right]\\=-\alpha a\sigma(ca)(1-\sigma(ca))+(1-\alpha)b\sigma(cb)(1-\sigma(cb))=0.
    \end{align*}
    Similar to example \ref{1Dexample}, the condition (\ref{logisticcondition}) holds as long as the probability $\alpha$ satisfies equation (\ref{alphachoice}), while the imbalance can still be of any severity because equation (\ref{imbalancetozero}) still holds.
    \end{example}

\section{Proofs}\label{sec:proofs}

\subsection{Proof of Theorem \ref{excess-lowerbound}}
By the first order optimality
\begin{align*}
    \|\nabla\tilde{\cR}(\btheta^*)\|=\|\nabla\tilde{\cR}(\btheta^*)-\nabla\tilde{\cR}(\tilde{\btheta})\|\leq M\|\btheta^*-\tilde{\btheta}\|,
\end{align*}
where the last step is from the M-smoothness of $\tilde{R}(\btheta)$ at $\tilde{\btheta}$. Thus we have the lower bound for parameter
\begin{equation}\label{eq90}
\|\btheta^*-\tilde{\btheta}\|\geq \frac{1}{M} \|\nabla\tilde{\cR}(\btheta^*)\|.
\end{equation}
The balanced risk $\cR(\btheta)$ is $\mu$-strongly convex near the minimizer $\btheta^*$, that is
\[
\cR(\btheta)-\cR(\btheta^*)\geq \frac{\mu}{2}\|\btheta-\btheta^*\|^2.
\]
Combining equation (\ref{eq90}), we have
\[
\cR(\tilde{\btheta})-\cR(\btheta^*)\geq \frac{\mu}{2}\|\tilde{\btheta}-\btheta^*\|^2\geq \frac{\mu}{2M^2}\|\nabla\tilde{\cR}(\btheta^*)\|^2.
\]
Recall the relationship combined with first order optimality
\begin{equation*}
    \nabla\tilde{\cR}(\btheta^*)=\nabla\cR(\btheta^*)+\left(\pi_0-\frac{1}{2}\right)\nabla\phi(\btheta^*)+\tilde{\pi}\nabla\psi(\btheta^*)=\left(\pi_0-\frac{1}{2}\right)\nabla\phi(\btheta^*)+\tilde{\pi}\nabla\psi(\btheta^*).
\end{equation*}
Thus we have
\begin{align}\label{eq91}
\cR(\tilde{\btheta})-\cR(\btheta^*)\geq \frac{\mu}{2M^2}\left\|\left(\pi_0-\frac{1}{2}\right)\nabla\phi(\btheta^*)+\tilde{\pi}\nabla\psi(\btheta^*)\right\|^2.
\end{align}

Next we analyze $\left|\cR(\hat{\btheta})-\cR(\tilde{\btheta})\right|$. We first show that $\hat{\btheta}$ and $\tilde{\btheta}$ are close to each other. For fixed $\epsilon>0$, define the set 
\[
A_\epsilon:=\left\{\btheta\in\Theta:\|\btheta-\tilde{\btheta}\|\geq\epsilon\right\}.
\]
Thus it is easy to verify that
\[
\mathop{\inf}\limits_{\btheta\in A_\epsilon}\left(\tilde{\cR}(\btheta)-\tilde{\cR}(\tilde{\btheta})\right)>0.
\]
Then, define the set 
\begin{equation}\label{eq94}
E_n=\left\{\mathop{\sup}\limits_{\btheta\in \Theta}\left|\hat{\cR}(\btheta)-\tilde{\cR}(\btheta)\right|\leq \mathop{\inf}\limits_{\btheta\in A_\epsilon}\left(\tilde{\cR}(\btheta)-\tilde{\cR}(\tilde{\btheta})\right)/3\right\}.
\end{equation}
By definition, we have
\begin{align}\label{eq93}
    \nonumber&\left|\hat{\cR}(\btheta)-\tilde{\cR}(\btheta)\right|\leq \pi_0\left|\frac{1}{n_0}\sum_{i=1}^{n_0}\ell(f_{\btheta}(\bx_i^{(0)}),0)-\bE_{\cP_0}\ell(\btheta;\bx,0)\right|\\
    &+\pi_1\left|\frac{1}{n_1}\sum_{i=1}^{n_1}\ell(f_{\btheta}(\bx_i^{(1)}),1)-\bE_{\cP_1}\ell(\btheta;\bx,1)\right|+\tilde{\pi}\left|\frac{1}{\tilde{n}}\sum_{i=1}^{\tilde{n}}\ell(f_{\btheta}(\tilde{\bx}_i),1)-\bE_{\cP_{\mathrm{syn}}}\ell(\btheta;\bx,1)\right|.
\end{align}
For the first two terms, we have compact $\btheta$, continuous $\ell$ and bounded loss, thus by Uniform Law of Large Numbers (e.g. Lemma 2.4 in \cite{newey1994large}), the uniform convergence holds that
\[
\mathop{\sup}\limits_{\btheta\in \Theta}\left|\frac{1}{n_0}\sum_{i=1}^{n_0}\ell(\btheta;\bx_i^{(0)},0)-\bE_{\cP_0}\ell(\btheta;\bx,0)\right|\mathop{\to}^P0,
\]
\[
\mathop{\sup}\limits_{\btheta\in \Theta}\left|\frac{1}{n_1}\sum_{i=1}^{n_1}\ell(\btheta;\bx_i^{(1)},1)-\bE_{\cP_1}\ell(\btheta;\bx,0)\right|\mathop{\to}^P0.
\]
For the third term, if the synthetic size $\tilde{n}\to \infty$, we similarly have 
\[
\mathop{\sup}\limits_{\btheta\in \Theta}\left|\frac{1}{\tilde{n}}\sum_{i=1}^{\tilde{n}}\ell(\btheta;\tilde{\bx}_i,1)-\bE_{\cP_{\mathrm{syn}}}\ell(\btheta;\bx,1)\right|\mathop{\to}^P0,
\]
while if the synthetic size is bounded by some large constant $\tilde{n}\leq M$, we have $\tilde{\pi}\to 0$. Combined with the bounded loss, we still have the convergence in probability
\[
\tilde{\pi}\mathop{\sup}\limits_{\btheta\in \Theta}\left|\frac{1}{\tilde{n}}\sum_{i=1}^{\tilde{n}}\ell(\btheta;\tilde{\bx}_i,1)-\bE_{\cP_{\mathrm{syn}}}\ell(\btheta;\bx,1)\right|\mathop{\to}^P0.
\]
Thus combining the three terms and equation (\ref{eq93}), we have the uniform convergence
\[
\mathop{\sup}\limits_{\btheta\in \Theta}\left|\hat{\cR}(\btheta)-\tilde{\cR}(\btheta)\right|\mathop{\to}^P0.
\]
Therefore, the set (\ref{eq94}) satisfies
\[
P(E_n)\to 1,
\]
considering $\mathop{\inf}\limits_{\btheta\in A_\epsilon}\left(\tilde{\cR}(\btheta)-\tilde{\cR}(\tilde{\btheta})\right)/3$ is a fixed positive number.
To show the parameter convergence, suppose for contradiction that on event $E_n$, we have $\hat{\btheta}\in A_\epsilon$. Then by the definition of $E_n$, we have
\begin{align*}
    \hat{\cR}(\hat{\btheta})\geq &  \tilde{\cR}(\hat{\btheta})-\frac{1}{3}\mathop{\inf}\limits_{\btheta\in A_\epsilon}\left(\tilde{\cR}(\btheta)-\tilde{\cR}(\tilde{\btheta})\right)=\tilde{\cR}(\hat{\btheta})-\tilde{\cR}(\tilde{\btheta})+\tilde{\cR}(\tilde{\btheta})-\frac{1}{3}\mathop{\inf}\limits_{\btheta\in A_\epsilon}\left(\tilde{\cR}(\btheta)-\tilde{\cR}(\tilde{\btheta})\right)\\
    \geq & \tilde{\cR}(\tilde{\btheta})+\frac{2}{3}\mathop{\inf}\limits_{\btheta\in A_\epsilon}\left(\tilde{\cR}(\btheta)-\tilde{\cR}(\tilde{\btheta})\right).
\end{align*}
Also by definition,
\[
\hat{\cR}(\tilde{\btheta})\leq\tilde{\cR}(\tilde{\btheta})+\frac{1}{3}\mathop{\inf}\limits_{\btheta\in A_\epsilon}\left(\tilde{\cR}(\btheta)-\tilde{\cR}(\tilde{\btheta})\right).
\]
Then we conclude $\hat{\cR}(\hat{\btheta})>\hat{\cR}(\tilde{\btheta})$, which contradicts the fact that $\hat{\btheta}$ is the minimizer of $\hat{\cR}(\btheta)$. Therefore, under event $E_n$, we have $\hat{\btheta}\notin A_\epsilon$. That is $\|\hat{\btheta}-\tilde{\btheta}\|\leq \epsilon$ on $E_n$. Thus
\[
\bP(\|\hat{\btheta}-\tilde{\btheta}\|>\epsilon)\leq \bP(E_n^c)\to 0.
\]
That is the parameter convergence 
\[\|\hat{\btheta}-\tilde{\btheta}\|\mathop{\to}^P0.\]
Considering $\cR(\btheta)$ is continuous, for any $\eta>0$, $\exists\delta>0$ s.t. for $\|\btheta-\tilde{\btheta}\|\leq \delta$ we have $|\cR(\btheta)-\cR(\tilde{\btheta})|\leq\eta$. Thus
\[
\bP(|\cR(\hat{\btheta})-\cR(\tilde{\btheta})|>\eta)\leq \bP(\|\hat{\btheta}-\tilde{\btheta}\|> \delta)\to 0.
\]
That is,
\begin{align}\label{eq92}
\left|\cR(\hat{\btheta})-\cR(\tilde{\btheta})\right|=o_P(1).
\end{align}
Thus, combining equations (\ref{eq91}) and (\ref{eq92}), we have
\begin{align*}
    \cR(\hat{\btheta})-\cR(\btheta^*)&=\cR(\hat{\btheta})-\cR(\tilde{\btheta})+\cR(\tilde{\btheta})-\cR(\btheta^*)\\
    &\geq \frac{\mu}{2M^2}\left\|\left(\pi_0-\frac{1}{2}\right)\nabla\phi(\btheta^*)+\tilde{\pi}\nabla\psi(\btheta^*)\right\|^2+o_P(1).
\end{align*}

\subsection{Proof of Theorem \ref{representation}}
    We start by making a decomposition to the excess risk:
    \begin{align}\label{eq3}
    \cR(\hat{\btheta})-\cR(\btheta^*)&=\left(\cR(\hat{\btheta})-\cR(\tilde{\btheta})\right)+\left(\cR(\tilde{\btheta})-\cR(\btheta^*)\right).
    \end{align}
    
    We first give an upper bound for the population risk $\cR(\tilde{\btheta})-\cR(\btheta^*)$. For function $F(\bx)$ that is differentiable and has minimizer $\bx^*$, satisfying
    \[
    F(\bx^*)\geq F(\bx)+\nabla F(\bx)^\top (\bx^*-\bx)+\frac{\mu}{2}\|\bx^*-\bx\|^2,
    \]
    we have
    \begin{align}\label{eq1}
    \nonumber F(\bx)-F(\bx^*)&\leq \nabla F(\bx)^\top (\bx-\bx^*)-\frac{\mu}{2}\|\bx-\bx^*\|^2\\
    &\leq \frac{1}{2\mu}\|\nabla F(\bx)\|^2+\frac{\mu}{2}\|\bx-\bx^*\|^2-\frac{\mu}{2}\|\bx-\bx^*\|^2=\frac{1}{2\mu}\|\nabla F(\bx)\|^2,
    \end{align}
    where the second inequality is from the fact that $0\leq \|\frac{1}{\sqrt{\mu}}\nabla F(\bx)-\sqrt{\mu}(\bx-\bx^*)\|^2$.

    Applying equation (\ref{eq1}) on the balanced risk $\cR$ gives
    \begin{equation}\label{eq2}
    \cR(\tilde{\btheta})-\cR(\btheta^*)\leq \frac{1}{2\mu}\|\nabla \cR(\tilde{\btheta})\|^2.
    \end{equation}
    For the synthetic risk, we can write it into the form
    \begin{align}
\nonumber\tilde{\cR}(\btheta)=\cR(\btheta)+\left(\pi_0-\frac{1}{2}\right)\phi(\btheta)+\tilde{\pi}\psi(\btheta),
\end{align}
    Take derivative to both sides gives 
    \[
    \nabla\tilde{\cR}(\btheta)=\nabla\cR(\btheta)+\left(\pi_0-\frac{1}{2}\right)\nabla\phi(\btheta)+\tilde{\pi}\nabla\psi(\btheta).
    \]
    Considering $\tilde{\btheta}$ is the minimizer of $\tilde{\cR}(\btheta)$, plugging in and by first order optimality,
    \[
    0=\nabla\tilde{\cR}(\tilde{\btheta})=\nabla\cR(\tilde{\btheta})+\left(\pi_0-\frac{1}{2}\right)\nabla\phi(\tilde{\btheta})+\tilde{\pi}\nabla\psi(\tilde{\btheta}).
    \]
    Take norms 
    \begin{equation}\label{R0gradbound}
    \|\nabla\cR(\tilde{\btheta})\|=\left\|\left(\pi_0-\frac{1}{2}\right)\nabla\phi(\tilde{\btheta})+\tilde{\pi}\nabla\psi(\tilde{\btheta})\right\|.
    \end{equation}
    Combining equation (\ref{eq2}), we have the bound
    \begin{equation}\label{eq8}
    \cR(\tilde{\btheta})-\cR(\btheta^*)\leq\frac{1}{2\mu}\left\|\left(\pi_0-\frac{1}{2}\right)\nabla\phi(\tilde{\btheta})+\tilde{\pi}\nabla\psi(\tilde{\btheta})\right\|.
    \end{equation}
    Further by the quadratic growth assumption, it holds that
    \begin{equation}\label{tilde-star}
    \|\tilde{\btheta}-\btheta^*\|\leq \sqrt{\frac{1}{2\mu L}}\left\|\left(\pi_0-\frac{1}{2}\right)\nabla\phi(\tilde{\btheta})+\tilde{\pi}\nabla\psi(\tilde{\btheta})\right\|
    \end{equation}

    With this upper bound, we then give a representation for $\tilde{\btheta}-\btheta^*$. Take derivative at $\tilde{\btheta}$ for equation (\ref{psinotation}) gives
    \begin{equation}\label{eq9}
    0=\nabla\tilde{\cR}(\tilde{\btheta})=\nabla\cR(\tilde{\btheta})+\left(\pi_0-\frac{1}{2}\right)\nabla\phi(\tilde{\btheta})+\tilde{\pi}\nabla\psi(\tilde{\btheta}).
    \end{equation}
    Standard Taylor expansion gives
    \begin{align}\label{eq10}
        0=\nabla\tilde{\cR}(\tilde{\theta})=\nabla\tilde{\cR}(\theta^*)+\nabla^2\tilde{\cR}(\theta^*)(\tilde{\theta}-\theta^*)+r_1,
    \end{align}
    where the remainder has the integral form
    \begin{align*}
        r_1=\int_0^1\left(\nabla^2\tilde{\cR}(\theta^*+t(\tilde{\theta}-\theta^*))-\nabla^2\tilde{\cR}(\theta^*)\right)^\top (\tilde{\theta}-\theta^*)dt.
    \end{align*}
    By Assumption (A3), we have the Lipschitzness
    \begin{align*}
        \left\|\nabla^2\tilde{\cR}(\theta^*+t(\tilde{\theta}-\theta^*))-\nabla^2\tilde{\cR}(\theta^*)\right\|\leq L\|\tilde{\theta}-\theta^*\|.
    \end{align*}
    Thus we have the remainder term satisfying
    \begin{align*}
        \|r_1\|=O\left(\|\tilde{\theta}-\theta^*\|\right).
    \end{align*}
    Recall the expression in equation (\ref{eq:decomp_compact}). Taking derivative gives
    \begin{align*}
        \nabla\tilde{\cR}(\theta)=\nabla\cR(\theta)+b(\theta).
    \end{align*}
    Taking $\theta^*$ and by the first order optimality, we have
    \begin{align*}
        \nabla\tilde{\cR}(\theta^*)=\nabla\cR(\theta^*)+b(\theta^*)=b(\theta^*).
    \end{align*}
    Also, taking second derivative to equation (\ref{eq:decomp_compact}) and at $\theta^*$, we have
    \begin{align*}
        \nabla^2\tilde{\cR}(\theta^*)=\nabla^2\cR(\theta^*)+\nabla b(\theta^*).
    \end{align*}
    Plugging back in equation (\ref{eq10}), we have
    \begin{align*}
        0=b(\theta^*)+\left(\nabla^2\cR(\theta^*)+\nabla b(\theta^*)\right)(\tilde{\theta}-\theta^*)+r_1.
    \end{align*}
    Solving the equation gives a representation
    \begin{align}\label{alpha-presentation}
    \nonumber&\tilde{\btheta}-\btheta^*\\&=-\left[\nabla^2\cR(\btheta^*)+\left(\pi_0-\frac{1}{2}\right)\nabla^2\phi(\btheta^*)+\tilde{\pi}\nabla^2\psi(\btheta^*)\right]^{-1}\left[\left(\pi_0-\frac{1}{2}\right)\nabla\phi(\btheta^*)+\tilde{\pi}\nabla\psi(\btheta^*)\right]+\br,
    \end{align}
    where $\|\br\|=O(\|\tilde{\btheta}-\btheta^*\|^2)=O\left(\left\|\left(\pi_0-\frac{1}{2}\right)\nabla\phi(\tilde{\btheta})+\tilde{\pi}\nabla\psi(\tilde{\btheta})\right\|^2\right)$ because the smallest eigenvalue of $\nabla^2\cR(\btheta^*)+\left(\pi_0-\frac{1}{2}\right)\nabla^2\phi(\btheta^*)+\tilde{\pi}\nabla^2\psi(\btheta^*)$ is lower bounded by assumption $\nabla^2\tilde{\cR}(\btheta)\succeq \lambda I$.

For the second term in equation (\ref{eq3}), standard Taylor expansion gives
\begin{align}\label{eq5}
    \cR(\tilde{\btheta})-\cR(\btheta^*)=\nabla\cR(\btheta^*)(\tilde{\btheta}-\btheta^*)+\frac{1}{2}(\tilde{\btheta}-\btheta^*)^\top \nabla^2\cR(\btheta^*)(\tilde{\btheta}-\btheta^*)+O(\|\tilde{\btheta}-\btheta^*\|^3)
\end{align}
Combining equation (\ref{eq5}) and the representation in equation (\ref{alpha-presentation}) gives a representation for the second term in equation (\ref{eq3})
\begin{align}\label{eq6}
    \nonumber\cR(\tilde{\btheta})-\cR(\btheta^*)=\frac{1}{2}\bb(\btheta^*)^\top \left(\nabla^2\cR(\btheta^*)+\nabla \bb(\btheta^*)\right)^{-1}\nabla^2\cR(\btheta^*)\left(\nabla^2\cR(\btheta^*)+\nabla \bb(\btheta^*)\right)^{-1} \bb(\btheta^*)\\
    +O_P\left(\left\|\left(\pi_0-\frac{1}{2}\right)\nabla\phi(\tilde{\btheta})+\tilde{\pi}\nabla\psi(\tilde{\btheta})\right\|^3\right)
\end{align}

To give a representation for the first term in equation (\ref{eq3}), we need a representation for $\hat{\btheta}-\tilde{\btheta}$. We first give an upper bound for $\|\hat{\btheta}-\tilde{\btheta}\|$. By first order optimality and Taylor expansion, 
\[
0=\nabla\hat{\cR}(\hat{\btheta})=\nabla\hat{\cR}(\tilde{\btheta})+\nabla^2\hat{\cR}(\tilde{\btheta})(\hat{\btheta}-\tilde{\btheta})+\br_4,
\]
where the remainder term satisfies $\|\br_4\|=O(\|\hat{\btheta}-\tilde{\btheta}\|^2)$. By matrix concentration, we have the minimal eigenvalue of Hessian matrix $\left|\lambda_{\min}(\nabla^2\hat{\cR}(\btheta))-\lambda_{\min}(\nabla^2\tilde{\cR}(\btheta))\right|\to 0$ in probability and $\nabla^2\tilde{\cR}(\btheta)\succeq \lambda I$ by assumption. Thus, for large enough $n_0$ and $n_1$, we have $\lambda_{\min}(\nabla^2\hat{\cR}(\tilde{\btheta}))>\frac{\lambda}{c}>0$ for some constant $c>0$ in probability. Therefore $\nabla^2\hat{\cR}(\tilde{\btheta})$ is invertible and the norm satisfies $\left\|\left[\nabla^2\hat{\cR}(\tilde{\btheta})\right]^{-1}\right\|\leq \frac{c}{\lambda}$. That is,
\begin{equation}\label{alphahat-alpha}
\hat{\btheta}-\tilde{\btheta}=-\left[\nabla^2\hat{\cR}(\tilde{\btheta})\right]^{-1}\nabla\hat{\cR}(\tilde{\btheta})+\br_5,
\end{equation}
where $\|\br_5\|=O(\|\hat{\btheta}-\tilde{\btheta}\|^2)$.

By optimality, $\bE\nabla\hat{\cR}(\tilde{\btheta})=\nabla \tilde{\cR}(\tilde{\btheta})=0$. Therefore,
\begin{align}\label{eq13}
    \nonumber&\left\|\nabla\hat{\cR}(\tilde{\btheta})\right\|=\left\|\frac{\pi_0}{n_0}\sum_{i=1}^{n_0}\nabla\ell(\btheta;\bx_i,0)+\frac{\pi_1}{n_1}\sum_{i=1}^{n_1}\nabla\ell(\btheta;\bx_i,1)+\frac{\tilde{\pi}}{\tilde{n}}\sum_{i=1}^{\tilde{n}}\nabla\ell(\btheta;\tilde{\bx}_i,1)-\bE\nabla\hat{\cR}(\tilde{\btheta})\right\|\\
    \nonumber&\leq\pi_0\left\|\frac{1}{n_0}\sum_{i=1}^{n_0}\nabla\ell(\btheta;\bx_i,0)-\bE_{\cP_0}\left[\nabla\ell(\btheta;\bx,0)\right]\right\|+\pi_1\left\|\frac{1}{n_1}\sum_{i=1}^{n_1}\nabla\ell(\btheta;\bx_i,1)-\bE_{\cP_1}\left[\nabla\ell(\btheta;\bx,1)\right]\right\|\\
    &+\tilde{\pi}\left\|\frac{1}{\tilde{n}}\sum_{i=1}^{\tilde{n}}\nabla\ell(\btheta;\tilde{\bx}_i,1)-\bE_{\cP_{\text{syn}}}\left[\nabla\ell(\btheta;\tilde{\bx},1)\right]\right\|
\end{align}

By union bound and Hoeffding's inequality, for fixed $\theta$,
\begin{align}\label{eq12}
\nonumber&\mathbb{P}\left(\left\|\frac{1}{n_0}\sum_{i=1}^{n_0}\nabla\ell(\btheta;\bx_i,0)-\bE_{\cP_0}\nabla\ell(\btheta;\bx,0)\right\|>t\right)\\
\nonumber&\leq 
\mathbb{P}\left(\exists i\in|d|,s.t.\left|\left[\frac{1}{n_0}\sum_{i=1}^{n_0}\nabla\ell(\btheta;\bx_i,0)\right]_i-\left[\bE_{\cP_0}\nabla\ell(\btheta;\bx,0)\right]_i\right|>\frac{t}{\sqrt{d}}\right)\\
&\leq \sum_{i=1}^d \mathbb{P}\left(\left|\left[\frac{1}{n_0}\sum_{i=1}^{n_0}\nabla\ell(\btheta;\bx_i,0)\right]_i-\left[\bE_{\cP_0}\nabla\ell(\btheta;\bx,0)\right]_i\right|>\frac{t}{\sqrt{d}}\right)\leq 2d\exp\left\{-\frac{2t^2n_0}{dB}\right\},
\end{align}
where $B>0$ is a constant from bounded support assumption. Thus
\begin{align}
\nonumber&\mathbb{P}\left(\pi_0\left\|\frac{1}{n_0}\sum_{i=1}^{n_0}\nabla\ell(\btheta;\bx_i,0)-\bE_{\cP_0}\nabla\ell(\btheta;\bx,0)\right\|>t\right)\\
&=\mathbb{P}\left(\left\|\frac{1}{n_0}\sum_{i=1}^{n_0}\nabla\ell(\btheta;\bx_i,0)-\bE_{\cP_0}\nabla\ell(\btheta;\bx,0)\right\|>\frac{t}{\pi_0}\right)\leq 2d\exp\left\{-\frac{2t^2(n_0+n_1+\tilde{n})^2}{dBn_0}\right\}.
\end{align}
Therefore, with probability at least $1-\delta$, 
\[
\pi_0\left\|\frac{1}{n_0}\sum_{i=1}^{n_0}\nabla\ell(\btheta;\bx_i,0)-\bE_{\cP_0}\nabla\ell(\btheta;\bx,0)\right\|\leq \sqrt{\frac{Bd\log(2d/\delta)}{2}}\frac{\sqrt{n_0}}{n_0+n_1+\tilde{n}}.
\]
That is, we have
\begin{equation}\label{eq14}
\pi_0\left\|\frac{1}{n_0}\sum_{i=1}^{n_0}\nabla\ell(\btheta;\bx_i,0)-\bE_{\cP_0}\nabla\ell(\btheta;\bx,0)\right\|=O_P\left(\frac{\sqrt{n_0}}{n_0+n_1+\tilde{n}}\right).
\end{equation}
By essentially the same procedure with equation (\ref{eq12}), one can have
\[
\mathbb{P}\left(\pi_1\left\|\frac{1}{n_1}\sum_{i=1}^{n_1}\nabla\ell(\btheta;\bx_i,1)-\bE_{\cP_1}\nabla\ell(\btheta;\bx,1)\right\|>t\right)\leq 2d\exp\left\{-\frac{2t^2(n_0+n_1+\tilde{n})^2}{Bdn_1}\right\},
\]
\[
\mathbb{P}\left(\tilde{\pi}\left\|\frac{1}{\tilde    n}\sum_{i=1}^{\tilde{n}}\nabla\ell(\btheta;\tilde{\bx}_i,1)-\bE_{\cP_{\text{syn}}}\nabla\ell(\btheta;\tilde{\bx},1)\right\|>t\right)\leq 2d\exp\left\{-\frac{2t^2(n_0+n_1+\tilde{n})^2}{Bd\tilde{n}}\right\},
\]
that is
\begin{equation}\label{eq141}
\pi_1\left\|\frac{1}{n_1}\sum_{i=1}^{n_1}\nabla\ell(\btheta;\bx_i,1)-\bE_{\cP_1}\nabla\ell(\btheta;\bx,1)\right\|=O_P\left(\frac{\sqrt{n_1}}{n_0+n_1+\tilde{n}}\right).
\end{equation}
\begin{equation}\label{eq142}
\tilde{\pi}\left\|\frac{1}{\tilde{n}}\sum_{i=1}^{\tilde{n}}\nabla\ell(\btheta;\tilde{\bx}_i,1)-\bE_{\cP_{\text{syn}}}\nabla\ell(\btheta;\tilde{\bx},1)\right\|=O_P\left(\frac{\sqrt{\tilde{n}}}{n_0+n_1+\tilde{n}}\right).
\end{equation}
Combining equation (\ref{eq12}), (\ref{eq14}), (\ref{eq141}) and (\ref{eq142}), we have 
\[
\left\|\nabla\hat{\cR}(\tilde{\btheta})\right\|=O_P\left(\frac{1}{\sqrt{n_0+n_1+\tilde{n}}}\right)
\]
Also equation (\ref{alphahat-alpha}) gives
\begin{equation}\label{alphahat-alpha-solve}
    \|\hat{\btheta}-\tilde{\btheta}\|\leq \left\|\left[\nabla^2\hat{\cR}(\tilde{\btheta})\right]^{-1}\right\|\left\|\nabla\hat{\cR}(\tilde{\btheta})\right\|+c\|\hat{\btheta}-\tilde{\btheta}\|^2
\end{equation}
Therefore, solving equation (\ref{alphahat-alpha-solve}), we have
\begin{equation}\label{alphahat-alpharate}
    \|\hat{\btheta}-\tilde{\btheta}\|=O_P\left(\frac{1}{\sqrt{n_0+n_1+\tilde{n}}}\right).
\end{equation}

To give a representation for $\hat{\btheta}-\tilde{\btheta}$, we need to study the expansion of probability density functions for sample average. Let $\bx_i,i=1,...,n$ be i.i.d.  $d$-dimensional random vectors with finite mean $\bmu$, positive definite covariance matrix $\Sigma$ and bounded support. Define \begin{equation}\label{eq25}
W_n=\sqrt{n}\Sigma^{-\frac{1}{2}}\left(\bar{\bx}_n-\bmu\right),~~\bar{\bx}_n=\frac{1}{n}\sum_{i=1}^n\bx_i.
\end{equation}
Denote the probability density function of $W_n$ as $f_n(\bx)$ for $\bx\in\mathbb{R}^d$, and denote the density of $d$-dimensional standard Gaussian random vector $\mathcal{N}(0,I_d)$ as $\phi_d(\bx)$.
By standard multivariate Berry-Esseen local limit theorem (for example \cite{bobkov2024berry}), we have the uniform bound on the density
\begin{equation}\label{berryesseen}
\mathop{\sup}\limits_{\bx}\left|f_n(\bx)-\phi_d(\bx)\right|\lesssim \frac{1}{\sqrt{n}}
\end{equation}
For notational simplicity, we define $b_n:=\frac{c}{\sqrt{n}}$ with a constant $c>0$ such that $\mathop{\sup}\limits_{\bx}\left|f_n(\bx)-\phi_d(\bx)\right|\leq \frac{c}{\sqrt{n}}$, and two sets
\[
A_n=\left\{\bx\in\mathbb{R}^d:\phi_d(\bx)\geq 2b_n\right\}, \quad B_n:=A_n^c
=\left\{\bx\in\mathbb{R}^d:\phi_d(\bx)< 2b_n\right\}.\]
Also, we define
\[
h_n:=f_n-\phi_d,\quad h_n^+=\max(f_n-\phi_d,0), \quad h_n^-=\max(\phi_d-f_n,0).
\]
Therefore, simply we have
\[
\int f_n=\int\phi_d=1~\Rightarrow~ \int h_n=0.
\]
Also, we have
\[
h_n=h_n^+-h_n^-,~\text{and }~|h_n|=h_n^++h_n^-. 
\]
Combining above we also have
\[
\int h_n^+-h_n^-=0~\Rightarrow ~
\int h_n^+=\int h_n^-.
\]
Therefore, we have the total variation satisfying
\begin{align}\label{TVdecomposition}
    TV(f_n,\phi_d)&=\frac{1}{2}\int |h_n|=\frac{1}{2}\int h_n^++\frac{1}{2}\int h_n^-=\int h_n^-=\int_{A_n}h_n^-+\int_{B_n}h_n^-.
\end{align}
Under the set $A_n$, by the Berry-Esseen bound (\ref{berryesseen}), which is uniform in $x$, we have $h_n^-\leq b_n$. Under the set $b_n$, considering two distributions are positive, we have $h_n^-\leq \phi_d$. That is
\begin{equation}\label{hn-split}
h_n^-(x)\leq \left\{\begin{aligned}
    b_n \quad \bx\in A_n\\
    \phi_d \quad \bx\in B_n
\end{aligned}\right.
\end{equation}
Combining equations (\ref{TVdecomposition}) and (\ref{hn-split}), the total variation can be upper bounded by
\begin{equation}\label{TVupperdecompose}
TV(f_n,\phi_d)\leq b_n|A_n|+\int_{B_n}\phi_d.
\end{equation}
Therefore, we can study the two terms in equation (\ref{TVupperdecompose}) separately. For the first term, we only need to derive the volume $|A_n|$. By the definition of $A_n$, we solve the equation
\[
\phi_d(\bx)=(2\pi)^{-\frac{d}{2}}\exp\left\{-\frac{\|\bx\|^2}{2}\right\}\geq 2b_n=\frac{2c}{\sqrt{n}}.
\]
The solution of the equation satisfies $\|\bx\|^2\leq \log n-\log(2\pi)^{d/2}c$, which is positive for large enough $n$. Therefore, the set $A_n$ satisfies that
\begin{equation}\label{bnAnbound}
|A_n|=O\left((\log n)^{\frac{d}{2}}\right)~\Rightarrow~b_n|A_n|=O\left(\frac{(\log n)^{\frac{d}{2}}}{\sqrt{n}}\right)\to 0.
\end{equation}
For the second term in equation (\ref{TVupperdecompose}), we have
\[
\int_{B_n}\phi_d(\bx)d\bx=\int_{\|\bx\|^2\geq \log n-\log(2\pi)^{d/2}c}\phi_d(\bx)d\bx.
\]
By standard tail bound for multivariate Gaussian (e.g. $\int_{\|\bx\|>r}\phi_d(\bx)d\bx\leq c_dr^{d-2}e^{-\frac{r^2}{2}}$ for large enough $r$), we have
\begin{equation}\label{gaussiantail}
\int_{B_n}\phi_d(\bx)d\bx\leq c\frac{(\log n)^{\frac{d-2}{2}}}{\sqrt{n}}\to0.
\end{equation}
Combining equations (\ref{TVupperdecompose}), (\ref{bnAnbound}) and (\ref{gaussiantail}) gives the convergence of total variation distance 
\[
TV(f_n,\phi_d)\leq c\frac{(\log n)^{\frac{d}{2}}}{\sqrt{n}}+c\frac{(\log n)^{\frac{d-2}{2}}}{\sqrt{n}}\to 0.
\]
Therefore, by maximal coupling principle, there exists a coupling between $W_n$ and $Z$ such that
\[
\mathbb{P}(W_n\neq Z)=TV(f_n,\phi_d).
\]
That is
\[
\mathbb{P}\left(\|W_n-Z\|>\frac{c}{n}\right)\leq TV(f_n,\phi_d)\to 0.
\]
That is, we can write
\[
W_n=Z+\br,\quad \|\br\|=O_P\left(\frac{1}{n}\right),
\]
for some $Z$ having marginal distribution $\phi_d$. Recall the definition of $W_n$, we solve for the sample average as
\begin{equation}\label{multivariate-berry-representation}
    \bar{\bx}_n=\bmu+\frac{1}{\sqrt{n}}\Sigma^{\frac{1}{2}}Z+\br,\quad \|\br\|=O_P\left(\frac{1}{n}\right).
\end{equation}

Now we are ready to give a representation for $\hat{\btheta}-\tilde{\btheta}$. By optimality and Taylor expansion
\[
0=\nabla\hat{\cR}(\hat{\btheta})=\nabla\hat{\cR}(\tilde{\btheta})+\nabla^2\hat{\cR}(\tilde{\btheta})(\hat{\btheta}-\tilde{\btheta})+O(\|\hat{\btheta}-\tilde{\btheta}\|^2)
\]
By the convergence of sample mean to expectation and applying equation (\ref{multivariate-berry-representation}),
\begin{align}
    \nonumber&0=\nabla\tilde{\cR}(\tilde{\btheta})+\frac{1}{n_0+n_1+\tilde{n}}\left[\sqrt{n_0}\Sigma_0^{\frac{1}{2}}Z_1+\sqrt{n_1}\Sigma_1^{\frac{1}{2}}Z_2+\sqrt{\tilde{n}}\tilde{\Sigma}^{\frac{1}{2}}Z_3\right]+O_P\left(\frac{1}{n_0+n_1+\tilde{n}}\right)\\
    \nonumber&+\nabla^2\tilde{\cR}(\tilde{\btheta})(\hat{\btheta}-\tilde{\btheta})+O_P\left(\left(\frac{1}{\sqrt{n_0+n_1+\tilde{n}}}\right)\|\hat{\btheta}-\tilde{\btheta}\|\right)+O\left(\|\hat{\btheta}-\tilde{\btheta}\|^2\right)\\
    \nonumber&=\frac{1}{n_0+n_1+\tilde{n}}\left[\sqrt{n_0}\Sigma_0^{\frac{1}{2}}Z_1+\sqrt{n_1}\Sigma_1^{\frac{1}{2}}Z_2+\sqrt{\tilde{n}}\tilde{\Sigma}^{\frac{1}{2}}Z_3\right]+\nabla^2\tilde{\cR}(\tilde{\btheta})(\hat{\btheta}-\tilde{\btheta})+O_P\left(\frac{1}{n_0+n_1+\tilde{n}}\right),
\end{align}
where $Z_1$, $Z_2$ and $Z_3$ are independent $d$-dimensional standard Gaussian random vectors. Considering these are three independent Gaussians, we can combine them into one Gaussian, that is
\begin{equation}
    0=\frac{1}{n_0+n_1+\tilde{n}}\left(n_0\Sigma_0+n_1\Sigma_1+\tilde{n}\tilde{\Sigma}\right)^{\frac{1}{2}}Z+\nabla^2\tilde{\cR}(\tilde{\btheta})(\hat{\btheta}-\tilde{\btheta})+O_P\left(\frac{1}{n_0+n_1+\tilde{n}}\right),
\end{equation}
where the Gaussian $Z\sim\mathcal{N}(0,I_d)$. 
Solving for $\hat{\btheta}-\tilde{\btheta}$ gives the representation
\begin{align}\label{eq16}
    \hat{\btheta}-\tilde{\btheta}=-\frac{1}{n_0+n_1+\tilde{n}}\left[\nabla^2\tilde{\cR}(\tilde{\btheta})\right]^{-1}\left(n_0\Sigma_0+n_1\Sigma_1+\tilde{n}\tilde{\Sigma}\right)^{\frac{1}{2}}Z+O_P\left(\frac{1}{n_0+n_1+\tilde{n}}\right).
\end{align}
Taylor expansion, combining equation (\ref{R0gradbound}) and (\ref{eq16}) gives a representation for the first term in equation (\ref{eq3}),
\begin{align}\label{eq17}
    \nonumber\cR(\hat{\btheta})-\cR(\tilde{\btheta})&=\nabla\cR(\tilde{\btheta})^\top (\hat{\btheta}-\tilde{\btheta})+\frac{1}{2}(\hat{\btheta}-\tilde{\btheta})^\top \nabla^2\cR(\tilde{\btheta})(\hat{\btheta}-\tilde{\btheta})+O(\|\hat{\btheta}-\tilde{\btheta}\|^3)\\
    \nonumber&=\frac{1}{\sqrt{n_0+n_1+\tilde{n}}}\nabla\cR(\tilde{\btheta})^\top \left[\nabla^2\tilde{\cR}(\tilde{\btheta})\right]^{-1}\Sigma^{\frac{1}{2}}Z\\
    \nonumber&\quad+\frac{1}{2(n_0+n_1+\tilde{n})}Z^\top \Sigma^{\frac{1}{2}}\left[\nabla^2\tilde{\cR}(\tilde{\btheta})\right]^{-1}\nabla^2\cR(\tilde{\btheta})\left[\nabla^2\tilde{\cR}(\tilde{\btheta})\right]^{-1}\Sigma^{\frac{1}{2}}Z\\
    &\quad+O_P\left((n_0+n_1+\tilde{n})^{-\frac{3}{2}}+\frac{\left\|\bb(\tilde{\btheta})\right\|}{n_0+n_1+\tilde{n}}\right),
\end{align}
where \[
\Sigma=\frac{n_0\Sigma_0+n_1\Sigma_1+\tilde{n}\tilde{\Sigma}}{n_0+n_1+\tilde{n}}.
\]

Combining equations (\ref{eq3}), (\ref{eq6}) and (\ref{eq17}) gives
\begin{align}\label{representation-with-tilde}\nonumber&\cR(\hat{\btheta})-\cR(\btheta^*)=\frac{1}{2}\bb(\btheta^*)^\top \left(\nabla^2\cR(\btheta^*)+\nabla \bb(\btheta^*)\right)^{-1}\nabla^2\cR(\btheta^*)\left(\nabla^2\cR(\btheta^*)+\nabla \bb(\btheta^*)\right)^{-1} \bb(\btheta^*)\\
    \nonumber&\quad+\frac{1}{n_0+n_1+\tilde{n}}\nabla\cR(\tilde{\btheta})^\top \left[\nabla^2\tilde{\cR}(\tilde{\btheta})\right]^{-1}\left(n_0\Sigma_0(\tilde{\btheta})+n_1\Sigma_1(\tilde{\btheta})+\tilde{n}\tilde{\Sigma}(\tilde{\btheta})\right)^{\frac{1}{2}}Z\\
   \nonumber &\quad+\frac{1}{2(n_0+n_1+\tilde{n})^2}Z^\top \Sigma(\tilde{\btheta})^{\frac{1}{2}}\left[\nabla^2\tilde{\cR}(\tilde{\btheta})\right]^{-1}\nabla^2\cR(\tilde{\btheta})\left[\nabla^2\tilde{\cR}(\tilde{\btheta})\right]^{-1}\Sigma(\tilde{\btheta})^{\frac{1}{2}}Z\\
    &\quad +O_P\left(\|\bb(\tilde{\btheta})\|+(n_0+n_1+\tilde{n})^{-\frac{3}{2}}+\frac{\|\bb(\tilde{\btheta})\|}{n_0+n_1+\tilde{n}}\right),
    \end{align}

Next, we need to get rid of $\tilde{\btheta}$ in the final representation. For $\nabla\cR(\tilde{\btheta})$, by Taylor expansion,
\[
\nabla\cR(\tilde{\btheta})=\nabla\cR(\btheta^*)+\nabla^2\cR(\btheta^*)(\tilde{\btheta}-\btheta^*)+\br_6=\nabla^2\cR(\btheta^*)(\tilde{\btheta}-\btheta^*)+\br_6,
\]
where $\|\br_6\|=O(\|\tilde{\btheta}-\btheta^*\|^2)$. Then combining equations (\ref{tilde-star}) and (\ref{alpha-presentation}) gives a representation
\begin{align}\label{eq18}
    \nabla\cR(\tilde{\btheta})=-\nabla^2\cR(\btheta^*)\left[\nabla^2\cR(\btheta^*)+\nabla \bb(\btheta^*)\right]^{-1}\left[\left(\pi_0-\frac{1}{2}\right)\nabla\phi(\btheta^*)+\tilde{\pi}\nabla\psi(\btheta^*)\right]+\br,
\end{align}
where $\|\br\|=O\left(\left\|\bb(\tilde{\btheta})\right\|^2\right)$. 

For the term $\left[\nabla^2\tilde{\cR}(\tilde{\btheta})\right]^{-1}$, the inverse satisfies,
\[\left[\nabla^2\tilde{\cR}(\tilde{\btheta})\right]^{-1}-\left[\nabla^2\tilde{\cR}(\btheta^*)\right]^{-1}=\left[\nabla^2\tilde{\cR}(\btheta^*)\right]^{-1}\left[\nabla^2\tilde{\cR}(\btheta^*)-\nabla^2\tilde{\cR}(\tilde{\btheta})\right]\left[\nabla^2\tilde{\cR}(\tilde{\btheta})\right]^{-1},\]
Thus, taking norms,
\begin{align*}
    \left\|\left[\nabla^2\tilde{\cR}(\tilde{\btheta})\right]^{-1}-\left[\nabla^2\tilde{\cR}(\btheta^*)\right]^{-1}\right\|&=\left\|\left[\nabla^2\tilde{\cR}(\btheta^*)\right]^{-1}\right\|\left\|\nabla^2\tilde{\cR}(\btheta^*)-\nabla^2\tilde{\cR}(\tilde{\btheta})\right\|\left\|\left[\nabla^2\tilde{\cR}(\tilde{\btheta})\right]^{-1}\right\|\\
    &\leq \frac{L}{\lambda^2}\|\tilde{\btheta}-\btheta^*\|.
\end{align*}
Therefore, we can write
\begin{equation}\label{eq19}
    \left[\nabla^2\tilde{\cR}(\tilde{\btheta})\right]^{-1}=\left[\nabla^2\tilde{\cR}(\btheta^*)\right]^{-1}+\br=\left[\nabla^2\cR(\btheta^*)+\nabla \bb(\btheta^*)\right]^{-1}+\br,
\end{equation}
where $\|\br\|\leq \frac{L}{\lambda^2}\|\tilde{\btheta}-\btheta^*\|=O\left(\left\|\bb(\tilde{\btheta})\right\|\right)$.

For the term $\left(n_0\Sigma_0(\tilde{\btheta})+n_1\Sigma_1(\tilde{\btheta})+\tilde{n}\tilde{\Sigma}(\tilde{\btheta})\right)^{\frac{1}{2}}$, denote $\Sigma(\btheta)=n_0\Sigma_0(\tilde{\btheta})+n_1\Sigma_1(\tilde{\btheta})+\tilde{n}\tilde{\Sigma}(\tilde{\btheta})$ for brevity. By assumption, $\Sigma(\btheta)\succ \lambda I$ and is Lipschitz. By simple algebra, we have the Sylvester type equation
\begin{align*}
    \Sigma(\tilde{\btheta})^{\frac{1}{2}}(\Sigma(\tilde{\btheta})^{\frac{1}{2}}-\Sigma(\btheta^*)^{\frac{1}{2}})+(\Sigma(\tilde{\btheta})^{\frac{1}{2}}-\Sigma(\btheta^*)^{\frac{1}{2}})\Sigma(\btheta^*)^{\frac{1}{2}}=\Sigma(\tilde{\btheta})-\Sigma(\btheta^*).
\end{align*}
We need to solve this equation to give a representation for $\Sigma(\tilde{\btheta})^{\frac{1}{2}}$. Define 
\[
F(t):=e^{-t\Sigma(\tilde{\btheta})^{1/2}}(\Sigma(\tilde{\btheta})^{\frac{1}{2}}-\Sigma(\btheta^*)^{\frac{1}{2}})e^{-t\Sigma(\btheta^*)^{1/2}}.
\]
Take derivative to $t$,
\begin{align*}
    \frac{d}{dt}F(t)&=e^{-t\Sigma(\tilde{\btheta})^{1/2}}\left[\Sigma(\tilde{\btheta})^{\frac{1}{2}}\left(\Sigma(\tilde{\btheta})^{\frac{1}{2}}-\Sigma(\btheta^*)^{\frac{1}{2}}\right)-\left(\Sigma(\tilde{\btheta})^{\frac{1}{2}}-\Sigma(\btheta^*)^{\frac{1}{2}}\right)\Sigma(\btheta^*)^{\frac{1}{2}}\right]e^{-t\Sigma(\btheta^*)^{1/2}}\\
    &=e^{-t\Sigma(\tilde{\btheta})^{1/2}}\left[\Sigma(\tilde{\btheta})-\Sigma(\btheta^*)\right]e^{-t\Sigma(\btheta^*)^{1/2}}.
\end{align*}
Then take integral on $[0,T]$, we have
\begin{align*}
    F(T)-F(0)=-\int_0^\top e^{-t\Sigma(\tilde{\btheta})^{1/2}}\left[\Sigma(\tilde{\btheta})-\Sigma(\btheta^*)\right]e^{-t\Sigma(\btheta^*)^{1/2}}dt.
\end{align*}
By definition of $F(t)$, we have
\begin{align*}
    \|F(t)\|\leq \left\|e^{-t\Sigma(\tilde{\btheta})^{1/2}}\right\|\left\|\Sigma(\tilde{\btheta})^{\frac{1}{2}}-\Sigma(\btheta^*)^{\frac{1}{2}}\right\|\left\|e^{-t\Sigma(\btheta^*)^{1/2}}\right\|\leq \left\|e^{-t\sqrt{\lambda}}\right\|^2\left\|\Sigma(\tilde{\btheta})^{\frac{1}{2}}-\Sigma(\btheta^*)^{\frac{1}{2}}\right\|\to 0,
\end{align*}
when $t\to \infty$. Also,
\begin{align*}
    F(0)=\Sigma(\tilde{\btheta})^{\frac{1}{2}}-\Sigma(\btheta^*)^{\frac{1}{2}}.
\end{align*}
Therefore, taking $T\to \infty$ in the integral gives
\begin{align*}
    \Sigma(\tilde{\btheta})^{\frac{1}{2}}-\Sigma(\btheta^*)^{\frac{1}{2}}=-\int_0^\infty e^{-t\Sigma(\tilde{\btheta})^{1/2}}\left[\Sigma(\tilde{\btheta})-\Sigma(\btheta^*)\right]e^{-t\Sigma(\btheta^*)^{1/2}}dt.
\end{align*}
That is, we have the norm,
\begin{align*}
    \left\|\Sigma(\tilde{\btheta})^{\frac{1}{2}}-\Sigma(\btheta^*)^{\frac{1}{2}}\right\|&=\int_0^\infty \left\|e^{-t\Sigma(\tilde{\btheta})^{1/2}}\right\|\left\|\Sigma(\tilde{\btheta})-\Sigma(\btheta^*)\right\|\left\|e^{-t\Sigma(\btheta^*)^{1/2}}\right\|dt\\
    &\leq \int_0^\infty e^{-2t\sqrt{\lambda}}dt\left\|\Sigma(\tilde{\btheta})-\Sigma(\btheta^*)\right\|\leq \frac{1}{2\sqrt{\lambda}}\left\|\Sigma(\tilde{\btheta})-\Sigma(\btheta^*)\right\|\\
    &\leq \frac{L}{2\sqrt{\lambda}}\left\|\tilde{\btheta}-\btheta^*\right\|=O\left(\left\|\bb(\tilde{\btheta})\right\|\right).
\end{align*}
That is, we can write 
\begin{align}\label{eq20}
    \Sigma(\tilde{\btheta})^{\frac{1}{2}}=\Sigma(\btheta^*)^{\frac{1}{2}}+\br_7,
\end{align}
where $\|\br_7\|=O\left(\left\|\bb(\tilde{\btheta})\right\|\right)$.

Then, combine the representation in equations (\ref{eq18}), (\ref{eq19}) and (\ref{eq20}), we have a representation without $\tilde{\btheta}$ for the second term (denoted as $T_2$ for brevity) in equation (\ref{representation-with-tilde}):
\begin{align}\label{eq32}
    T_2=-\frac{\bb(\btheta^*)\left[\nabla^2\cR(\btheta^*)+\nabla \bb(\btheta^*)\right]^{-1}\nabla^2\cR(\btheta^*)\left[\nabla^2\cR(\btheta^*)+\nabla \bb(\btheta^*)\right]^{-1}\Sigma(\btheta^*)^{\frac{1}{2}}Z}{n_0+n_1+\tilde{n}}+\br,
\end{align}
where $\|\br\|=O_P\left(\frac{1}{\sqrt{n_0+n_1+\tilde{n}}}\left\|\bb(\tilde{\btheta})\right\|^2\right)$.

For the third term (denoted as $T_3$) in the representation (\ref{representation-with-tilde}), plugging in the equations (\ref{eq19}), (\ref{eq20}), and together with lipschitz condition of $\nabla^2\cR(\btheta)$, we have
\begin{align}\label{eq33}
    T_3=\frac{Z^\top \Sigma(\btheta^*)^{\frac{1}{2}}\left[\nabla^2\cR(\btheta^*)+\nabla \bb(\btheta^*)\right]^{-1}\nabla^2\cR(\btheta^*)\left[\nabla^2\cR(\btheta^*)+\nabla \bb(\btheta^*)\right]^{-1}\Sigma(\btheta^*)^{\frac{1}{2}}Z}{2(n_0+n_1+\tilde{n})^2}+\br,
\end{align}
where $\|\br\|=O_P\left(\left\|\bb(\tilde{\btheta})\right\|\right)$.

Finally, combine the representation for the second and third terms (equations (\ref{eq32}) and (\ref{eq33})), we have
{\small \begin{align*}
\nonumber&\cR(\hat{\btheta})-\cR(\btheta^*)=\frac{1}{2}\bb(\btheta^*)^\top \left(\nabla^2\cR(\btheta^*)+\nabla \bb(\btheta^*)\right)^{-1}\nabla^2\cR(\btheta^*)\left(\nabla^2\cR(\btheta^*)+\nabla \bb(\btheta^*)\right)^{-1} \bb(\btheta^*)\\
    \nonumber&-\frac{1}{n_0+n_1+\tilde{n}}\bb(\btheta^*)\left[\nabla^2\cR(\btheta^*)+\nabla \bb(\btheta^*)\right]^{-1}\nabla^2\cR(\btheta^*)\left[\nabla^2\cR(\btheta^*)+\nabla \bb(\btheta^*)\right]^{-1}\Sigma(\btheta^*)^{\frac{1}{2}}Z\\
   \nonumber &+\frac{1}{2(n_0+n_1+\tilde{n})^2}Z^\top \Sigma(\btheta^*)^{\frac{1}{2}}\left[\nabla^2\cR(\btheta^*)+\nabla \bb(\btheta^*)\right]^{-1}\nabla^2\cR(\btheta^*)\left[\nabla^2\cR(\btheta^*)+\nabla \bb(\btheta^*)\right]^{-1}\Sigma(\btheta^*)^{\frac{1}{2}}Z+R,
    \end{align*}}
where the remainder term $R$ satisfies
\begin{align*}
    R=&O_P\left(\|\bb(\tilde{\btheta})\|^3+(n_0+n_1+\tilde{n})^{-\frac{3}{2}}+\frac{1}{n_0+n_1+\tilde{n}}\|\bb(\tilde{\btheta})\|+\frac{1}{\sqrt{n_0+n_1+\tilde{n}}}\|\bb(\tilde{\btheta})\|^2\right).
\end{align*}
For the remaining term, by integral form of Taylor expansion,
\begin{align*}
    0=\nabla\tilde{\cR}(\tilde{\btheta})=\nabla\tilde{\cR}(\btheta^*)+\left(\int_0^1\nabla^2\tilde{\cR}(\btheta^*+t(\tilde{\btheta}-\btheta^*))dt\right)(\tilde{\btheta}-\btheta^*)\\=\bb(\btheta^*)+\left(\int_0^1\nabla^2\tilde{\cR}(\btheta^*+t(\tilde{\btheta}-\btheta^*))dt\right)(\tilde{\btheta}-\btheta^*).
\end{align*}
Therefore, by positive definite $\nabla^2\cR(\btheta)$ assumption,
\begin{align*}
    \|\tilde{\btheta}-\btheta^*\|\leq \left\|\left(\int_0^1\nabla^2\tilde{\cR}(\btheta^*+t(\tilde{\btheta}-\btheta^*))dt\right)^{-1}\right\|\|\bb(\btheta^*)\|\lesssim \|\bb(\btheta^*)\|.
\end{align*}
Again by integral mean value theorem
\begin{align*}
    \bb(\tilde{\btheta})&=\bb(\btheta^*)+\left(\int_0^1\nabla \bb(\btheta^*+t(\tilde{\btheta}-\btheta^*))dt\right)(\tilde{\btheta}-\btheta^*).
\end{align*}
By finite $\|\nabla \bb(\btheta)\|$ assumption, we have
\[
\left\|\int_0^1\nabla \bb(\btheta^*+t(\tilde{\btheta}-\btheta^*))dt\right\|\leq \sup \|\nabla \bb(\btheta)\|<\infty.
\]
That is
\[
\|\bb(\tilde{\btheta})\|\lesssim \|\bb(\btheta^*)\|+\left\|\int_0^1\nabla \bb(\btheta^*+t(\tilde{\btheta}-\btheta^*))dt\right\|\|\tilde{\btheta}-\btheta^*\|\lesssim \|\bb(\btheta^*)\|.
\]
We thus can get rid of $\tilde{\btheta}$ in the remaining term by bounding it as 
\begin{align*}
    R=&O_P\left(\|\bb(\btheta^*)\|^3+(n_0+n_1+\tilde{n})^{-\frac{3}{2}}+\frac{1}{n_0+n_1+\tilde{n}}\|\bb(\btheta^*)\|+\frac{1}{\sqrt{n_0+n_1+\tilde{n}}}\|\bb(\btheta^*)\|^2\right).
\end{align*}
We then finish the proof.

\subsection{Proof of Theorem \ref{thm:good-generator}}
    By the choice of $\mathscr{C}$, the synthetic size within $\mathscr{C}$ satisfies
    \[
    \left(\frac{n_0}{n_0+n_1+\tilde{n}}-\frac{1}{2}\right)^2\ll\frac{1}{n_0}.
    \]
    Applying the decomposition in Theorem \ref{representation} (and the same order bounds used in the proof of Theorem \ref{rate-improve-angle}), 
    \begin{equation}\label{eq28}
    T_1\asymp \|\bb(\btheta^*)\|^2,\quad |T_2|=\Theta_P\left(\frac{1}{\sqrt{n_0}}\|\bb(\btheta^*)\|\right),\quad T_3=\Theta_P\left(\frac{1}{n_0}\right).
    \end{equation}
    By the projection detailed in equation (\ref{eq27}), the norm satisfies
    \begin{align*}
    \|\bb(\btheta^*)\|^2&=\left(\pi_0-\frac{1}{2}+\alpha\tilde{\pi}\frac{\|\nabla\psi(\btheta^*)\|}{\|\nabla\phi(\btheta^*)\|}\right)^2\|\nabla\phi(\btheta^*)\|^2+\beta^2\tilde{\pi}^2\|\nabla\psi(\btheta^*)\|^2\\
    &=\left(\pi_0-\frac{1}{2}\right)^2\|\nabla\phi(\btheta^*)\|^2+2\alpha\tilde{\pi}\left(\pi_0-\frac{1}{2}\right)\|\nabla\phi(\btheta^*)\|\|\nabla\psi(\btheta^*)\|\\
    &\quad\quad+\alpha^2\tilde{\pi}^2\|\nabla\psi(\btheta^*)\|^2+\beta^2\tilde{\pi}^2\|\nabla\psi(\btheta^*)\|^2,
    \end{align*}
    Combining the definition of $\mathscr{C}$ and the synthetic generator quality $\|\nabla\psi(\btheta^*)\|\ll \frac{1}{\sqrt{n_0}}$, we have
    \[
    \|\bb(\btheta^*)\|\ll \frac{1}{\sqrt{n_0}}.
    \]
    Together with equation (\ref{eq28}), it holds that
    \[
    T_1\ll \frac{1}{n_0},\quad |T_2|\ll \Theta_P\left(\frac{1}{n_0}\right),\quad T_3=\Theta_P\left(\frac{1}{n_0}\right).
    \]
    We thus finish the proof by noting all terms are small terms compared with $T_3$.

\subsection{Proof of Theorem \ref{rate-improve-angle}}
For the normalized expected loss gradient difference between synthetic data and minority data, we can make a projection, such that
\[
\frac{\nabla\psi(\btheta^*)}{\|\nabla\psi(\btheta^*)\|}=\alpha \frac{\nabla\phi(\btheta^*)}{\|\nabla\phi(\btheta^*)\|}+\beta \bu,
\]
where $\bu$ is a unit vector ($\|\bu\|=1$) with direction depending on $\nabla\psi(\btheta^*)$, and we have orthogonality $\bu^\top \nabla\phi(\btheta^*)=0$. It is easy to verify that here 
\[
\alpha=\cos\angle\left(\nabla\phi(\btheta^*),\nabla\psi(\btheta^*)\right), \quad \beta =\sin\angle \left(\nabla\phi(\btheta^*),\nabla\psi(\btheta^*)\right).
\]
Therefore,
\begin{equation}\label{projection}
\nabla\psi(\btheta^*)=\alpha \frac{\|\nabla\psi(\btheta^*)\|}{\|\nabla\phi(\btheta^*)\|}\nabla\phi(\btheta^*)+\beta \|\nabla\psi(\btheta^*)\|\bu,
\end{equation}
Recall the first term 
\begin{equation}\label{eq26}
T_1=\frac{1}{2}\bb(\btheta^*)^\top \left(\nabla^2\cR(\btheta^*)+\nabla \bb(\btheta^*)\right)^{-1}\nabla^2\cR(\btheta^*)\left(\nabla^2\cR(\btheta^*)+\nabla \bb(\btheta^*)\right)^{-1} \bb(\btheta^*),
\end{equation}
we have
\[
\nabla^2\cR(\btheta^*)+\nabla \bb(\btheta^*)=\nabla^2\cR(\btheta^*)+\left(\pi_0-\frac{1}{2}\right)\nabla^2\phi(\btheta^*)+\tilde{\pi}\nabla^2\psi(\btheta^*).
\]
By assumption, we have
\[
\lambda_{\min}\left(\nabla^2\cR(\btheta^*)+\nabla \bb(\btheta^*)\right)\geq \lambda,
\]
which is lower bounded by a constant. Therefore, the first term in Theorem \ref{representation} satisfies
\begin{equation}\label{T1upper}
T_1\lesssim \|\bb(\btheta^*)\|^2.
\end{equation}
For the lower bound of the first term, we have by assumption on bounded $\|\nabla \bb(\btheta^*)\|$ and both of them are fixed numbers that
\[
\lambda_{\max}\left(\nabla^2\cR(\btheta^*)+\nabla \bb(\btheta^*)\right)\leq c\lambda_{\max}\left(\nabla^2\cR(\btheta^*)\right),
\]
which is a constant. Therefore, by the representation of the first term (\ref{eq26}), we have 
\begin{align}\label{T1lower}
    T_1\geq\frac{\lambda}{c\lambda_{\max}\left(\nabla^2\cR(\btheta^*)\right)^2}\bb(\btheta^*)^\top \bb(\btheta^*)\gtrsim \|\bb(\btheta^*)\|^2.
\end{align}
Therefore, combining equation (\ref{T1upper}) and (\ref{T1lower}) gives 
\[
T_1\asymp \|\bb(\btheta^*)\|^2.
\]
We then analyze the term $\|\bb(\btheta^*)\|^2$. By equation (\ref{projection}), it holds that
\begin{equation}\label{eq27}
\bb(\btheta^*)=\left(\pi_0-\frac{1}{2}+\alpha\tilde{\pi}\frac{\|\nabla\psi(\btheta^*)\|}{\|\nabla\phi(\btheta^*)\|}\right)\cdot\nabla\phi(\btheta^*)+\beta\tilde{\pi}\|\nabla\psi(\btheta^*)\|\cdot \bu.
\end{equation}
Taking norms and setting synthetic size as
\begin{align*}
\tilde{n}=\frac{n_0-n_1+O_P(\sqrt{n_0})}{1-2\|\nabla\psi(\btheta^*)\|/\|\nabla\phi(\btheta^*)\|\alpha+O_P(1/\sqrt{n_0})},
\end{align*}
we have
\begin{align*}
    \left\|\left(\pi_0-\frac{1}{2}+\alpha\tilde{\pi}\frac{\|\nabla\psi(\btheta^*)\|}{\|\nabla\phi(\btheta^*)\|}\right)\cdot\nabla\phi(\btheta^*)\right\|=O_P\left(\frac{1}{\sqrt{n_0}}\right),
\end{align*}
while the second term
\begin{align*}
    \left\|\beta\tilde{\pi}\|\nabla\psi(\btheta^*)\|\cdot \bu\right\|=O_P\left(\frac{1}{\sqrt{n_0}}\right),
\end{align*}
where the inequality follows by the assumption
\[
\beta=\sin\angle\left(\nabla\phi(\btheta^*),\nabla\psi(\btheta^*)\right)\lesssim \frac{1}{\|\nabla\psi(\btheta^*)\|\sqrt{n_0}}.
\]
Combining, we have
\begin{align*}
\|\bb(\btheta^*)\|^2\lesssim  \left\|\left(\pi_0-\frac{1}{2}+\alpha\tilde{\pi}\frac{\|\nabla\psi(\btheta^*)\|}{\|\nabla\phi(\btheta^*)\|}\right)\cdot\nabla\phi(\btheta^*)\right\|^2+\left\|\beta\tilde{\pi}\|\nabla\psi(\btheta^*)\|\cdot \bu\right\|^2=O_P\left(\frac{1}{n_0}\right),
\end{align*}
Similarly, it is easy to show that the second and third term follows that
\[
|T_2|=\Theta_P\left(\frac{1}{\sqrt{n_0}}\|\bb(\btheta^*)\|\right)\lesssim\Theta_P\left(\frac{1}{n_0}\right),\quad T_3=\Theta_P\left(\frac{1}{n_0}\right).
\]
Therefore, the excess risk satisfies
\[
        \cR(\hat{\btheta})-\cR(\btheta^*)=\Theta_P\left( \frac{1}{n_0}\right).
        \]

For the second statement, simply plugging in the synthetic size $\tilde{n}=n_0-n_1$ gives
\[
\bb(\btheta^*)=\alpha\frac{n_0-n_1}{2n_0}\frac{\|\nabla\psi(\btheta^*)\|}{\|\nabla\phi(\btheta^*)\|}\cdot\nabla\phi(\btheta^*)+\beta\frac{n_0-n_1}{2n_0}\|\nabla\psi(\btheta^*)\|\cdot \bu.
\]
Taking norms and squaring both sides,
\[
\|\bb(\btheta^*)\|^2=\alpha^2\left(\frac{n_0-n_1}{2n_0}\right)^2\|\nabla\psi(\btheta^*)\|^2+\beta^2\left(\frac{n_0-n_1}{2n_0}\right)^2\|\nabla\psi(\btheta^*)\|^2\asymp\|\nabla\psi(\btheta^*)\|^2.
\]
That is, it follows that
\[
T_1\asymp\|\nabla\psi(\btheta^*)\|^2,\quad |T_2|=\Theta_P\left(\frac{1}{\sqrt{n_0}}\|\nabla\psi(\btheta^*)\|\right),\quad T_3=\Theta_P\left(\frac{1}{n_0}\right).
\]
Finally, we have the excess risk
\[
        \cR(\hat{\btheta})-\cR(\btheta^*)=\Theta_P\left(\|\nabla\psi(\btheta^*)\|^2\right)\gg \Theta_P\left(\frac{1}{n_0}\right).
        \]
We then finish the proof.

\subsection{Proof of Theorem \ref{inconsistent}}

By similar procedure with proof of theorem \ref{rate-improve-angle}, we make projection
\[
\frac{\nabla\psi(\btheta^*)}{\|\nabla\psi(\btheta^*)\|}=\alpha \frac{\nabla\phi(\btheta^*)}{\|\nabla\phi(\btheta^*)\|}+\beta \bu,
\]
where $\bu$ is a unit vector ($\|\bu\|=1$) with direction depending on $\nabla\psi(\btheta^*)$, and we have orthogonality $\bu^\top \nabla\phi(\btheta^*)=0$. That is,
\begin{equation*}
\nabla\psi(\btheta^*)=\alpha \frac{\|\nabla\psi(\btheta^*)\|}{\|\nabla\phi(\btheta^*)\|}\nabla\phi(\btheta^*)+\beta \|\nabla\psi(\btheta^*)\|\bu,
\end{equation*}
Plugging in gives
\begin{align*}
    \bb(\btheta^*)=\left(\pi_0-\frac{1}{2}+\alpha\tilde{\pi} \frac{\|\nabla\psi(\btheta^*)\|}{\|\nabla\phi(\btheta^*)\|}\right)\nabla\phi(\btheta^*)+\beta\tilde{\pi}\|\nabla\psi(\btheta^*)\|\bu.
\end{align*}
With the selection of 
\[
    \tilde{n}=\frac{n_0-n_1+(n_0+n_1)o(1)}{1-2\|\nabla\psi(\btheta^*)\|/\|\nabla\phi(\btheta^*)\|\alpha+o(1)},
    \]
we have 
    \[
    \pi_0-\frac{1}{2}+\alpha\tilde{\pi} \frac{\|\nabla\psi(\btheta^*)\|}{\|\nabla\phi(\btheta^*)\|}=o_P(1).
    \]
Also by assumption, $\beta=o(1)$. Therefore, we have the total synthetic bias
\[
\|\bb(\btheta^*)\|=o_P(1).
\]
By previous analysis of the excess risk, we have
\[
\cR(\hat{\btheta})-\cR(\btheta^*)=o_P(1).
\]

If we still choose synthetic size $\tilde{n}=n_0-n_1$, it holds that
\begin{align*}
    \liminf_{n_0,n_1\to\infty}\|\bb(\btheta^*)\|\geq \liminf_{n_0,n_1\to\infty}\left|\alpha\left(\frac{1}{2}-\frac{n_1}{2n_0}\right)\right|\|\nabla\psi(\btheta^*)\|\geq c>0.
\end{align*}
Thus by Theorem \ref{excess-lowerbound}, the excess risk
\[
\liminf_{n_0,n_1\to\infty}\cR(\hat{\btheta})-\cR(\btheta^*)\geq c>0,
\]
in probability.

When synthetic direction is not aligned such that $\sin\angle\left(\nabla\phi(\btheta^*),\nabla\psi(\btheta^*)\right)>c>0$ in probability, and $\tilde{n}/n_0\to \rho$, we have $\tilde{\pi}\to \frac{\rho}{1+\rho+n_1/n_0}$, $\pi_0\to \frac{1}{1+\rho+n_1/n_0}$. Therefore,
\begin{align*}
    \|\bb(\btheta^*)\|^2&=\left(\pi_0-\frac{1}{2}+\alpha\tilde{\pi} \frac{\|\nabla\psi(\btheta^*)\|}{\|\nabla\phi(\btheta^*)\|}\right)^2\|\nabla\phi(\btheta^*)\|^2+\beta^2\tilde{\pi}^2\|\nabla\psi(\btheta^*)\|^2\|\bu\|^2\gtrsim c
\end{align*}
Thus by Theorem \ref{excess-lowerbound}, the excess risk satisfies
\[
\cR(\hat{\btheta})-\cR(\btheta^*)\geq c+o_P(1).
\]

\subsection{Proof of Theorem \ref{thm:cannothelp}}

    For diminishing synthetic size $\tilde{n}=O(\sqrt{n_0}/\|\nabla\psi(\btheta^*)\|)$, under $\nabla\phi(\btheta^*)=0$ and a realistic generator, we have 
    \[
    \bb(\btheta^*)=\tilde{\pi}\nabla\psi(\btheta^*)\quad \Rightarrow \quad \|\bb(\btheta^*)\|=\frac{\tilde{n}}{n_0+n_1+\tilde{n}}\|\nabla\psi(\btheta^*)\|=O(n_0^{-1/2}).
    \]
    From previous analysis of Theorem \ref{representation}, it holds that
    \[
    T_1\asymp \|\bb(\btheta^*)\|^2,\quad T_2=\Theta_P\left(\frac{\|\bb(\btheta^*)\|}{\sqrt{n_0+n_1+\tilde{n}}}\right),\quad T_3=\Theta_P\left( \frac{1}{n_0+n_1+\tilde{n}}\right),
    \]
    and remaining term $R$ is smaller order term under this setting. Thus, we conclude that
    \[
    \cR(\hat{\btheta})-\cR(\btheta^*)=O_P(n_0^{-1}).
    \]

    For the non-diminishing synthetic size $\tilde{n}\gtrsim n_0$, we have
    \[
    \|\bb(\btheta^*)\|=\frac{\tilde{n}}{n_0+n_1+\tilde{n}}\|\nabla\psi(\btheta^*)\|\gtrsim \|\nabla\psi(\btheta^*)\|.
    \]
    Similarly by the order analysis of Theorem \ref{representation},
    \[
    \cR(\hat{\btheta})-\cR(\btheta^*)=\Theta_P(\|\bb(\btheta^*)\|^2)\gtrsim \Omega_P(\|\nabla\psi(\btheta^*)\|^2)\gg \Theta_P(n_0^{-1}).
    \]

\subsection{Proof of Proposition \ref{meanshiftcancel}}
    For the squared loss, we have
    \[
f_{\btheta}(\bx)=\btheta^\top \bx,\quad \ell(\btheta;\bx,y)=(y-1/2-\btheta^\top \bx)^2,\quad \nabla\ell(\btheta;\bx,y)=-2(y-1/2-\btheta^\top \bx)\bx.
\]
Thus for the population of majority,
\begin{align*}
    \bE_{\cP_0}\nabla\ell(\btheta;\bx,0)=\bE_{\cP_0}[(1+2\btheta^\top \bx)\bx]=2(\bmu\bmu^\top +\bE[\bxi\bxi^\top ])\btheta-\bmu.
\end{align*}
Similarly for minority
\begin{align*}
    \bE_{\cP_1}\nabla\ell(\btheta;\bx,1)=\bE_{\cP_1}[(2\btheta^\top \bx-1)\bx]=2(\bmu\bmu^\top +\bE[\bxi\bxi^\top ])\btheta-\bmu.
\end{align*}
Thus the derivative of the balanced risk is
\[
\nabla\cR(\btheta)=\frac{1}{2}\bE_{\cP_0}\nabla\ell(\btheta;\bx,0)+\frac{1}{2}\bE_{\cP_1}\nabla\ell(\btheta;\bx,1)=2(\bmu\bmu^\top +\bE[\bxi\bxi^\top ])\btheta-\bmu=0.
\]
Solving gives the minimum
\[
\btheta^*=\frac{1}{2}\left[\bmu\bmu^\top +\bE[\bxi\bxi^\top ]\right]^{-1}\bmu.
\]
Also, plugging in gives the majority-minority bias cancels
\[
\nabla\phi(\btheta^*)=\bE_{\cP_0}\nabla\ell(\btheta;\bx,0)-\bE_{\cP_1}\nabla\ell(\btheta;\bx,1)=0.
\]

\subsection{Proof of Proposition \ref{Logisticcancel}}
    Under the data generating mechanism of sigmoid Bernoulli model, the class conditionals are
    \begin{align*}
        &\cP_1\sim\mathbb{P}(\bx~|~y=1)=\frac{\mathbb{P}(y=1~|~\bx)p(\bx)}{\mathbb{P}(y=1)}=\frac{p(\bx)\sigma(\btheta_{\mathrm{true}}^\top \bx)}{\mathbb{P}(y=1)},\\
        &\cP_0\sim\mathbb{P}(\bx~|~y=0)=\frac{\mathbb{P}(y=0~|~\bx)p(\bx)}{\mathbb{P}(y=0)}=\frac{p(\bx)(1-\sigma(\btheta_{\mathrm{true}}^\top \bx))}{\mathbb{P}(y=0)},
    \end{align*}
    where 
    \begin{align*}
    &\mathbb{P}(y=1)=\int \mathbb{P}(\bx~,~y=1)=\int \mathbb{P}(y=1~|~\bx)p(\bx)=\bE_{p}[\sigma(\btheta^\top _{\mathrm{true}}\bx)],\\ &\mathbb{P}(y=0)=1-\bE_{p}[\sigma(\btheta^\top _{\mathrm{true}}\bx)].
    \end{align*}
    The derivative of the logistic loss is
    \[
    \nabla\ell(\btheta;\bx,y)=(\sigma(\btheta^\top \bx)-y)\bx.
    \]
    Then for majority under the true parameter,
    \begin{align*}
    &\bE_{\cP_0}\nabla\ell(\btheta_{\mathrm{true}};\bx,0)=\bE_{\cP_0}\sigma(\btheta_{\mathrm{true}}^\top \bx)\bx\\
    &=\int \sigma(\btheta_{\mathrm{true}}^\top \bx)\bx\frac{p(\bx)(1-\sigma(\btheta_{\mathrm{true}}^\top \bx))}{\mathbb{P}(y=0)}=\frac{1}{\mathbb{P}(y=0)}\bE_{p(\bx)}\left[\bx\sigma(\btheta_{\mathrm{true}}^\top \bx)\left(1-\sigma(\btheta_{\mathrm{true}}^\top \bx)\right)\right].
    \end{align*}
    Similarly for minority
    \[
    \bE_{\cP_1}\nabla\ell(\btheta_{\mathrm{true}};\bx,1)=-\frac{1}{\mathbb{P}(y=1)}\bE_{p(\bx)}\left[\bx\sigma(\btheta_{\mathrm{true}}^\top \bx)\left(1-\sigma(\btheta_{\mathrm{true}}^\top \bx)\right)\right].
    \]
    Then we conclude that $\btheta_{\mathrm{true}}$ is the minimizer of the balanced risk by noticing
    \begin{align*}
    \nabla\cR(\btheta_{\mathrm{true}})&=\frac{1}{2}\bE_{\cP_0}\nabla\ell(\btheta_{\mathrm{true}};\bx,0)+\frac{1}{2}\bE_{\cP_1}\nabla\ell(\btheta_{\mathrm{true}};\bx,1)\\
    &=\frac{1}{2}\left(\frac{1}{\mathbb{P}(y=0)}-\frac{1}{\mathbb{P}(y=1)}\right)\bE_{p(\bx)}\left[\bx\sigma(\btheta_{\mathrm{true}}^\top \bx)\left(1-\sigma(\btheta_{\mathrm{true}}^\top \bx)\right)\right]=0.
    \end{align*}
    By strong convexity around the minimizer of $\cR(\btheta)$, we have uniqueness of the minimizer, and thus $\btheta^*=\btheta_{\mathrm{true}}$. Therefore the majority-minority difference cancels
    \begin{align*}
    \nabla\phi(\btheta^*)&=\bE_{\cP_0}\nabla\ell(\btheta_{\mathrm{true}};\bx,0)-\bE_{\cP_1}\nabla\ell(\btheta_{\mathrm{true}};\bx,1)\\
    &=\left(\frac{1}{\mathbb{P}(y=0)}+\frac{1}{\mathbb{P}(y=1)}\right)\bE_{p(\bx)}\left[\bx\sigma(\btheta_{\mathrm{true}}^\top \bx)\left(1-\sigma(\btheta_{\mathrm{true}}^\top \bx)\right)\right]=0.
    \end{align*}
    We thus finish the proof.

\section{Examples for Choice of Loss}\label{sec:loss}
In this section, we summarize common choices of loss functions for classification.
\paragraph*{Cross-entropy/logistic loss.} If the model outputs a probability $p_{\btheta}(\boldsymbol{x})\in(0,1)$, the cross-entropy loss is
$\ell(\btheta;\boldsymbol{x},y) = -y\log p_{\btheta}(\boldsymbol{x})-(1-y)\log(1-p_{\btheta}(\boldsymbol{x}))$.
A common parameterization is $p_{\btheta}(\boldsymbol{x})=\sigma(f_{\btheta}(\boldsymbol{x}))$ with sigmoid $\sigma(t)=1/(1+e^{-t})$. This is equivalent to maximum likelihood for a Bernoulli model and typically yields well-calibrated probabilities when the model is correctly specified. Under oversampling, however, the effective class prior in training changes (because $\tilde\pi$ increases the frequency of label 1), so probability calibration may require adjustment if one wants probabilities under the original prior.

\paragraph*{Square loss.} With the square loss $\ell(\btheta;\boldsymbol{x},y)=(y-p_{\btheta}(\boldsymbol{x}))^2$,  binary classification is treated as regression to the label $y\in \{0,1\}$ using the model output $p_{\btheta}(\boldsymbol{x})$ as a real-valued prediction. This objective is simple to optimize and often convenient analytically. However, it is less robust to outliers in probability space.

\paragraph*{Hinge loss.}
For output score $f_{\btheta}(\boldsymbol{x})$, the hinge loss can be written as $\ell(\btheta;\boldsymbol{x},y)=\max(0,1-(2y-1)f_{\btheta}(\boldsymbol{x}))$. It encourages a large margin between classes and underlies SVMs. Margin-based losses tend to be less sensitive to probability calibration but are sensitive to the geometry of the feature distribution. Synthetic samples that distort the minority geometry can therefore affect the margin and support vectors. When labels are written in $\{-1,+1\}$, we can map $y\in\{0,1\}$ to $y'\in\{-1,+1\}$ via $y' = 2y-1$, the hinge loss becomes
$\ell(\btheta;\boldsymbol{x},y')=\max(0,1-y'f_{\btheta}(\boldsymbol{x}))$. 

\paragraph*{Exponential loss (boosting).}
The exponential loss is $\ell_{\exp}(\btheta;\boldsymbol{x},y)=\exp(-(2y-1)f_{\btheta}(\boldsymbol{x}))$, where $f_{\btheta}(\boldsymbol{x})$ is a real-valued score. This loss heavily penalizes misclassified points with large negative margins and is associated with AdaBoost-like algorithms. Because it emphasizes hard-to-classify points, it can interact strongly with synthetic samples: ``too-easy'' synthetic points may have little effect, while poorly generated synthetic points can dominate the gradient. Similarly when labels are written in $y'\in\{-1,+1\}$, the exponential loss becomes
$\ell(\btheta;\boldsymbol{x},y')=\exp(-y'f_{\btheta}(\boldsymbol{x}))$. 

\section{Related Classifiers}\label{sec:classifiers}

In this section, we review several common classifiers and relate them to the empirical risk minimization (ERM) framework used in this paper. Many discriminative methods, such as logistic regression, support vector machines, and boosting, can be written as minimizing an empirical average loss (with regularization), and therefore fall directly within our setting.
In contrast, some classical procedures are not naturally expressed as minimizing a fixed discriminative risk over a decision function. For example, linear discriminant analysis (LDA) is derived from a generative Gaussian model and applies the plug-in Bayes rule after estimating distributional parameters; although this corresponds to likelihood optimization over generative parameters, it is not a standard discriminative ERM formulation.

\subsection{Discriminative classifiers in the ERM framework}

\paragraph*{Logistic regression.}
Logistic regression is a probabilistic linear classifier for binary outcomes that models the log-odds of the positive class as a linear function of features. Logistic regression also sits naturally inside the generalized linear model (GLM) framework via the logit link for binomial data.

Given data $\{(\bx_i,y_i)\}_{i=1}^n$ with $y_i\in\{0,1\}$, logistic regression models
\[
\mathbb P_{\btheta}(y=1\mid \bx)=\sigma(\btheta^\top \bx),\qquad 
\sigma(u)=\frac{1}{1+e^{-u}}.
\]
The negative log-likelihood corresponds to the logistic (cross-entropy) loss
\[
\ell(\btheta; \bx_i,y_i)
= -y_i\log \sigma(\btheta^\top \bx_i) -(1-y_i)\log\big(1-\sigma(\btheta^\top \bx_i)\big),
\]
and the estimator is an empirical risk minimizer
\[
\hat\btheta\in\arg\min_{\btheta\in\Theta}\ \frac{1}{n}\sum_{i=1}^n \ell(\btheta;\bx_i,y_i).
\]

Common extensions of logistic regression still fit the same empirical risk minimization framework. Main extensions include Softmax Logistic Regression for $K>2$ classes, Penalized Logistic Regression for high-dimensional settings and Kernel Logistic Regression for learning nonlinear decision boundaries.

\paragraph*{Support vector machine (SVM).}
Support Vector Machines originated from the maximum-margin idea: learn a separating hyperplane that maximizes the margin between classes. Early foundational formulations include \cite{boser1992training} and \cite{cortes1995support}.

In the binary case with labels $y_i\in\{0,1\}$, the (soft-margin) SVM can be written as regularized
empirical risk minimization with the hinge loss:
\[
(\hat{\boldsymbol{w}},\hat b)\in \arg\min_{\boldsymbol{w},b}\;
\frac{\lambda}{2}\|\boldsymbol{w}\|_2^2
+\frac{1}{n}\sum_{i=1}^n
\max\Bigl\{0,\;1-(2y_i-1)\,(\boldsymbol{w}^\top \phi(\bx_i)+b)\Bigr\}.
\]
where $\phi(\cdot)$ is a feature map and $\lambda>0$ controls
regularization. This is equivalent (up to reparameterization) to the classic constrained soft-margin form:
\[
\min_{\boldsymbol{w},b,\xi\ge 0}\;\frac12\|\boldsymbol{w}\|_2^2 + C\sum_{i=1}^n \xi_i
\quad\text{s.t.}\quad
\begin{cases}
\boldsymbol{w}^\top\phi(\bx_i)+b \ge 1-\xi_i & \text{if } y_i=1,\\
\boldsymbol{w}^\top\phi(\bx_i)+b \le -1+\xi_i & \text{if } y_i=0.
\end{cases}
\]

\paragraph*{Boosting.}
Boosting is an ensemble methodology that builds a strong predictor by combining many “weak” base learners. The original theoretical foundation is \cite{schapire1990strength}, which formalized why boosting can work in principle. A widely used practical instantiation is AdaBoost, derived by \cite{freund1997decision} in their decision-theoretic online learning framework and then applied to boosting.

A common view is that boosting learns an additive model
\[
F_M(\bx)=\sum_{m=1}^M \alpha_m h_m(\bx),\qquad h_m\in\mathcal H,
\]
by (approximately) minimizing an empirical risk
\[
\hat R(F)=\frac{1}{n}\sum_{i=1}^n L \big(y_i, F(\bx_i)\big),
\]
using a greedy or stagewise procedure (choosing the next $h_m$ and step size $\alpha_m$ to reduce $\hat R$).
In particular, AdaBoost can be interpreted as stagewise minimization of the exponential loss
\[
L(y,F)=\exp\{-yF\},\qquad y\in\{-1,+1\},
\]
which induces the familiar reweighting scheme: at iteration $m$, examples receive weights $\boldsymbol{w}_i^{(m)}\propto \exp\{-y_i F_{m-1}(\bx_i)\}$, so misclassified or low-margin points are emphasized when fitting the next weak learner.

\paragraph*{Random forest.}
Random Forests are an ensemble of decision trees built with two main sources of randomness, bootstrap resampling of the training set (bagging) and random feature selection at each split. The results are then aggregated through majority vote for classification \citep{breiman2001random}. The method explicitly builds on bagging and is closely related to the random subspace idea for constructing tree ensembles.

While a random forest is an ensemble rather than a single parametric ERM, it fits the empirical-risk
minimization view at two levels. First, each decision tree is built by greedy minimization of a
node-wise empirical loss. For a node containing samples $S$, a split rule $s$ partitions $S$
into children $S_L(s)$ and $S_R(s)$, and a CART-style tree selects
\[
s^* \in \arg\min_{s\in\mathcal S}\ 
\frac{|S_L(s)|}{|S|}  I \big(S_L(s)\big)
+\frac{|S_R(s)|}{|S|}  I \big(S_R(s)\big),
\]
where $I(\cdot)$ is an impurity, e.g., Gini or entropy for classification. Random forests introduce additional randomness by fitting each tree on a
bootstrap resample of the training data and by restricting $\mathcal S$ at each node to splits over a
random subset of features. Second, the forest aggregates $B$ trees to form the predictor
\[
\hat f(\bx)=\frac{1}{B}\sum_{b=1}^B f_b(\bx)\quad\text{(regression)},\qquad
\hat y(\bx)=\mathrm{majority\_vote}\{f_b(\bx)\}_{b=1}^B\quad\text{(classification)},
\]
which reduces prediction loss primarily through variance reduction from averaging de-correlated trees.

\paragraph*{Neural networks.}
Neural networks for classification trace back to the perceptron, a linear threshold model for binary classification introduced by \cite{rosenblatt1958perceptron}. Modern multi-layer perceptrons (MLPs) extend this to multiple nonlinear hidden layers. Their practical success was enabled by efficient gradient-based training via backpropagation.

Given data $\{(\bx_i,y_i)\}_{i=1}^n$, a neural network classifier is a parametric function $f_{\btheta}$
(e.g., an MLP) producing logits $s_{\btheta}(\bx)\in\mathbb{R}^K$. A standard probabilistic classifier uses
the softmax model
\[
p_{\btheta}(y=k\mid \bx)=\frac{\exp(s_{\btheta,k}(\bx))}{\sum_{j=1}^K \exp(s_{\btheta,j}(\bx))}.
\]
Training fits the empirical-risk framework by minimizing a regularized empirical loss:
\[
\hat\btheta \in \arg\min_{\btheta}\ \frac{1}{n}\sum_{i=1}^n \ell\big(y_i, s_{\btheta}(\bx_i)\big)
+ \lambda  \Omega(\btheta),
\]
where $\ell$ is typically the cross-entropy (negative log-likelihood under the softmax model) and
$\Omega$ is a regularizer (e.g., $\|\btheta\|_2^2$). In practice, $\hat\btheta$ is often found by stochastic
gradient methods using backpropagation.

\subsection{Generative classification}

\paragraph*{Linear Discriminant analysis.} 
LDA is a classical generative classifier that models each class as a multivariate Gaussian distribution sharing a common covariance matrix. Assume $K$ classes $y\in\{1,\dots,K\}$ and $\bx\in\mathbb{R}^d$. LDA assumes
\[\bx\mid (y=k)\sim \cN(\mu_k,\Sigma),\qquad \mathbb P(y=k)=\pi_k,\]
with shared $\Sigma$ across classes. The parameters are typically estimated by maximum likelihood (equivalently, minimizing the negative log-likelihood over the generative parameters), after which the classifier is obtained via the plug-in Bayes rule:
\[
\hat y(\bx)=\arg\max_{k\in\{1,\dots,K\}}\delta_k(\bx),
\]
with the usual linear discriminant scores $\delta_k(\bx)$. Thus, LDA aligns with an ERM view over \emph{generative} parameters, but it is not naturally expressed as minimizing a fixed \emph{discriminative} loss $\frac{1}{n}\sum_{i=1}^n \ell(y_i,f(\bx_i))$ over a decision function $f$ of the form emphasized in our framework.

\section{Related Synthetic Sample Generators}\label{sec:generators}

In this section, we briefly review the related synthetic data generators. These generators span both classical resampling-based approaches and modern model-based generative methods.

\subsection{Classical Resampling and Interpolation Methods}

\subsubsection{Bootstrap and Reweighting}

The bootstrap is a general-purpose resampling method introduced by \citet{efron1994introduction} for approximating sampling distributions and constructing uncertainty estimates. In its basic form, given a dataset of size $n$, one (i) draws a bootstrap resample by sampling $n$ observations with replacement, (ii) fits the model or computes the statistic on the resample, and (iii) repeats this procedure $B$ times to obtain an empirical distribution of the estimator (or predictions). In imbalanced learning, bootstrap-style resampling is often used as a simple data-level intervention: by resampling within each class, one can create approximately balanced training sets.

Beyond the classical bootstrap, several variants improve statistical behavior or scalability, such as the Bayesian bootstrap \citep{rubin1981bayesian} and the bag of little bootstraps for massive datasets
\citep{kleiner2014scalable}.

Reweighting addresses imbalance by modifying the training objective rather than generating new samples: one assigns class- or example-dependent weights and minimizes a weighted empirical risk, often choosing weights inversely proportional to class frequency or according to a misclassification cost matrix. At the level of the objective, resampling and reweighting are closely connected: duplicating an example in a resampled dataset is equivalent to assigning it a larger weight in a weighted empirical risk, and importance weights can be implemented either by passing weights to the learner or by appropriate resampling schemes \citep{chen2024survey}.

As a baseline ``generator,'' bootstrap-based oversampling has important drawbacks: it creates duplicates rather than new minority information, which can increase overfitting \citep{he2009learning}. It can also be sensitive to noise or outliers, since a rare noisy point may be selected multiple times and thus receive inflated influence \citep{chen2024survey}.

\subsubsection{SMOTE and Variants}\label{sec:supp-SMOTE}

SMOTE (Synthetic Minority Over-sampling Technique) is one of the most widely used synthetic oversampling methods for imbalanced classification. It was introduced by \cite{chawla2002smote} as an alternative to naive minority replication, with the goal of expanding the minority region in feature space and improving classifier performance. Conceptually, SMOTE treats “synthetic data generation” as interpolation rather than duplication: it creates new minority samples between existing minority points and their neighbors. For a desired oversampling rate, SMOTE repeatedly: (i) selects a minority instance $\bx$, (ii) finds its $k$ nearest minority neighbors, (iii) randomly chooses one neighbor $\bx_{(j)}$, and (iv) generates a synthetic point by linear interpolation
\[
\bx_{\mathrm{new}}=\bx+\gamma(\bx_{(j)}-\bx),\quad \gamma\sim \mathrm{Uniform}(0,1),
\]
so the synthetic example lies on the line segment between $\bx$ and a minority neighbor. This procedure is applied until the required number of synthetic minority samples is produced.

Many SMOTE variants modify where and how interpolation occurs to better target difficult regions or reduce noise. Borderline-SMOTE oversamples primarily minority points near the decision boundary (“borderline” instances), motivated by the idea that these are more informative yet more error-prone \citep{han2005borderline}. ADASYN (Adaptive Synthetic Sampling) further makes the number of generated synthetic points adaptive, producing more samples in regions where minority points are harder to learn \citep{he2008adasyn}. Also, more recent work incorporates structure in the minority distribution. For example, k-means SMOTE uses clustering to guide where synthetic samples are most helpful while aiming to avoid generating unnecessary noise \citep{last2017oversampling}.

A key advantage of SMOTE over naive random oversampling is that it generates new minority samples by interpolation rather than duplicating existing points, which can reduce overfitting relative to simple replication-based oversampling, and it can also fill in the interior of minority clusters to make them easier for a classifier to model. On the downside, because SMOTE interpolates from existing minority points, if some minority instances are noisy or outliers, the algorithm can generate additional synthetic samples around these abnormal points, effectively propagating noise and hurting classification \citep{matharaarachchi2024enhancing}. Moreover, SMOTE generates points along line segments between neighbors (i.e., within the convex hull implied by observed minority samples). When the true minority support is non-convex (e.g., two separated regions or curved manifolds), interpolation can place synthetic samples in low-density regions or even inside majority regions, thereby degrading the decision boundary.

\subsection{Deep Generative Models}

\subsubsection{Generative Adversarial Network (GAN)}
Generative Adversarial Network (GAN) is a class of implicit generative models introduced by \cite{goodfellow2014generative}, where a generator learns to synthesize samples that resemble the training data while a discriminator learns to distinguish real from generated samples. Learning proceeds as a two-player minimax game.

The GAN can be summarized as follows.
Let $\bz\sim p(\bz)$ be a noise source (e.g., $\bz\sim\cN(0,I)$), $G_{\btheta}(\bz)$ be the generator, and
$D_\phi(\bx)\in(0,1)$ be the discriminator output interpreted as the probability that $\bx$ is real.
The original GAN objective \citep{goodfellow2014generative} is the minimax game
\[
\min_{\btheta}\max_{\phi}   V(\phi,\btheta)
=
\mathbb E_{\bx\sim p_{\text{data}}}\big[\log D_\phi(\bx)\big]
+
\mathbb E_{\bz\sim p(\bz)}\big[\log\big(1-D_\phi(G_{\btheta}(\bz))\big)\big].
\]
Training alternates between updating $\phi$ to increase $V(\phi,\btheta)$ (improving discrimination) and
updating $\btheta$ to decrease $V(\phi,\btheta)$ (fooling the discriminator), typically via stochastic
gradient steps on mini-batches. After convergence, synthetic sampling is performed by drawing $\bz\sim p(\bz)$ and outputting $\bx=G_{\btheta}(\bz)$.

Many GAN variants modify architecture, conditioning, divergence or optimization to improve stability and fidelity. For example, DCGAN established convolutional architectural guidelines for stable image GAN training \citep{radford2015unsupervised}. Conditional GAN (cGAN) conditions both $G$ and $D$ on label information $y$ to generate $p(\bx|y)$ \citep{mirza2014conditional}. InfoGAN adds a mutual-information term to encourage disentangled and interpretable latent factors \citep{chen2016infogan}.

\subsubsection{Variational Autoencoders (VAEs)}
Variational Autoencoders (VAEs) are latent variable generative models that combine probabilistic modeling with neural-network encoder-decoder parameterizations. They were popularized by \cite{kingma2013auto}, which showed how to train deep latent variable models efficiently with gradient methods. At a high level, a VAE learns (i) a generative model that maps latent variables to data, and (ii) an inference (recognition) model that approximates the intractable posterior over latents given data.

Variational autoencoders posit a latent variable model
\[
p_{\btheta}(\bx,\bz)=p(\bz)  p_{\btheta}(\bx\mid \bz),
\]
typically with a simple prior $p(\bz)=\cN(0,I)$. Since the posterior $p_{\btheta}(\bz\mid \bx)$ is generally
intractable, VAEs introduce an amortized variational approximation $q_\phi(\bz\mid \bx)$ and maximize the
evidence lower bound (ELBO):
\[
\log p_{\btheta}(\bx)\ \ge\ \mathcal L(\btheta,\phi;\bx)
:=\mathbb E_{q_\phi(\bz\mid \bx)} \big[\log p_{\btheta}(\bx\mid \bz)\big]
-\mathrm{KL} \left(q_\phi(\bz\mid \bx)  \|  p(\bz)\right).
\]
Training proceeds by stochastic gradient ascent on $\sum_{i=1}^n \mathcal L(\btheta,\phi;\bx_i)$.
For continuous latents (e.g., a Gaussian encoder), the reparameterization trick writes a sample as
\[
\bz=\mu_\phi(\bx)+\sigma_\phi(\bx)\odot \varepsilon,\qquad \varepsilon\sim\cN(0,I),
\]
so Monte Carlo estimates of $\nabla_{\btheta,\phi}\mathcal L(\btheta,\phi;\bx)$ can be backpropagated through
the stochastic node. After training, synthetic sampling is performed by drawing $\bz\sim p(\bz)$ and then
drawing $\bx\sim p_{\btheta}(\bx\mid \bz)$.

Many VAE variants improve expressiveness, conditioning, or sample quality. IWAE tightens the ELBO by using an importance-weighted bound with multiple samples from $q_\phi(\bz|\bx)$, improving density modeling and learned representations \citep{burda2015importance}. $\beta$-VAE modifies the KL term to encourage disentangled latent factors \citep{higgins2017beta}. Conditional VAE (CVAE) conditions generation on side information $y$ to model $p_{\btheta}(x|y)$ via latent variable formulations, supporting controlled prediction \citep{sohn2015learning}.

\subsubsection{Diffusion Models}\label{sec:supp-diffusion}
Diffusion models are generative models that learn to sample from a data distribution by reversing a gradual noising process. The original diffusion-probabilistic framing is often traced to \cite{sohl2015deep}, who proposed transforming data into noise through a Markov diffusion and learning the reverse-time dynamics for generation. A highly influential modern formulation is DDPM (Denoising Diffusion Probabilistic Models) \citep{ho2020denoising}, which showed that diffusion models can produce high-quality image samples using a denoising training objective closely related to score matching, without adversarial training.

The basic steps of DDPM can be summarized as follows. Let $\bx_0 \sim p_{\text{data}}$ and choose a variance schedule $\{\beta_t\}_{t=1}^T$ with
$\alpha_t = 1-\beta_t$ and $\bar\alpha_t = \prod_{s=1}^T \alpha_s$.

\emph{(1) Forward diffusion (noising).}
Define the Markov chain
\[
q(\bx_t \mid \bx_{t-1}) = \cN \big(\sqrt{\alpha_t}  \bx_{t-1},   \beta_t I\big),
\]
which implies the closed form marginal
\[
q(\bx_t \mid \bx_0) = \cN \big(\sqrt{\bar\alpha_t}  \bx_0,  (1-\bar\alpha_t)I\big),
\quad \text{equivalently} \quad
\bx_t = \sqrt{\bar\alpha_t}  \bx_0 + \sqrt{1-\bar\alpha_t}  \varepsilon,\ \varepsilon \sim \cN(0,I).
\]

\emph{(2) Learn the reverse denoiser.}
Parameterize a network $\varepsilon_{\btheta}(\bx_t,t)$ to predict the injected noise and train via the
(simplified) denoising objective
\[
\min_{\btheta}\ 
\mathbb E_{\substack{t\sim \mathrm{Unif}\{1,\dots,T\}\\ \bx_0\sim p_{\text{data}}\\ \varepsilon\sim \cN(0,I)}}
\big\|\varepsilon - \varepsilon_{\btheta}(\bx_t,t)\big\|_2^2.
\]
This corresponds to a Gaussian reverse kernel
\[
p_{\btheta}(\bx_{t-1}\mid \bx_t) = \cN \big(\mu_{\btheta}(\bx_t,t),  \Sigma_t\big),
\]
with a common parameterization
\[
\mu_{\btheta}(\bx_t,t) = \frac{1}{\sqrt{\alpha_t}}
\left(
\bx_t - \frac{\beta_t}{\sqrt{1-\bar\alpha_t}}\ \varepsilon_{\btheta}(\bx_t,t)
\right),
\]
and $\Sigma_t$ chosen as $\beta_t I$ or learned or modified.

\emph{(3) Sampling (generation).}
Draw $\bx_T \sim \cN(0,I)$ and iterate for $t=T,\dots,1$:
\[
\bx_{t-1} = \mu_{\btheta}(\bx_t,t) + \Sigma_t^{1/2} \bz,\quad \bz\sim \cN(0,I),
\]
yielding $\bx_0$ as the synthetic sample.

There are many variants and extensions of diffusion models. For example, a major unifying view is score-based diffusion via SDEs, which formulates diffusion in continuous time and connects sampling to solving a reverse-time SDE \citep{song2020score}. To address slow sampling, DDIM (denoising diffusion implicit models) introduces a non-Markovian formulation that can sample in far fewer steps while using the same training procedure as DDPMs \citep{song2020denoising}. For efficiency at high resolution, Latent Diffusion Models (LDMs) perform diffusion in a learned latent space to reduce computation while maintaining strong synthesis performance \citep{rombach2022high}.

\subsubsection{Flow Matching}
Flow Matching is a training approach for continuous normalizing flows (CNFs) that learns a time-dependent vector field whose ODE transports a simple base distribution (e.g., Gaussian noise) to the data distribution. It was introduced by \cite{lipman2022flow}, with the key idea that instead of maximizing likelihood via expensive ODE simulation, one can train the CNF by regressing the model’s vector field to a known conditional velocity field induced by a chosen probability path connecting noise to data.

Let $\pi_0$ be a base distribution on $\mathbb{R}^d$ (e.g., $\cN(0,I)$) and $\pi_1=p_{\text{data}}$.
Choose a family of intermediate distributions $\{\rho_t\}_{t\in[0,1]}$ (a ``probability path'') such that
$\rho_0=\pi_0$ and $\rho_1=\pi_1$. A common construction specifies a conditional path
$\bx_t=\Psi_t(\bx_0,\bx_1,\epsilon)$ with $\bx_0\sim \pi_0$, $\bx_1\sim \pi_1$, and auxiliary noise $\epsilon$
(e.g., $\epsilon\sim\cN(0,I)$ for Gaussian paths). This induces a conditional distribution
$\rho_t(\cdot\mid \bx_1)$ and an associated conditional velocity field $u_t(\bx\mid \bx_1)$ (or $u_t(\bx\mid \bx_0,\bx_1)$),
often available in closed form for the chosen path. Flow Matching trains a neural vector field
$v_{\btheta}(\bx,t)$ by the regression objective
\[
\min_{\btheta}\ 
\mathbb E_{t\sim \mathrm{Unif}[0,1]}\ 
\mathbb E_{\bx_1\sim \pi_1}\ 
\mathbb E_{\bx\sim \rho_t(\cdot\mid \bx_1)}
\big\|v_{\btheta}(\bx,t)-u_t(\bx\mid \bx_1)\big\|_2^2.
\]
After training, generation is performed by deterministic ODE sampling: draw $\bx(0)\sim \pi_0$ and solve
\[
\frac{d\bx}{dt}=v_{\btheta}(\bx(t),t),\qquad t\in[0,1],
\]
to obtain a synthetic sample $\bx(1)$.

\subsection{Large Language Models}

Attention-based large language models (LLMs) use the Transformer architecture, built around self-attention to model and generate sequences. The foundational paper is \cite{vaswani2017attention}, which introduced multi-head self-attention as a fully parallel alternative to recurrence or convolutions for sequence modeling. Building on this architecture, modern LLMs are typically trained as auto-regressive models, where generation is done token-by-token.

An attention-based LLM defines an explicit probability model over token sequences $\bx_{1:T}$
(e.g., words, subwords, or bytes) by the autoregressive factorization
\[
p_{\btheta}(\bx_{1:T})=\prod_{t=1}^\top  p_{\btheta}(\bx_t\mid \bx_{<t}).
\]
Given a training corpus $\{\bx^{(i)}_{1:T_i}\}_{i=1}^n$, parameters $\btheta$ are learned by maximum likelihood
(equivalently, minimizing cross-entropy):
\[
\hat\btheta \in \arg\max_{\btheta} \sum_{i=1}^n \sum_{t=1}^{T_i}
\log p_{\btheta} \left(\bx^{(i)}_t \mid \bx^{(i)}_{<t}\right).
\]
The conditionals $p_{\btheta}(\cdot\mid \bx_{<t})$ are produced by a Transformer decoder with causal self-attention.
Writing the hidden states as $H\in\mathbb{R}^{t\times d}$, each attention layer forms
$Q=HW_Q,\ K=HW_K,\ V=HW_V$ and computes
\[
\mathrm{Attn}(Q,K,V)=\mathrm{softmax} \left(\frac{QK^\top}{\sqrt{d_k}}+M\right)V,
\]
where $M$ is a causal mask (future positions set to $-\infty$), and multi-head attention concatenates several
such heads.

After training, the LLM can generate synthetic sequences by sampling from the learned distribution $p_{\hat\btheta}$.
Starting from a prompt or a prefix $\bx_{1:t_0}$, the model outputs a categorical distribution
over the next token:
\[
\pi_{t}(\cdot)=p_{\hat\btheta}(\bx_t=\cdot\mid \bx_{<t}).
\]
A synthetic token is then drawn by a sampling rule such as ancestral sampling ($\bx_t \sim \mathrm{Categorical}(\pi_t)$) or temperature sampling ($\bx_t \sim \mathrm{Categorical}(\mathrm{softmax}(\mathrm{logits}/\tau))$). Appending the sampled token yields a longer
prefix $\bx_{1:t}$, and the process is repeated until an end-of-sequence token or a length cap is reached. The resulting
sequence $\tilde \bx_{1:\tilde T}$ is a synthetic sample from $p_{\hat\btheta}$ or from the conditional distribution
$p_{\hat\btheta}(\cdot\mid \bx_{1:t_0})$ when a prompt is used to steer attributes such as class, style, or covariates.


%% file: ref.bib
@article{shen2023boosting,
  title={Boosting data analytics with synthetic volume expansion},
  author={Shen, Xiaotong and Liu, Yifei and Shen, Rex},
  journal={The Annals of Applied Statistics},
  year={2026}
}

@article{keret2025glm,
  title={GLM Inference with AI-Generated Synthetic Data Using Misspecified Linear Regression},
  author={Keret, Nir and Shojaie, Ali},
  journal={arXiv preprint arXiv:2503.21968},
  year={2025}
}

@article{last2017oversampling,
  title={Oversampling for imbalanced learning based on K-means and SMOTE},
  author={Last, Felix and Douzas, Georgios and Bacao, Fernando},
  journal={arXiv preprint arXiv:1711.00837},
  year={2017}
}

@article{wang2025self,
  title={Self-improving generative foundation model for synthetic medical image generation and clinical applications},
  author={Wang, Jinzhuo and Wang, Kai and Yu, Yunfang and Lu, Yuxing and Xiao, Wenchao and Sun, Zhuo and Liu, Fei and Zou, Zixing and Gao, Yuanxu and Yang, Lei and others},
  journal={Nature Medicine},
  volume={31},
  number={2},
  pages={609--617},
  year={2025},
  publisher={Nature Publishing Group US New York}
}

@article{ktena2024generative,
  title={Generative models improve fairness of medical classifiers under distribution shifts},
  author={Ktena, Ira and Wiles, Olivia and Albuquerque, Isabela and Rebuffi, Sylvestre-Alvise and Tanno, Ryutaro and Roy, Abhijit Guha and Azizi, Shekoofeh and Belgrave, Danielle and Kohli, Pushmeet and Cemgil, Taylan and others},
  journal={Nature Medicine},
  volume={30},
  number={4},
  pages={1166--1173},
  year={2024},
  publisher={Nature Publishing Group US New York}
}

@article{lyu2025bias,
  title={Bias-Corrected Data Synthesis for Imbalanced Learning},
  author={Lyu, Pengfei and Ma, Zhengchi and Zhang, Linjun and Zhang, Anru R},
  journal={arXiv preprint arXiv:2510.26046},
  year={2025}
}

@inproceedings{he2008adasyn,
  title={{ADASYN}: Adaptive synthetic sampling approach for imbalanced learning},
  author={He, Haibo and Bai, Yang and Garcia, Edwardo A and Li, Shutao},
  booktitle={2008 IEEE International Joint Conference on Neural Networks (IEEE World Congress on Computational Intelligence)},
  pages={1322--1328},
  year={2008},
  organization={IEEE}
}

@article{chawla2002smote,
  title={{SMOTE}: Synthetic minority over-sampling technique},
  author={Chawla, Nitesh V and Bowyer, Kevin W and Hall, Lawrence O and Kegelmeyer, W Philip},
  journal={Journal of Artificial Intelligence Research},
  volume={16},
  pages={321--357},
  year={2002}
}

@book{efron1994introduction,
  title={An introduction to the bootstrap},
  author={Efron, Bradley and Tibshirani, Robert J},
  year={1994},
  publisher={Chapman and Hall/CRC}
}

@inproceedings{han2005borderline,
  title={Borderline-{SMOTE}: A new over-sampling method in imbalanced data sets learning},
  author={Han, Hui and Wang, Wen Yuan and Mao, Bing Huan},
  booktitle={International Conference on Intelligent Computing},
  pages={878--887},
  year={2005},
  organization={Springer}
}

@inproceedings{bunkhumpornpat2009safe,
  title={Safe-level-{SMOTE}: Safe-level-synthetic minority over-sampling technique for handling the class imbalanced problem},
  author={Bunkhumpornpat, Chumphol and Sinapiromsaran, Krung and Lursinsap, Chidchanok},
  booktitle={Pacific-Asia Conference on Knowledge Discovery and Data Mining},
  pages={475--482},
  year={2009},
  organization={Springer}
}

@article{tian2025conditional,
  title={Conditional data synthesis augmentation},
  author={Tian, Xinyu and Shen, Xiaotong},
  journal={arXiv preprint arXiv:2504.07426},
  year={2025}
}

@article{zhang2017mixup,
  title={{M}ixup: Beyond empirical risk minimization},
  author={Zhang, Hongyi and Cisse, Moustapha and Dauphin, Yann N and Lopez-Paz, David},
  journal={arXiv preprint arXiv:1710.09412},
  year={2017}
}

@article{ho2020denoising,
  title={Denoising diffusion probabilistic models},
  author={Ho, Jonathan and Jain, Ajay and Abbeel, Pieter},
  journal={Advances in neural information processing systems},
  volume={33},
  pages={6840--6851},
  year={2020}
}

@article{song2020score,
  title={Score-based generative modeling through stochastic differential equations},
  author={Song, Yang and Sohl-Dickstein, Jascha and Kingma, Diederik P and Kumar, Abhishek and Ermon, Stefano and Poole, Ben},
  journal={arXiv preprint arXiv:2011.13456},
  year={2020}
}

@article{lipman2022flow,
  title={Flow matching for generative modeling},
  author={Lipman, Yaron and Chen, Ricky TQ and Ben-Hamu, Heli and Nickel, Maximilian and Le, Matt},
  journal={arXiv preprint arXiv:2210.02747},
  year={2022}
}

@article{goodfellow2014generative,
  title={Generative adversarial nets},
  author={Goodfellow, Ian J and Pouget-Abadie, Jean and Mirza, Mehdi and Xu, Bing and Warde-Farley, David and Ozair, Sherjil and Courville, Aaron and Bengio, Yoshua},
  journal={Advances in neural information processing systems},
  volume={27},
  year={2014}
}

@article{kingma2013auto,
  title={Auto-encoding variational {B}ayes},
  author={Kingma, Diederik P and Welling, Max},
  journal={arXiv preprint arXiv:1312.6114},
  year={2013}
}

@article{newey1994large,
  title={Large sample estimation and hypothesis testing},
  author={Newey, Whitney K and McFadden, Daniel},
  journal={Handbook of econometrics},
  volume={4},
  pages={2111--2245},
  year={1994},
  publisher={Elsevier}
}

@article{rubin1981bayesian,
  title={The bayesian bootstrap},
  author={Rubin, Donald B},
  journal={The annals of statistics},
  pages={130--134},
  year={1981},
  publisher={JSTOR}
}

@article{kleiner2014scalable,
  title={A scalable bootstrap for massive data},
  author={Kleiner, Ariel and Talwalkar, Ameet and Sarkar, Purnamrita and Jordan, Michael I},
  journal={Journal of the Royal Statistical Society Series B: Statistical Methodology},
  volume={76},
  number={4},
  pages={795--816},
  year={2014},
  publisher={Oxford University Press}
}

@article{he2009learning,
  title={Learning from imbalanced data},
  author={He, Haibo and Garcia, Edwardo A},
  journal={IEEE Transactions on knowledge and data engineering},
  volume={21},
  number={9},
  pages={1263--1284},
  year={2009},
  publisher={Ieee}
}

@article{chen2024survey,
  title={A survey on imbalanced learning: latest research, applications and future directions},
  author={Chen, Wuxing and Yang, Kaixiang and Yu, Zhiwen and Shi, Yifan and Chen, CL Philip},
  journal={Artificial Intelligence Review},
  volume={57},
  number={6},
  pages={137},
  year={2024},
  publisher={Springer}
}

@article{matharaarachchi2024enhancing,
  title={Enhancing SMOTE for imbalanced data with abnormal minority instances},
  author={Matharaarachchi, Surani and Domaratzki, Mike and Muthukumarana, Saman},
  journal={Machine Learning with Applications},
  volume={18},
  pages={100597},
  year={2024},
  publisher={Elsevier}
}

@inproceedings{sohl2015deep,
  title={Deep unsupervised learning using nonequilibrium thermodynamics},
  author={Sohl-Dickstein, Jascha and Weiss, Eric and Maheswaranathan, Niru and Ganguli, Surya},
  booktitle={International conference on machine learning},
  pages={2256--2265},
  year={2015},
  organization={pmlr}
}

@article{song2020denoising,
  title={Denoising diffusion implicit models},
  author={Song, Jiaming and Meng, Chenlin and Ermon, Stefano},
  journal={arXiv preprint arXiv:2010.02502},
  year={2020}
}

@inproceedings{rombach2022high,
  title={High-resolution image synthesis with latent diffusion models},
  author={Rombach, Robin and Blattmann, Andreas and Lorenz, Dominik and Esser, Patrick and Ommer, Bj{\"o}rn},
  booktitle={Proceedings of the IEEE/CVF conference on computer vision and pattern recognition},
  pages={10684--10695},
  year={2022}
}

@article{radford2015unsupervised,
  title={Unsupervised representation learning with deep convolutional generative adversarial networks},
  author={Radford, Alec and Metz, Luke and Chintala, Soumith},
  journal={arXiv preprint arXiv:1511.06434},
  year={2015}
}

@article{mirza2014conditional,
  title={Conditional generative adversarial nets},
  author={Mirza, Mehdi and Osindero, Simon},
  journal={arXiv preprint arXiv:1411.1784},
  year={2014}
}

@article{chen2016infogan,
  title={Infogan: Interpretable representation learning by information maximizing generative adversarial nets},
  author={Chen, Xi and Duan, Yan and Houthooft, Rein and Schulman, John and Sutskever, Ilya and Abbeel, Pieter},
  journal={Advances in neural information processing systems},
  volume={29},
  year={2016}
}

@article{burda2015importance,
  title={Importance weighted autoencoders},
  author={Burda, Yuri and Grosse, Roger and Salakhutdinov, Ruslan},
  journal={arXiv preprint arXiv:1509.00519},
  year={2015}
}

@inproceedings{higgins2017beta,
  title={beta-vae: Learning basic visual concepts with a constrained variational framework},
  author={Higgins, Irina and Matthey, Loic and Pal, Arka and Burgess, Christopher and Glorot, Xavier and Botvinick, Matthew and Mohamed, Shakir and Lerchner, Alexander},
  booktitle={International conference on learning representations},
  year={2017}
}

@article{sohn2015learning,
  title={Learning structured output representation using deep conditional generative models},
  author={Sohn, Kihyuk and Lee, Honglak and Yan, Xinchen},
  journal={Advances in neural information processing systems},
  volume={28},
  year={2015}
}

@article{vaswani2017attention,
  title={Attention is all you need},
  author={Vaswani, Ashish and Shazeer, Noam and Parmar, Niki and Uszkoreit, Jakob and Jones, Llion and Gomez, Aidan N and Kaiser, {\L}ukasz and Polosukhin, Illia},
  journal={Advances in neural information processing systems},
  volume={30},
  year={2017}
}

@inproceedings{boser1992training,
  title={A training algorithm for optimal margin classifiers},
  author={Boser, Bernhard E and Guyon, Isabelle M and Vapnik, Vladimir N},
  booktitle={Proceedings of the fifth annual workshop on Computational learning theory},
  pages={144--152},
  year={1992}
}

@article{cortes1995support,
  title={Support-vector networks},
  author={Cortes, Corinna and Vapnik, Vladimir},
  journal={Machine learning},
  volume={20},
  number={3},
  pages={273--297},
  year={1995},
  publisher={Springer}
}

@article{schapire1990strength,
  title={The strength of weak learnability},
  author={Schapire, Robert E},
  journal={Machine learning},
  volume={5},
  number={2},
  pages={197--227},
  year={1990},
  publisher={Springer}
}

@article{freund1997decision,
  title={A decision-theoretic generalization of on-line learning and an application to boosting},
  author={Freund, Yoav and Schapire, Robert E},
  journal={Journal of computer and system sciences},
  volume={55},
  number={1},
  pages={119--139},
  year={1997},
  publisher={Elsevier}
}

@article{breiman2001random,
  title={Random forests},
  author={Breiman, Leo},
  journal={Machine learning},
  volume={45},
  number={1},
  pages={5--32},
  year={2001},
  publisher={Springer}
}

@article{rosenblatt1958perceptron,
  title={The perceptron: a probabilistic model for information storage and organization in the brain.},
  author={Rosenblatt, Frank},
  journal={Psychological review},
  volume={65},
  number={6},
  pages={386},
  year={1958},
  publisher={American Psychological Association}
}

@article{salmi2024handling,
  title={Handling imbalanced medical datasets: review of a decade of research},
  author={Salmi, Mabrouka and Atif, Dalia and Oliva, Diego and Abraham, Ajith and Ventura, Sebastian},
  journal={Artificial intelligence review},
  volume={57},
  number={10},
  pages={273},
  year={2024},
  publisher={Springer}
}

@article{chen2024interpretable,
  title={Interpretable machine learning for imbalanced credit scoring datasets},
  author={Chen, Yujia and Calabrese, Raffaella and Martin-Barragan, Belen},
  journal={European Journal of Operational Research},
  volume={312},
  number={1},
  pages={357--372},
  year={2024},
  publisher={Elsevier}
}

@article{bougaham2024composite,
  title={Composite score for anomaly detection in imbalanced real-world industrial dataset},
  author={Bougaham, Arnaud and El Adoui, Mohammed and Linden, Isabelle and Fr{\'e}nay, Beno{\^\i}t},
  journal={Machine Learning},
  volume={113},
  number={7},
  pages={4381--4406},
  year={2024},
  publisher={Springer}
}

@article{bunkhumpornpat2012dbsmote,
  title={DBSMOTE: density-based synthetic minority over-sampling technique},
  author={Bunkhumpornpat, Chumphol and Sinapiromsaran, Krung and Lursinsap, Chidchanok},
  journal={Applied Intelligence},
  volume={36},
  number={3},
  pages={664--684},
  year={2012},
  publisher={Springer}
}

@article{papamakarios2021normalizing,
  title={Normalizing flows for probabilistic modeling and inference},
  author={Papamakarios, George and Nalisnick, Eric and Rezende, Danilo Jimenez and Mohamed, Shakir and Lakshminarayanan, Balaji},
  journal={Journal of Machine Learning Research},
  volume={22},
  number={57},
  pages={1--64},
  year={2021}
}

@article{nakada2024synthetic,
  title={Synthetic oversampling: Theory and a practical approach using llms to address data imbalance},
  author={Nakada, Ryumei and Xu, Yichen and Li, Lexin and Zhang, Linjun},
  journal={arXiv preprint arXiv:2406.03628},
  year={2024}
}

@inproceedings{roy2024frauddiffuse,
  title={FraudDiffuse: Diffusion-aided Synthetic Fraud Augmentation for Improved Fraud Detection},
  author={Roy, Ruma and Tiwari, Darshika and Pandey, Anubha},
  booktitle={Proceedings of the 5th ACM International Conference on AI in Finance},
  pages={90--98},
  year={2024}
}

@article{stanton2024data,
  title={Data augmentation for predictive maintenance: Synthesising aircraft landing gear datasets},
  author={Stanton, Izaak and Munir, Kamran and Ikram, Ahsan and El-Bakry, Murad},
  journal={Engineering Reports},
  volume={6},
  number={12},
  pages={e12946},
  year={2024},
  publisher={Wiley Online Library}
}

@article{johnson2016mimic,
  title={MIMIC-III, a freely accessible critical care database},
  author={Johnson, Alistair EW and Pollard, Tom J and Shen, Lu and Lehman, Li-wei H and Feng, Mengling and Ghassemi, Mohammad and Moody, Benjamin and Szolovits, Peter and Anthony Celi, Leo and Mark, Roger G},
  journal={Scientific data},
  volume={3},
  number={1},
  pages={1--9},
  year={2016},
  publisher={Nature Publishing Group}
}

@article{singer2016third,
  title={The third international consensus definitions for sepsis and septic shock (Sepsis-3)},
  author={Singer, Mervyn and Deutschman, Clifford S and Seymour, Christopher Warren and Shankar-Hari, Manu and Annane, Djillali and Bauer, Michael and Bellomo, Rinaldo and Bernard, Gordon R and Chiche, Jean-Daniel and Coopersmith, Craig M and others},
  journal={Jama},
  volume={315},
  number={8},
  pages={801--810},
  year={2016},
  publisher={American Medical Association}
}

@article{raisa2025consistent,
  title={On consistent Bayesian inference from synthetic data},
  author={R{\"a}is{\"a}, Ossi and J{\"a}lk{\"o}, Joonas and Honkela, Antti},
  journal={Journal of Machine Learning Research},
  volume={26},
  number={74},
  pages={1--65},
  year={2025}
}

@article{bobkov2024berry,
  title={Berry-Esseen bounds in local limit theorems},
  author={Bobkov, Sergey and G{\"o}tze, Friedrich},
  journal={arXiv preprint arXiv:2407.20744},
  year={2024}
}

@article{xia2026classification,
  title={Classification Imbalance as Transfer Learning},
  author={Xia, Eric and Klusowski, Jason M},
  journal={arXiv preprint arXiv:2601.10630},
  year={2026}
}

@article{ahmad2025concentration,
  title={Concentration and excess risk bounds for imbalanced classification with synthetic oversampling},
  author={Ahmad, Touqeer and Kalan, Mohammadreza M and Portier, Fran{\c{c}}ois and Stupfler, Gilles},
  journal={arXiv preprint arXiv:2510.20472},
  year={2025}
}

@article{xia2024advancing,
  title={Advancing retail data science: Comprehensive evaluation of synthetic data},
  author={Xia, Yu and Wang, Chi-Hua and Mabry, Joshua and Cheng, Guang},
  journal={arXiv preprint arXiv:2406.13130},
  year={2024}
}

@article{feng2021imbalanced,
  title={Imbalanced classification: a paradigm-based review},
  author={Feng, Yang and Zhou, Min and Tong, Xin},
  journal={Statistical Analysis and Data Mining: The ASA Data Science Journal},
  volume={14},
  number={5},
  pages={383--406},
  year={2021},
  publisher={Wiley Online Library}
}

@article{rigollet2011neyman,
  title={Neyman-pearson classification, convexity and stochastic constraints},
  author={Rigollet, Philippe and Tong, Xin},
  journal={Journal of machine learning research},
  year={2011}
}

@article{tong2020neyman,
  title={Neyman-Pearson classification: parametrics and sample size requirement},
  author={Tong, Xin and Xia, Lucy and Wang, Jiacheng and Feng, Yang},
  journal={Journal of Machine Learning Research},
  volume={21},
  number={12},
  pages={1--48},
  year={2020}
}
